%% file: main.tex
\documentclass{article}

\usepackage{amsfonts} 
\usepackage{listings} 
\usepackage{url} 
\usepackage{cite} 
\usepackage{float}
\usepackage{amsthm,amssymb,amsmath}
\usepackage{xspace}
\usepackage[margin=1in]{geometry}

\usepackage{graphicx} 
\usepackage{subcaption} 
\graphicspath{{./}{figs/}}

\newtheorem{theorem}{Theorem}[section]

\theoremstyle{definition}

\theoremstyle{remark}
\newtheorem{remark}[theorem]{Remark}

\usepackage{lipsum}

\usepackage[utf8]{inputenc}
\usepackage{bbm}
\usepackage{algorithm}
\usepackage{algpseudocode}
\usepackage{amsthm}
\usepackage{amsfonts}
\usepackage{amsmath}
\usepackage{hyperref}
\usepackage{cleveref}
\usepackage{color}
\usepackage{xspace}
\usepackage[bb=dsserif]{mathalpha}
\usepackage{rotating}
\usepackage{mathrsfs}
\usepackage{rsfso}
\usepackage{rotating}
\usepackage{xspace}
\usepackage[normalem]{ulem}

\crefname{figure}{figure}{figures}
\crefname{table}{table}{tables}

\newcommand\boxdim{\dim_{\mathrm{box}}}
\newcommand{\mstdim}{\dim_{\mathrm{MST}}}
\newcommand{\phdim}{\dim_{\mathrm{PH}}}
\newcommand{\magdim}{\dim_{\mathrm{Mag}}}
\newcommand{\capdim}{\dim_{\mathrm{cap}}}

\newcommand{\geomle}{\ensuremath{\mathsf{GeoMLE}}\xspace}
\newcommand{\gride}{\ensuremath{\mathsf{GRIDE}}\xspace}
\newcommand{\phzero}{\ensuremath{\mathsf{PH}_0}\xspace}
\newcommand{\mle}{\ensuremath{\mathsf{MLE}}\xspace}
\newcommand{\mst}{\ensuremath{\mathsf{MST}}\xspace}
\newcommand{\pca}{\ensuremath{\mathsf{PCA}}\xspace}
\newcommand{\lpca}{\ensuremath{\mathsf{lPCA}}\xspace}
\newcommand{\mds}{\ensuremath{\mathsf{MDS}}\xspace}
\newcommand{\cdim}{\ensuremath{\mathsf{CDim}}\xspace}
\newcommand{\corrint}{\ensuremath{\mathsf{CorrInt}}\xspace}
\newcommand{\twonn}{\ensuremath{\mathsf{TwoNN}}\xspace}
\newcommand{\ess}{\ensuremath{\mathsf{ESS}}\xspace}
\newcommand{\essa}{\ensuremath{\mathsf{ESSa}}\xspace}
\newcommand{\essb}{\ensuremath{\mathsf{ESSb}}\xspace}
\newcommand{\wodcap}{\ensuremath{\mathsf{WODCap}}\xspace}
\newcommand{\danco}{\ensuremath{\mathsf{DANCo}}\xspace}
\newcommand{\knnestimator}{\ensuremath{\mathsf{KNN}}\xspace}
\newcommand{\fishers}{\ensuremath{\mathsf{FisherS}}\xspace}
\newcommand{\mindml}{\ensuremath{\mathsf{MiND\_ML}}\xspace}
\newcommand{\mindkl}{\ensuremath{\mathsf{MiND\_KL}}\xspace}
\newcommand{\idea}{\ensuremath{\mathsf{IDEA}}\xspace}
\newcommand{\tle}{\ensuremath{\mathsf{TLE}}\xspace}

\newcommand{\PHDim}{\ensuremath{\mathsf{PD}}\xspace}

\newcommand{\knn}{\ensuremath{\mathrm{knn}}\xspace}
\newcommand{\bpeps}{\ensuremath{B(p, \varepsilon)}\xspace}

\newcommand{\scikitdimension}{\texttt{scikit-dimension}\xspace}

\newcommand\bbr{\mathbb{R}}
\newcommand\rd{\mathbb{R}^D}
\newcommand\bbx{\mathbb{X}}
\newcommand\indicator{\mathbb{1}}
\newcommand{\sff}{{\mathrm{I\!I}}}

\newcommand\diff{\,\mathrm{d}}
\DeclareMathOperator{\diam}{diam}
\DeclareMathOperator*{\argmax}{argmax}
\DeclareMathOperator*{\argmin}{argmin}

\makeatletter
\DeclareRobustCommand\onedot{\futurelet\@let@token\@onedot}
\def\@onedot{\ifx\@let@token.\else.\null\fi\xspace}
\def\etal{\emph{et al}\onedot}
\makeatother

\makeatletter
\let\@fnsymbol\@arabic
\makeatother

\title{A Survey of Dimension Estimation Methods}

\author{James A. D. Binnie\thanks{Cardiff University, Cardiff, United Kingdom and Heilbronn Institute for Mathematical Research, Bristol, United Kingdom. Email:~\texttt{BinnieJA@cardiff.ac.uk}}
\and Pawe\l{} D\l{}otko\thanks{Dioscuri Centre in Topological Data Analysis,
IMPAN, Polish Academy of Sciences, Warsaw,
Poland. Email:~\texttt{pawel.dlotko@impan.pl}}
\and John Harvey\thanks{Cardiff University, Cardiff, United Kingdom. Email:~\texttt{HarveyJ13@cardiff.ac.uk}}
\and Jakub Malinowski\thanks{Wroc\l{}aw University of Science and Technology, Wroc\l{}aw, Poland and Dioscuri Centre in Topological Data Analysis, IMPAN, Polish Academy of Sciences, Warsaw, Poland. Email:~\texttt{jakub.malinowski@pwr.edu.pl}}
\and Ka Man Yim\thanks{Cardiff University, Cardiff, United Kingdom. Email:~\texttt{YimKM@cardiff.ac.uk}}
}

\begin{document}
\maketitle


\begin{abstract}

It is a standard assumption that datasets in high dimension have an internal structure which means that they in fact lie on, or near, subsets of a lower dimension. In many instances it is important to understand the real dimension of the data, hence the complexity of the dataset at hand.
A great variety of dimension estimators have been developed to find the intrinsic dimension of the data but there is little guidance on how to reliably use these estimators.

This survey reviews a wide range of dimension estimation methods, categorising them by the geometric information they exploit: tangential estimators which detect a local affine structure; parametric estimators which rely on dimension-dependent probability distributions; and estimators which use topological or metric invariants. 

The paper evaluates the performance of these methods, as well as investigating varying responses to curvature and noise. Key issues addressed include robustness to hyperparameter selection, sample size requirements, accuracy in high dimensions, precision, and performance on non-linear geometries. In identifying the best hyperparameters for benchmark datasets, overfitting is frequent, indicating that many estimators may not generalise well beyond the datasets on which they have been tested.
\end{abstract}

\section{Introduction}
\label{sec:intro}

Data analysis in high-dimensional space poses significant challenges due to the ``curse of dimensionality''.
However, the data are often believed to lie in a latent space with lower \emph{intrinsic dimension} than that of the ambient feature space.
Techniques that estimate the intrinsic dimension of data inform how we visualise data in biological settings~\cite{Way2020CompressingRepresentations, Zang2021AbstractSets, Risso2018AData}, build efficient classifiers~\cite{Geng2005SupervisedClassification}, forecast time series~\cite{Papaioannou2022Time-seriesHarmonics}, analyse crystal structures~\cite{Oganov2009HowSolids} and even assess the generalisation capabilities of neural networks~\cite{Birdal2021IntrinsicNetworks,Dupuis2023GeneralizationDimensions,Andreeva2023MetricNetworks,Simsekli2020HausdorffNetworks}. Moreover they have been proposed as being able to distinguish human written text from AI written text~\cite{Tulchinskii2023IntrinsicTexts}.
Accurately identifying the intrinsic dimension of the dataset is also key to understanding the accuracy, efficiency and limitations of dimensionality reduction methods~\cite{VanDerMaaten2008VisualizingT-SNE,Balasubramanian2002TheStability., McInnes2018UMAP:Projection}.

We study how estimators respond to practical challenges such as choosing appropriate hyperparameters, and the presence of noise and curvature in data. We do so by assessing their performance on a standard set of benchmark manifolds described in~\cite{HeinMatthias2005IntrinsicRd,Campadelli2015IntrinsicFramework,Rozza2012NovelEstimators}.
The datasets encompass a large range of intrinsic dimensions (1 to 70), codimensions (0 to 72), and ambient dimensions of the embedding space (3 to 96) as well as different geometries (flat, constant curvature, variable curvature).

Dimension estimators have been regularly surveyed, reflecting both their importance and the productivity of the field~\cite{Camastra2003DataSurvey, Campadelli2015IntrinsicFramework, Camastra2016IntrinsicProblems}.
As it is now nine years since the last survey, a renewed picture of the field is appropriate.
Key developments include the growth of topological data analysis, which adds new types of estimators.
We pay particular attention to the geometric underpinnings of the different estimators, focussing on the geometric information used in the estimation process and the impact this has, rather than on the \emph{local, global, pointwise} division.
In carrying out this survey, we have also significantly extended the capability of the \scikitdimension package~\cite{Bac2021Scikit-Dimension:Estimation} by adding a wide variety of estimators.

\subsection{What is dimension?}

In mathematics there are many different notions of dimension. These come from a wide variety of areas of mathematics: geometry, topology, algebra, dynamical systems etc. In this section, we survey the main different concepts of dimension. In the following, we will specialize to a
certain subset of the concepts discussed below;

\paragraph{Algebraic (Hamel) Dimension} Perhaps the simplest notion of dimension which can be defined is the algebraic dimension of a vector space $V$, which is  the cardinality of a basis $B$ of $V$ over the base field $F$.

\begin{figure}
    \centering
    \includegraphics[width=0.4\linewidth]{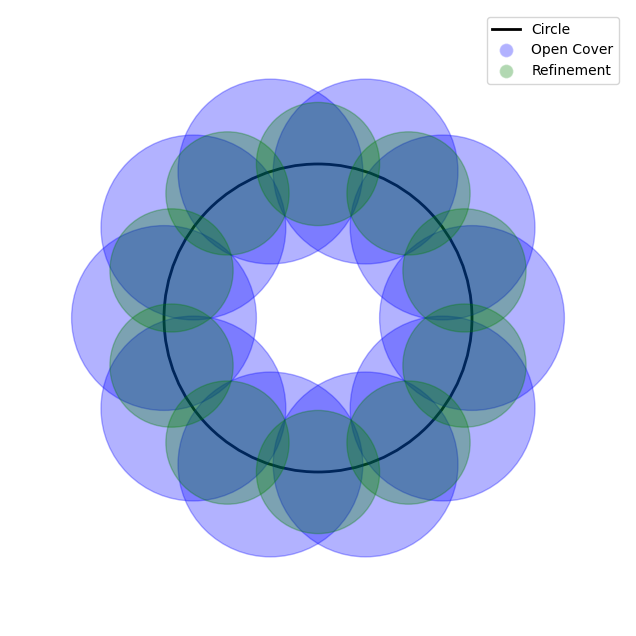}
    \caption{Example of a cover and a refinement. There are points on the circle which are contained in three elements of the initial cover. After refining, every point is contained in at most two elements. It would not be possible to refine until each point was contained in only one element of the cover, so that $\dim = 1$.}
    \label{fig:lcdEx}
\end{figure}

\paragraph{Topological (Lebesgue Covering) Dimension} A topological space $X$ is said to be finite dimensional if there is some integer $d$ such that for every open covering $\mathscr{A}$ of $X$, there is an open covering $\mathscr{B}$ of $X$ that refines $\mathscr{A}$ so that each $x \in X$ is contained in at most $d+1$ members of $\mathscr{B}$. The topological dimension of X is defined to be the smallest value of $d$ for which this statement holds\cite{Munkres2000TopologyEdition}. In the case of a topological vector space, its topological dimension is equal to its Hamel dimension~\cite{Tychonoff1935EinFixpunktsatz}. The concept is illustrated in \Cref{fig:lcdEx}.

\paragraph{Hausdorff Dimension}
We now specialise further to metric spaces. Let $X$ be a metric space and fix $d\ge0$. For each $\delta>0$, consider any countable cover $\{U_i\}$ of $X$ with $\diam(U_i)<\delta$ for all $i$, where $\diam(U_i)$ denotes the diameter of $U_i$. Define
\[
\mathcal{H}^d_\delta(X)
\;=\;\inf\Bigl\{\sum_i\bigl(\diam(U_i)\bigr)^d : X\subset\bigcup_iU_i,\ \diam(U_i)<\delta\Bigr\},
\]
and then
\[
\mathcal{H}^d(X)
\;=\;\lim_{\delta\to0}\mathcal{H}^d_\delta(X).
\]
If $d$ is too large, one finds $\mathcal{H}^d(X)=0$; if $d$ is too small, $\mathcal{H}^d(X)=+\infty$. The \emph{Hausdorff dimension} of $X$ is
\[
\dim_H(X)=\inf\{\,d\ge0:\mathcal{H}^d(X)=0\},
\]
equivalently the unique $d_0$ for which
\[
\mathcal{H}^d(X)=
\begin{cases}
+\infty,&d<d_0,\\
0,&d>d_0,
\end{cases}
\quad
\mathcal{H}^{d_0}(X)\in[0,\infty].
\]
This definition captures ``fractional'' scaling behavior—sets like the Cantor set or Julia sets often have non‐integer $\dim_H$~\cite{Bishop2016FractalsAnalysis}.

\paragraph{Box‐Counting (Minkowski‐Bouligand) Dimension}
Let $X\subset\mathbb{R}^n$. For each $\epsilon>0$, let $N(\epsilon)$ be the minimum number of $n$‐dimensional boxes (closed cubes) of side length $\epsilon$ required to cover $X$. Since the covering elements here are restricted to equal‐sized boxes, one records how $N(\epsilon)$ grows as $\epsilon\to0$. The box‐counting dimension $d_B$ of $X$ is defined as~\cite{Falconer2003AlternativeDimension}:
\[
d_B \;=\; \lim_{\epsilon\to0} \frac{\log N(\epsilon)}{\log\bigl(1/\epsilon\bigr)}.
\]
In other words, if $N(\epsilon)\sim\epsilon^{-d}$ as $\epsilon\to0$, then $d_B=d$. Note that this is a special (and generally coarser) case of the Hausdorff dimension, corresponding to restricting all coverings in the Hausdorff construction to equal‐sized boxes.

\paragraph{Correlation Dimension}
The correlation dimension $d_{\mathrm{C}}$ is a way to measure the dimensionality a space with a probability measure, particularly in applications like dynamical systems and data science. 
The correlation integral captures the spatial correlation between points by finding the probability that a pair of points $(x,y)$ is separated by a distance less than $\varepsilon$.
Let $\mu$ be a probability measure on $M$.
\[
C(\epsilon) = \int_{M \times M} \indicator_{\|x - y\| < \epsilon} \diff \mu^{\otimes 2},
\]
The correlation dimension $d_{\mathrm{C}}$ is defined as:
\[
d_C = \lim_{\epsilon \to 0} \frac{\log C(\epsilon)}{\log \epsilon}.
\]
This dimension estimates the scaling behaviour of point-pair correlations and is especially relevant for analysing the structure of attractors in dynamical systems. It is bounded above by the information dimension, another probability-based dimension, which is in turn bounded above by the Hausdorff dimension of $M$~\cite{Grassberger1983MeasuringAttractors}. Since it is dependent on the measure $\mu$, it takes account of the varying density of attractors.

\paragraph{Equivalence of dimensions for manifolds}
Consider the specific case of a smooth submanifold $M^d \subset\rd$. 
The tangent space at a point  $p \in M^d$, $T_pM$, is isomorphic to $\bbr^d$ and so will have algebraic dimension $d$. 
The Lebesgue covering dimension of a smooth manifold of dimension $d$ is $d$~\cite{vanMill1988Infinite-DimensionalTopology}.
Endowing the manifold with the Riemannian metric inherited from $\rd$, we can also calculate its dimension as a metric space.
The Hausdorff dimension of a Riemannian manifold is the same as its topological dimension~\cite{Mattila1995GeometrySpaces}, in this case $d$.
We also have that $M^d \subset \rd $ has box-counting dimension $d$~\cite{Falconer2003AlternativeDimension}.
If $\mu$ is a probability measure on $M$ which is absolutely continuous with respect to the volume measure, then the correlation dimension is the growth rate of tubes around the diagonal submanifold in $M \times M$, which can again be seen to be $d$ (by a result of Gray and Vanhecke reported in~\cite{Gray1982ComparisonFormula}).
Hence we have that for a smooth submanifold of $\rd$ of dimension $d$ all notions of dimension described above coincide.

\subsection{Intrinsic Dimension} 

If a dataset is generated from a probability measure supported on some subset $M \subset \rd$, the intrinsic dimension of $M$ can be defined using the notions of dimension described above.
This is the dimension of the latent space, where the internal structure of the data is believed to originate and on which the generating process of the data is ultimately defined.

For a given dataset $\bbx \subset \rd$, the intrinsic dimension has been described as being the ``minimum number of parameters necessary to describe the data'', and presented as equivalent to the intrinsic dimensionality of the data generating process itself~\cite{Fukunaga1971AnData}.
However, regardless of what $\bbx$ is, it is always possible to embed a 1-dimensional curve into $\rd$ in such a way that every data point in $\bbx$ lies on the curve (or, indeed, a disconnected 0-manifold comprised of precisely the points of $\bbx$).

When a very low-dimensional manifold of this type is used to model the data, the neighbourhood structure of the data is not preserved.
As a result, this representation of the data involves significant information loss.
Since the dataset does not come with a specification of the true latent neighbourhood structure, though, it is not possible to determine precisely when a representation does or does not involve information loss.

One might attempt to define intrinsic dimensionality of a dataset instead as the smallest $d$ such that there is a $d$-dimensional space $M$ which can be mapped into $\rd$ in a geometrically controlled way so that every point of $\bbx$ lies in the image of $M$.
The nature of the geometric control depends on what types of generating process we might consider and is related to the question of what the neighbourhood structure of $\bbx$ is.
The weaker the geometric control is, the more embeddings are considered, until eventually every point on $\bbx$ is permitted to lie on the image of a curve of possibly very high curvature.

If $M$ is understood to be an embedded submanifold, then control on the reach of $M$ is appropriate~\cite{Kim2019MinimaxManifold}.
Indeed, as manifolds of lower reach are considered, the probability of error in estimating the dimension grows.
It is therefore not possible to assign an intrinsic dimension to a dataset.
Only the underlying support of the distribution has an intrinsic dimension and aspects of the geometry of this support and of the distribution itself influence how challenging it is to estimate this dimension.

In this paper, we address the following core problem: 
construct a statistical procedure, an estimator \[\hat d_N : \left(\rd\right)^N \to \bbr,\]
such that if the finite point cloud \(\bbx = \{x_1, x_2, \dots, x_N\} \subset \rd\) 
is independently identically distributed according to an unknown distribution supported on a subset \(M\) 
of true intrinsic dimension \(d\) 
then $\hat d_N$ satisfies the consistency requirement
\[
\forall\,\varepsilon>0,\quad \lim_{N\to\infty} \Pr\bigl(|\hat d_N - d| > \varepsilon\bigr) \;=\; 0.
\]
Equivalently, \(\hat d_N\) must converge (in probability, or almost surely under stronger assumptions) to the true dimension \(d\) as the sample size \(N\) grows. Beyond mere convergence, one typically seeks quantifiable rates of convergence and robustness to perturbations—such as additive noise or non‐zero curvature on \(M\). 

Many estimators which we will consider take a hyperparameter $k$ which defines a local region in the dataset. Consistency will sometimes be known under the additional assumptions that $k \to \infty$ while $k/N \to 0$. The first condition requires that every local neighbourhood contains arbitrarily many points, while the second requires the local neighbourhoods to be arbitrarily small in comparison to the global dataset.

This paper will consider, compare and contrast a variety of estimators, evaluating them with respect to their computational feasibility, their consistency and convergence on a set of benchmark synthetic datasets and their sensitivity to hyperparameter selection.

\subsection{High-dimensional phenomena}

A set of $N$ points in Euclidean space $\rd$ will always, if $D > N$, lie on an affine subspace of dimension $N-1$.
This is the simplest example of a general principle that, in a high-dimensional space, a large number of points are necessary for the additional dimensions to be evident from the geometry of the point set alone.

It is clear from the relationship between volume and dimension that, to fully explore a region of $D$-dimensional space, a number of points which grows exponentially with $D$ is needed.
However, this need for a large sample size is still present even for the high dimensionality to have a much more limited influence on geometry.
For a set of $N$ points, if we require that the points can be mapped to a lower-dimensional space with relative distance distortion $\varepsilon$, the Johnson--Lindenstrauss Lemma~\cite{Johnson1984ExtensionsSpace} shows that the lower-dimensional space can have dimension $8 \varepsilon^{-2} \log N$, so that in the presence of any deviation from linearity the number of points needed to detect the dimensionality of the space by projection grows exponentially with $D$.

Uniform distributions of points in high-dimensional balls and cubes have been widely studied and demonstrate a variety of interesting behaviours that create statistical challenges.
This is discussed in more detail in Section \ref{sec:underestimation}.

All of these phenomena both provide a compelling reason to carry out dimension reduction as well as creating a challenge in estimating the dimension of higher-dimensional spaces.

\subsection{Notation}
In this paper, we assume that our finite sample lies on a smooth $d$-dimensional manifold $M^d \subset \mathbb{R}^D$. Concretely, let $M^d$ be a compact, smooth submanifold of $\mathbb{R}^D$, and let
\[
f : M^d \rightarrow \mathbb{R}
\]
be a probability density on $M^d$ (with respect to the $d$-dimensional Hausdorff measure). We write
\[
\bbx = \{\,x_1, x_2, \ldots, x_N\} \subset M^d
\]
for a finite sample of size $N$, drawn i.i.d.\ from the distribution defined by $f$.  

Many dimension estimation procedures are local in nature, so for each point $p$ of interest we need to describe a neighbourhood of $p$ within $\bbx$. 

  For any point $p\in\mathbb{R}^D$, let 
  \(
    r_k(p)
  \)
  denote the Euclidean distance from $p$ to its $k$th nearest neighbour in $\bbx$. If $p\in \bbx$, then by convention we do \emph{not} count $p$ itself as its own nearest neighbour, so that $r_1(p)>0$.  
  The ratio between two nearest neighbours will be denoted by
  \[
  \rho_{i,j}(p) = \frac{r_i(p)}{r_j(p)}.
  \]

  We write 
  \[
    \knn(p; k)
    \;=\;
    \bigl\{\,x \in \bbx : \|x - p\| \le r_k(p)\bigr\}
  \]
  for the set of $k$ nearest neighbours of $p$ in $\bbx$. When the value of $k$ is clear from context, we simply write $\knn(p)$.  
  For any $\varepsilon > 0$, define
  \[
    \bpeps
    \;=\;
    \bigl\{\,x \in \bbx : \|x - p\| < \varepsilon \bigr\}.
  \]
  Thus $\bpeps$ is the subset of $\bbx$ lying inside the open Euclidean ball of radius $\varepsilon$ centered at $p$. 
  Denote by $N(p,\varepsilon)$ the number of points in $\bpeps$, not counting $p$ itself.

Both $\knn(p)$ and $\bpeps$ provide local neighbourhoods of $p$ within $\bbx$, which we will use for various local dimension estimates.

\section{Families of Estimators}\label{sec:estimator_description}

The authors have systematically searched the literature to identify papers on dimension estimation published after the most recent survey (\cite{Camastra2016IntrinsicProblems}, which provides a good contextual overview of methods previous to 2015/16), indexed by Google Scholar using the search terms ``intrinsic dimension'' and ``dimension estimation''. Another survey of previous techniques from the same time~\cite{Karbauskaite2016Fractal-BasedSurvey} focuses on fractal dimensions.

This search is complicated by the vast number of diverse disciplines which are interested in the problem. 
The disparate nature of the work in the field means that new estimators are not always well contextualised within the historical research. 
Practitioners often appear unaware of well-established techniques. 

While we place a focus on papers published after 2015, we also seek to provide a description of the most essential established techniques.
This paper cannot describe every estimator in detail, and our reference list may not give an entirely comprehensive description of recent developments.
We encourage practitioners to consider more recent estimators rather than just relying on those that are familiar.

The dimension estimators which have been described in the literature fall into several broad families depending on which underlying geometric information is used to infer the intrinsic dimension.

\subsection{Tangential Estimators}

A small neighbourhood in a smooth manifold can always be viewed as the graph of a function in the following way.
Let $p \in M^d \subset \rd$.
After translating and rotating, we may assume that $p=0$ and $T_p M = \bbr^d$.
Then, near $p$, the manifold is the graph of a function $F \colon \bbr^d \to \rd$ and, furthermore, the differential $\diff F$ is the standard embedding $\bbr^d \to \rd$ along the first $d$ co-ordinates. 

The points in $\knn(p;k)$ or in $\bpeps$, for sufficiently small $k$ or $\varepsilon$, are therefore concentrated around an affine subspace of $\rd$ which is the tangent space to the manifold.
Many estimators seek to identify the dimension of the manifold by finding this affine subspace.
In this section we describe some of these, which we refer to as \emph{tangential estimators}.

\subsubsection{Principal components analysis}

Principal components analysis ($\pca$) is a long-established method for fitting an affine subspace to a sample of points. 
By applying this to $\knn(p)$ or $\bpeps$ we can estimate the dimension of the tangent space by considering eigenvectors corresponding to the dominant eigenvalues. More precisely since the eigenvalues $\lambda_1, \ldots, \lambda_D$ (always taken in decreasing order) of the covariance matrix of the data are in proportion to the amount of variance explained by each principal component, the first $d$ eigenvalues are expected to be large while the remaining $D-d$ eigenvalues are non-zero only in the presence of curvature (with the leading term of the deviation given by the second fundamental form $\sff_p$) or noise. This method is known as local principal components analysis ($\lpca$)~\cite{Fukunaga1971AnData, Fan2010IntrinsicAnalysis, Lin2008RiemannianLearning}.  The key decision in designing an $\lpca$ estimator is in determining a suitable threshold for identifying the ``large'' eigenvalues.

There are a number of thresholding methods giving rise to different variations of \lpca. We discuss some of them below.

\paragraph{Threshold parameters}

Some methods involve setting a parameter value which can be used to determine which eigenvalues are ``large''. As described below in Section \ref{sec:hyperparameter_choice}, the output of the estimator can be very sensitive to the selected value.

Fukunaga and Olsen~\cite{Fukunaga1971AnData} compare each eigenvalue to the largest, identifying an eigenvalue as ``large'' if it exceeds a certain proportion of the largest one. This gives an estimated dimension of $\max \{ u : \lambda_u > \alpha \lambda_1 \}$ for some fixed parameter $0 < \alpha < 1$.

Fan \etal use two different thresholding methods and recommend using the lower of the two estimates~\cite{Fan2010IntrinsicAnalysis}. 
One is to consider the ``gap'' $\frac{\lambda_u}{\lambda_{u+1}}$. If this number is large, there is little additional explained variance in adding a dimension, and a parameter $\beta > 1$ may be chosen so that the estimated dimension is $\min \{ u : \frac{\lambda_u}{\lambda_{u+1}} < \beta \}$.

Another is based on the total proportion of the variance estimated, so that the dimension is given by $\min \{ u : \sum_{i=1}^u \lambda_i > \gamma \cdot \sum_{i=1}^N \lambda_i \}$ for some parameter $0 < \gamma < 1$.
This method has been validated by Lim \etal~\cite{Lim2024TangentDistance}, who showed that with theoretical bounds on sampling the dimension is correctly estimated with high probability.

Kaiser takes a very simple approach to determining which eigenvalues are dominant, using only those which are above the mean eigenvalue~\cite{Kaiser1960TheAnalysis}. However, this can give very low estimates in cases of low codimension. Jolliffe has suggested that this should be softened somewhat, to 70\% of the mean~\cite{Jolliffe2002PrincipalAnalysis}. Clearly any other proportion could be used, so that this is a user-chosen parameter.

\paragraph{Probabilistic methods}

Other thresholding methods seek to label the $d$th eigenvalue as dominant if it is sufficiently larger than it would be under a null hypothesis that the eigenvalues arise solely from random variation. The challenge here is in selecting an appropriate distribution of eigenvalues to model the null hypothesis.

Frontier uses the ``broken stick'' distribution~\cite{Frontier1976EtudeBrise}. This distribution is generated by drawing $D-1$ uniform random variables on $[0,1]$ and arranging the lengths of the resulting $D$ subintervals in descending order. The assumption behind this method is that, if the data is unstructured, the eigenvalues will be distributed according to the lengths of these subintervals. However, there does not appear to be any theoretical basis for choosing this distribution.
 
A permutation test approach is given by Buja and Eyuboglu~\cite{Buja1992RemarksAnalysis} in the context of \pca, but it can easily be adapted for \lpca. If the local data is represented as a $k \times D$ matrix, each column can be reshuffled independently to give an alternative dataset. Each feature has the same variance, but the internal structure of the dataset has been removed. Carrying out \pca on these unstructured sets yields a null distribution for the eigenvalues. 

Minka~\cite{Minka2001AutomaticPCA} adopts a Bayesian model selection approach by maximising the Bayes evidence in fitting the data to normal distributions of different dimensions, extending the maximum likelihood framework of~\cite{Bishop1998BayesianPCA}. 

\paragraph{Other methods}

Craddock and Flood~\cite{Craddock1969EigenvectorsHemisphere} argue that eigenvalues corresponding to ‘noise’ should decay in a geometric progression. Plotting $\log \lambda_i$ against $i$, these eigenvalues will therefore appear as a trailing straight line, while dominant eigenvalues are those before the line. This gives a dimension estimate as the value $d$ such that $\lambda_{d+1}$ is the first point on the trailing straight line.

\pca based approaches have been augmented with an autoencoder for residual estimation~\cite{Karkkainen2023AdditiveEstimation}. For large sample sizes, alternative methods of finding eigenvalues have been proposed~\cite{Ozcoban2025AEstimation}. Where missing features pose a difficulty in using \pca, a method has been proposed in~\cite{Risso2018AData}.

\subsubsection{Kernel \pca}

An alternative form of $\pca$ is ``kernel $\pca$''. Rather than considering the data points themselves, it is possible to consider the inner products between points in some higher dimensional space. This inner product is known as a ``kernel''. If other kernels are used, such as similarity measures, it is possible to carry out $\pca$ on this similarity matrix to estimate the dimension.

The kernel mitigates the key limitation of \pca: that it is exact only in a linear case. It does so by implicitly mapping the data to a very high dimensional space $\phi : X \rightarrow H$ in the hope that the data lie on a linear manifold in this high dimensional space, then implicitly applying \pca in that space. The embedding itself is not carried out -- there is no need to explicitly know the map $\phi$. 
It is sufficient to know the kernel map $\mathcal{K} : X \times X \rightarrow \mathbb{R}_{+}$ which for each pair $x_i, x_j \in X$ yields $\mathcal{K}(x_i,x_j) = \langle \phi(x_i),\phi(x_j)\rangle$.
The matrix $K$ defined by $K_{ij} := \mathcal{K}(x_i,x_j)  $ is thus the Gram matrix of the set of points in this high dimensional feature space. 
Any symmetric positive semidefinite function can be used as a kernel, such as radial basis functions, quadratics, etc~\cite{Ghojogh2023ElementsLearning}.
When a linear kernel is used, kernel \pca reduces to the usual \pca.

The method of multidimensional scaling ($\mds$) is sometimes described as an alternative approach to \pca based methods, aiming to embed the data set in a lower-dimensional space to minimize a strain function. 
In classical \mds, the strain function to be minimised is the distortion of the Euclidean metric, while in generalised classical \mds it could be any valid kernel function.
Analysis of these cases shows that, in fact, these are equivalent to \pca and kernel \pca respectively~\cite{Ghojogh2023ElementsLearning}.
Despite their separate histories, then, the methods of kernel $\pca$ and $\mds$ are in fact not theoretically distinct.
Other forms of \mds move beyond this, using similarity functions which do not come from inner products.

Methods developed from the \mds school of thought tend to be focussed on producing global embeddings of the data in lower-dimensional space, rather than identifying its intrinsic dimension.
Examples include Isomap~\cite{Tenenbaum2000AReduction}, Sammon mapping~\cite{Sammon1969AAnalysis} and Shepard--Kruskal scaling~\cite{Kruskal1964NonmetricMethod}

\subsubsection{Nearest neighbour methods}

Another category of estimators aims to identify the tangent space by studying the directions to nearest neighbours. 
For any point $p \in \bbx$, the edges in the $\knn$ graph incident to $p$ represent directions which lie close to $T_p M$, provided the neighbours of $p$ are sufficiently close.

In~\cite{Lin2008RiemannianLearning}, edges are removed from the $\knn$ graph to emphasise the local structure.
A neighbour $p_i$ of $p$ can be considered to be ``behind'' another neighbour $p_j$ if the angle $\measuredangle p p_j p_i > \pi/2$.
By removing the edge from $p$ to $p_i$, only neighbours ``visible'' from $p$ remain connected to $p$.
By carrying out \pca on the edge set incident to $p$, beginning with the shortest edge and adding edges in order of increasing length, it is possible to identify which edges result in an increase in dimension.
Where these edges are larger than the preceding edges by some threshold parameter, they are identified as potentially ``short-circuiting'' the graph and are also removed.
The dimension estimate is the dimension of the largest simplex in the clique complex built from this reduced $\knn$ graph.

The conical dimension ($\cdim$)~\cite{Yang2007ConicalApplications} takes a slightly different approach -- its value at $p$ is the dimension of the smallest affine subspace $W$ through $p$ so that the angle at $p$ between each element of $\knn(p)$ and $W$ is less than $\pi/4$. This method is proven to be correct under conditions on the reach and the sampling density, conditions which require an exponentially large sample size in each neighbourhood.

\subsection{Parametric Estimators}

The assumption that the data lie on a manifold with locally approximately uniform density determines the approximate distributions of the number of points in a particular volume, of distances to nearest neighbours, and of angles subtended at a point by two neighbours. The dimension of the manifold is a parameter in these probability distributions and it can be estimated using parametric methods. 

In this section we considered methods based on parametric estimators. All of these methods can be expected to be less accurate if the underlying assumptions fail. Variable density, curvature, boundary and spatial correlation will all introduce error. Since larger neighbourhoods are more likely to suffer from these problems, and since these phenomena are also more evident on larger neighbourhoods, it is reasonable to expect increasing bias with $k$. However, as our results show, this does not always appear to be the case.

Unlike tangential estimators, the geometry and topology of the ambient space is generally not taken into account, so that estimators can produce values which are higher than the dimension of the ambient space.

\subsubsection{Volume growth}

Volumes of balls have also been used for parametric estimation from an early date. If $p \in \bbx$ is a fixed point, then the proportion of sampled points lying in a ball of radius $r_k(p)$ is, empirically, approximately $\frac{k}{N-1}$.
In other words, recalling that $f$ is the density on $M$ we have
\[\frac{k}{N-1} \simeq f(p) V_d r_k^d(p)\]
for small enough $r_k$, assuming $f$ is almost constant near $p$. Here \[V_d = \frac{\pi^{\frac{d}{2}}}{\Gamma(\frac{d}{2}+1)}\] is the volume of the unit ball in $\bbr^d$.

It is the rate of growth of the volume that is related to the dimension, so that another key equation is 
\[\frac{N(p,\varepsilon_2)}{N(p,\varepsilon_1)} \simeq \left(\frac{\varepsilon_2}{\varepsilon_1}\right)^d\]
for $0 < \varepsilon_1 < \varepsilon_2$, which holds if $B(p,\varepsilon_2)$ is close enough to lying in an affine space and the density $f$ can be assumed to be constant on $B(p,\varepsilon_2)$.
This is the basis of the estimator in~\cite{Farahmand2007Manifold-adaptiveEstimation}, which was later elaborated on in~\cite{Benko2022Manifold-adaptiveRevisited}

Volume-based estimators have also been developed for discrete spaces, such as lattices~\cite{Macocco2023IntrinsicMetrics}.

\paragraph{Correlation Integral}
The correlation integral $\corrint$~\cite{Grassberger1983MeasuringAttractors} is a global approach motivated by this geometry.
It is used in chaos theory to indicate the level of spatial correlation between points in the attractor. 
Let $\Delta(\varepsilon) \subset \bbr^{2D}$ be the neighbourhood of the diagonal given by $\Delta(\varepsilon) = \{ (x,y) \colon \|x-y\|<\varepsilon\}$.
Then the function \[C(\varepsilon) = \lim_{N \to \infty} \frac{1}{N^2} \sum_{i,j} \indicator_{\Delta(\varepsilon)}(x_i,x_j),\] known as the \emph{correlation integral}, counts the proportion of pairs of points lying within distance $\varepsilon$ of each other.
The growth rate of this function for small $\varepsilon$, $C(\varepsilon) \sim \varepsilon^d$, gives the correlation dimension of the attractor. This depends on the measure on the attractor rather than on the geometry of its support.

The value can be estimated by assuming that the fixed finite sample size $N$ is sufficiently large that \[C(\varepsilon) \simeq \frac{1}{N^2} \sum_{i,j} \indicator_{\Delta(\varepsilon)}(x_i,x_j),\] and approximating the slope with
\[ \frac{\log \sum_{i \neq j} \indicator_{\Delta(r_2)}(x_i,x_j) - \log \sum_{i \neq j} \indicator_{\Delta(r_1)}(x_i,x_j)}{\log r_2 - \log r_1} \]
for two suitable values $0 < r_1 < r_2$.
Here pairs with $i=j$ are ignored; in the large sample limit their contribution is negligible but for a finite sample they have the potential to distort the estimate.

Variants of this estimator have been proposed in~\cite{Camastra2001IntrinsicAlgorithm, Erba2019IntrinsicData, Krakovska2023SimpleCausality, Qiu2022IntrinsicMethod}.

The correlation dimension is sensitive to the measure. If the focus is purely geometric, this may not be a good choice.
The packing number method proposed by Kegl~\cite{Kegl2002IntrinsicNumbers} instead uses the capacity dimension, which will be described later, making this method particularly useful when the data are unevenly distributed.

The capacity dimension of a set $S$ is defined by the equation
\[\capdim(S) = \lim_{r \rightarrow 0 } \frac{\log(N(r))}{\log(r)},\]
where $N(r)$ is the $r$-covering number (the number of open balls of radius $r$ required to cover set $S$). It is worth noting that if both the topological dimension and capacity dimension exist for a given set, they are equal to each other.
For ease of computation, the $r$-packing number $M(r)$ can be used instead of $N(r)$. 
Given a metric space $X$ with distance metric $d(\cdot, \cdot)$, 
the set $U \subset X$ is said to
be $r$-separated if $d(x, y) \geq r$ for all distinct $x, y \in U$ . 
The $r$-packing number $M(r)$ of a set
$S \subset X$ is defined as \[M(r) = \max_{U \subset S} \{ |U| : U \text{ is $r$-separated} \}.\] It follows from the inequalities $N(r) \leq M(r) \leq N(r/2)$ that
\[\capdim(S) = \lim_{r \rightarrow 0 } \frac{\log(M(r))}{\log(r)}.\]

As with \corrint, this limit can only be approximated when using a finite sample. 
The dimension is estimated by the following scale-dependent formula
\[\widehat{\dim}_{\mathrm{cap}}(r_1, r_2) = \lim_{r \rightarrow 0 } \frac{\log(M(r_2)) - \log(M(r_1))}{\log(r_2) - \log(r_1)}.\]
Because finding the exact value of $M(r)$ is an NP-hard problem, a greedy estimate of this value is used.

\begin{figure}
    \centering
    \includegraphics[width=0.9\linewidth]{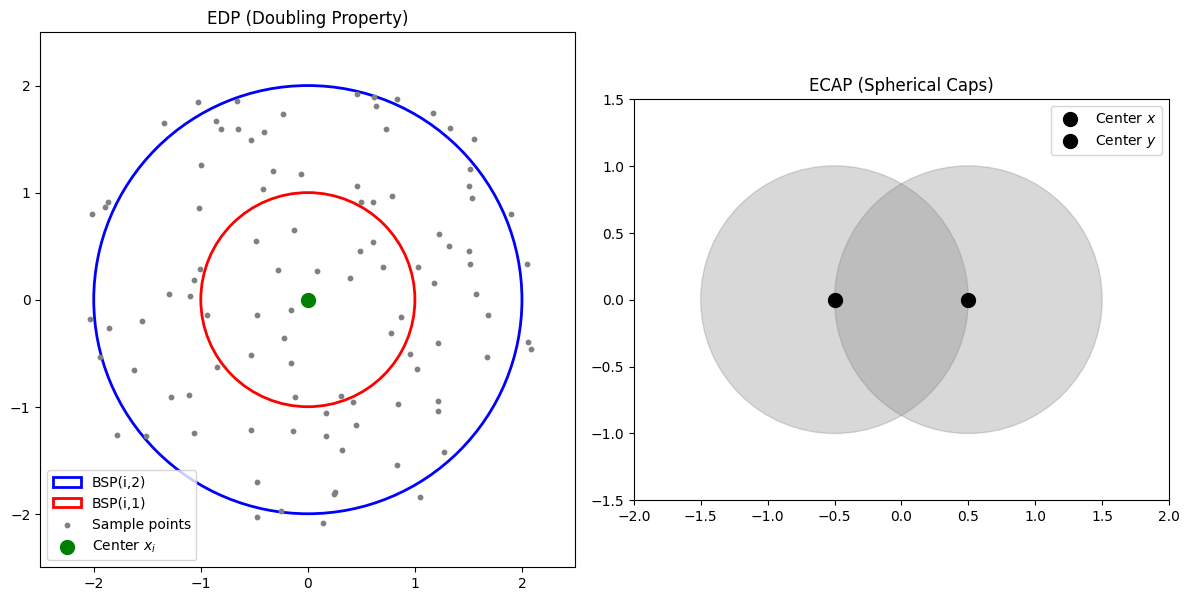}
    \caption{On the left the idea behind the doubling property estimator is represented. The ratio of the number of points in the two balls grows with dimension. On the right, \wodcap counts the number of points in the intersection of the two balls. \wodcap stands somewhat alone as being the only bi-local estimator that we have found. }
    \label{fig:wodexp}
\end{figure}

\paragraph{Volumes through graphs}
The methods of Kleindessner and von Luxburg are designed to provide an estimate without using distances; instead considering only the \knn graph~\cite{Kleindessner2015DimensionalityDistances}.
Since the ratio of the Lebesgue measure of the ball of radius 1 centred at $x\in \mathbb{X}$ to the Lebesgue measure of the ball of radius 2 is an injective function of dimension, a dimension estimate $\hat d_{\mathrm{DP}}(x)$ can be obtained from the ratio.
Similarly, the volume of the intersection of two balls of radius 1 with different centres can be compared to the unit ball, giving the ratio
\[S(d) = I_{\frac34}\left(\frac{d+1}{2}, \frac12\right),\]
where $I$ is the regularized incomplete beta function.
Inverting this function gives another estimate. The idea for both methods can be seen in \Cref{fig:wodexp}

To use these ratios to estimate dimension, construct the $\knn$ graph where points are connected by a directed unweighted edge from $i$ to $j$ if and only if $x_{j} \in \knn(x_{i})$. 
Using the edge length metric $d_{\mathrm{e}}$, the ball of radius $r$ is defined by $B_{\mathrm{e}}(i,r)= \{ j \in V | d_{\mathrm{e}}(i,j)<r \}$ where $d_{\mathrm{e}}(i,j)$ is the distance from $i$ to $j$ in the $\knn$ graph. 
They then consider the ratio of $|B_{\mathrm{e}}(i,1)| $ and $|B_{\mathrm{e}}(i,2)| $ and then average these ratios over a subset of the points to recover a dimension estimate. 
A similar construction can be used for the intersection of two unit balls with neighbouring centres, giving the \wodcap estimator $\hat d_{\mathrm{CAP}}$.

It has been shown that the sample version converges in probability to the theoretical construction as $n \rightarrow \infty$ in the special case where $M^d \subset \bbr^d$ is a compact domain in Euclidean space such that the boundary satisfies certain regularity assumptions and the density of $M$ is bounded both above and below.

\begin{theorem}\cite{Kleindessner2015DimensionalityDistances}
    Let $ \mathbb{X} =\{x_1 , \dots , x_n\} \subset M^d \subset \bbr^d $ be an i.i.d. sample from $f$ and let $G$
be the directed, unweighted kNN-graph on $\mathbb{X}$. Given
$G$ as input and a vertex $i \in \{1, \dots , n\} $ chosen uniformly at random, both $\hat d_{\mathrm{DP}}(x_i)$ and $\hat d_{\mathrm{CAP}}(x_i)$ converge to the true dimension $d$ in probability as $n \rightarrow \infty$
if $k = k(n)$ satisfies $k \in o(n)$, $\log (n) \in o(k)$, and there exists $k' = k'(n)$ with $k' \in o(k)$ and $\log (n) \in o(k')$.
\end{theorem}

These estimators are perhaps the most sensitive to choice of neighbourhood size, as discussed later in Section \ref{sec:NbhdSize}.
The first method tends to vastly underestimate the dimension, especially with high dimensional datasets. 
\wodcap performs very closely to Levina and Bickel's \mle~\cite{Levina2004MaximumDimension}\cite{Kleindessner2015DimensionalityDistances}. 

We refer the reader to~\cite{Serra2017DimensionModels, Qiu2021IntrinsicInformation, Qiu2023UnderestimationEstimation} which further develops these approaches.

\subsubsection{Nearest neighbour distances}

\paragraph{Maximum likelihood estimators}
The earliest estimator of this type is due to Trunk~\cite{Trunk1968StatisticalCollections}, who used a maximum likelihood estimator based on the joint distribution of both the ratios $\rho_{i,k}(p)$ and the angles $\theta_i(p)$ between the $(i+1)$th nearest neighbour and the span of the first $i$ nearest neighbours. For large $N$, the distribution of the distance ratios can be obtained from the distribution in the Euclidean setting, but the distribution of angles is less tractable, and an approximation was used.

Levina and Bickel applied the theory of maximum likelihood estimation to the point process to obtain an estimator calculated from nearest neighbour distances~\cite{Levina2004MaximumDimension}.
For $r$ small and $f$ almost constant near $p$, and approximating the binomial point process from which the $N$ points are an observation by a Poisson point process, the number of points within a distance $r$ of $p \in M$ has rate 
\[\lambda(r) = f(p) V_d d r^{d-1},\]
where $V_d$ is the volume of a unit ball in $\bbr^d$.

For a given $p \in \bbx$, this gives rise to a maximum likelihood estimate for $d$, which depends on how the neighbourhood of $p$ is selected.
Considering points in $\bpeps$, the maximum likelihood estimator for $d$ is 
\[ \hat{d}_{\varepsilon}(p) = 
\left( 
\frac{1}{N(p,\varepsilon)}
\sum_{j=1}^{N(p,\varepsilon)} \log \frac{\varepsilon}{r_j(p)}
\right)^{-1},\]
where $N(p,\varepsilon)$ is the number of points in $\bpeps$.
Similarly, considering points in $\knn(p;k)$, we obtain
\[ \hat{d}_{k}(p) = 
\left( 
\frac{1}{k-1}
\sum_{j=1}^{k-1} \left(\log \rho_{k,j}(p)\right)
\right)^{-1}.\]

Normalising by $k-2$ instead will make this estimator asymptotically unbiased (under the conditions $n, k \to \infty$ and $\frac k n \to 0$) and give variance $\frac{d^2}{k-3}$~\cite{Levina2004MaximumDimension}.

To obtain a global estimate, Levina and Bickel used the arithmetic mean of these local estimates over all samples. 
However, as noted by MacKay and Ghahramani~\cite{MacKay2005Comments2004}, making the approximate assumption that the rates at each point are independent allows one to write down a global likelihood function. 
The global maximum likelihood estimator for the dimension is not the arithmetic mean of the local estimates, but the harmonic mean.
This adjustment provides significant improvements at small $k$.
Levina and Bickel argue that the spatial dependence does not prevent the variance of the global estimator from behaving as $N^{-1}$, since for fixed $k$ this results in $N/k$ roughly independent groups of points.

A Bayesian approach to this problem is described in~\cite{Joukhadar2025BayesianDimensionality}.

Since the natural sizes of data clusters are typically unknown, it is preferable if the \mle approach can operate on minimal neighbourhoods, so that the majority of data points within a neighbourhood genuinely originate from the same local distribution. The \tle~\cite{Amsaleg2019IntrinsicLocalities} method is an extension of the standard \mle estimator specifically engineered for these tight neighbourhood scenarios.

Restricting neighbourhood size inherently limits the data available for estimation. To accurately estimate local dimensionality, it is advantageous to utilize all pairwise distances among neighbours rather than relying solely on distances from a central reference point. Denote the point at which the dimension estimation is carried out by $x \in \bbx$. The simplest application of this approach would involve computing the \mle for selected points within the neighbourhood of $x$ and averaging their results. However, simply aggregating estimators presents two primary challenges: it can violate the locality condition, by adding additional neighbours, or it can introduce a clipping bias, by restricting the data to the original neighbourhood. \tle seeks to address these challenges.

In its basic variant, local dimension estimation is performed at a selected point $x$ in a neighbourhood $V \subset \bbx$ centred at $q$, as well as at reflections of these points $x$ through $q$.
At each of these points, a dimension estimate is made using only elements from neighbourhood of point $q$, thereby preserving locality. To avoid clipping bias, the distances used are skewed to $d_{q,r}(x,v) = \frac{r(v-x) \cdot (v-x) }{2(q-x) \cdot (v-x)}$. Finally, these estimates are aggregated by harmonic mean to yield an estimator of the following form:

\[
\left(\widehat\dim_{\mathrm{TLE}}(q)\right)^{-1} = - \frac{1}{2 |V| (|V|-1)}\sum_{\substack{x,v \in V \\ x\neq v}} \left[ \ln\frac{d_{q,r}(x,v)}{r} + \ln\frac{d_{x,r}(2q-x,v)}{r} \right].
\]

This approach relies on the fundamental assumption that local intrinsic dimensionality exhibits uniform continuity across the data, given its use of dimension estimates for points within the neighbourhood.

The \mindml~\cite{Rozza2012NovelEstimators,Lombardi2011MinimumDimension} estimator, along with its derivatives \mindkl and \idea, which are discussed later, also builds on the work of Levina and Bickel while hewing closely to likelihood principles. The authors study the ratio $\rho_{1,k+1}(p)$ and calculate a maximum likelihood estimator as the root
\[ \hat{d} = \left\{ d \colon \frac{N}{d} 
    + \sum_{p \in \bbx} \left( \log \rho(p) - (k-1) \frac{\rho^d(p)\log\rho(p)}{1-\rho^d(p)} \right) = 0 \right\}.\]
In case $k=1$, this reduces to the harmonic mean of the Levina--Bickel estimator as recommended by MacKay and Ghahramani.

Another maximum likelihood based estimator, \gride~\cite{Denti2022TheEstimator}, uses only two neighbours. Unlike \twonn, which will be described later, these need not be the two nearest neighbours.
The intention is that the neighbours be sufficiently far away to overcome problems caused by noise.
For two parameters $n_1 < n_2$, the ratio $\rho_{n_2, n_1}$ is studied. Note that the \twonn method uses only $\rho_{2,1}$.
The distribution of these ratios yields a concave log-likelihood function which in general can be maximised by numerical optimization.
However, where $n_2 = n_1 + 1$ a closed form solution exists, which reduces to the standard MLE estimator for $k=2$ in case $n_1 = 1$.
In these cases, the authors note that if the data are generated from a Poisson process then the the ratios follow the Pareto distribution. Furthermore, the ratios $\rho_{n_1+1, n_1}(p)$ for each $n_1$ are jointly independent.
This allows for uncertainty quantification and, if data over a range of $n_1$ are considered, a reduction in the variance of the estimator.

A method for adaptively choosing the neighbourhood size to carry out maximum likelihood estimation is described in~\cite{DiNoia2024BeyondIdentification} while Amsaleg \etal demonstrate an approach using methods from extreme value theory in~\cite{Amsaleg2018Extreme-value-theoreticDimensionality}.

\paragraph{Fitting distributions}

Building on the premises of Levina and Bickel's work, Facco \etal~\cite{Facco2017EstimatingInformation} developed the \twonn method. This uses only distance information to the two nearest neighbours of each point. 
By using only the two smallest distances, the neighbourhoods involved can be assumed to be closer to Euclidean space and the probability distribution need only be approximately constant on the scale $r_2(p)$.
The ratio $\rho_{2,1}(p)$ is the local statistic of interest here.
Assuming that $\rho_{2,1}$ is independently identically distributed at each $p$, and letting $F$ be the cumulative distribution function (CDF)  of the distribution,
we have 
\[\frac{\log(1 - F(\rho_{2,1}))}{\log(\rho_{2,1})} = d.\]
The observed ratios provide an empirical CDF which is used to estimate the intrinsic dimension by linear regression, fitting a line through the origin.
The authors recommend truncating this data before fitting the line, as the higher values of the CDF tend to be noisier.

Rozza \etal claim that, for large $d$, the value of $k$ needed for the data to closely match the theoretical distribution grows exponentially with $d$~\cite{Rozza2012NovelEstimators}.
The grounds for this claim appear weak, based on an assessment of the probability of a point lying near the centre of its neighbourhood ball when, in fact, that event is already conditioned on.
It is nevertheless true that for many datasets a large value of $n$ is needed to ensure that a \knn neighbourhood does not meet the boundary of the manifold, so that the methods of this paper are still relevant.

To address the possibility that the underlying assumptions of maximum likelihood estimators are far from being true, one proposal, \mindkl, is to compare the distances to neighbours not to the theoretical distribution, but to an empirical distribution obtained distances to neighbours in a random sample of $N$ points from a unit ball of each dimension. The dimension estimate is obtained by using the Kullback--Leibler divergence to select the dimension which most closely matches the data.
This method is not so well motivated for the possibility that $M$ is not geometrically similar to a unit ball.

\paragraph{Other distance-based approaches}

Another proposal from the same paper, \idea, uses that $d = \frac{m}{1-m}$ where $m$ is the expected norm of a vector sampled uniformly from the ball. Estimating $m$ by \[\hat{m} = \frac{1}{Nk} \sum_{i=1}^N \sum_{j=1}^k \rho_{j,k+1}(x_i)\] they use \[\hat d = \frac{\hat m}{1- \hat m }\] as a consistent estimator of $d$. 
The authors note an underestimation bias, which they attribute to small sample size, but which may also be due to the convexity of the function $\frac{x}{1-x}$ on $(0,1)$. 
The bias in the estimator is corrected using a jackknife technique: subsamples are generated where each point is included with probability $p$ and the number of nearest neighbours is reduced to $k \sqrt{p}$, so that increasing $p$ emulates a situation where $k, N \to \infty$ and $\frac{k}{N} \to 0$. The resulting dimension estimators are fitted to a curve \[\hat d = a_0 - \frac{a_1}{\log_2 (\frac{pN}{a_2} + a_3)}.\] The horizontal asymptote $a_0$ is used as the estimate (unless $a_1<0$, indicating that $\hat d$ in fact decreases with $p$; in this case the original estimate $\hat d$ for the entire dataset is preferred).

The earlier work of Pettis \etal~\cite{Pettis1979AnInformation} is also relevant here.
These authors directly calculated the distribution of $r_k(x)$ and from this obtained that the expected value of the mean distance from $x$ to $\knn(x)$ is $\frac{1}{G_{k,d}}k^{1/d}C_n$ where $G_{k,d} = \frac{k^{1/d}\Gamma(k)}{\Gamma(k+\frac1d)}$ and $C_n$, while sample dependent, is independent of $k$.
Taking logarithms yields a regression problem
\[\log \bar r_k -\log G_{k,d} =  \frac1d \log k  + \log C_n,\]
where the slope is $\frac1d$. On the left hand side of this equation, while $\bar r_k = \frac1k\sum_{j=1}^k r_j(x)$ is fixed, the value of $G$ depends explicitly on $d$.
An iterative regression method is used, where $\log G$ is initially taken to be 0, allowing the regression to be carried out, and the resulting value of $d$ is then used to recalculate $\log G$, until the scheme converges.
This was later refined further by Verveer and Duin~\cite{Verveer1995AnEstimators}. 

Other estimators in this general category include~\cite{Ivanov2021ManifoldEstimation}.

\subsubsection{Angles between nearest neighbours}

The \danco method~\cite{Ceruti2014DANCo:Concentration} uses information about angles between neighbours in addition to distances.

A maximum likelihood estimate from the distances is made as suggested in~\cite{Lombardi2011MinimumDimension} by solving the problem,
$$\hat{d}_{\mathrm{ML}} =\argmax_{1\leq d \leq D} \left\{ N \log(kd) + (d-1) \sum_{x_i \in \bbx} \log(\rho_{1,k+1}(x_i)) + (k-1)\sum_{x_i \in \bbx} \log(1- \rho_{1,k+1}^d(x_i))\right\}.  $$

Let the vector $\hat{\theta_i}$ be given by the angles subtended at $x_i$ by each pair of points in $\knn(x_i)$.
As each component of $\hat{\theta_i}$ follows a von~Mises--Fisher distribution with parameters $\nu$ and $\tau$, at each point $x_i$ the parameters can be estimated with a maximum likelihood approach.
This yields vectors $\hat{\bf{\nu}} =(\hat{\nu}_{i})_{i=1}^{N} $ and $\hat{\bf{\tau}} =  (\hat{\tau}_{i})_{i=1}^{N}$ with means $\bar{\nu}$ and $\bar{\tau}$. 

The statistics $\hat{d}_{\mathrm{ML}}$, $\bar\nu$ and $\bar\tau$ are then compared with the statistics from a family datasets with a known intrinsic dimension (in this case a uniform sample of $N$ points from $S^n$ for $1 \leq n \leq D$) to provide an estimate.


Another method based on angles is the Expected Simplex Skewness method, \ess. This method has two variants: \essa and \essb. We describe only \essa in detail.
Consider $\knn(p) \subset \bbx$ and a target dimension, $m$.
We may form $(m+1)$-simplices with one vertex at the centroid of $\knn(p)$ and the others at $m+1$ points drawn from $\knn(p)$.
A comparison simplex can be formed where at one vertex all edges meet orthogonally, and the side lengths of those edges are the same as the side lengths incident to the centroid.
For each simplex, a statistic referred to as ``simplex skewness'' is obtained as the ratio of the volume of the simplex to this comparison simplex.

For example, in the case of $m=1$ the area of a triangle with one vertex at the centroid is compared to the area of a right triangle with short side lengths equal to the lengths of the edges incident to the centroid.
The simplex skewness for this 2-simplex is simply $\sin(\theta)$, where $\theta$ is the angle at the centroid. 

For uniformly distributed points in a $d$-dimensional ball $B^d$ with volume measure $\mu$, the expected simplex skewness in case $m=1$ is $$ s^{(1)}_d = \frac{1}{V_d} \int_{B^d} |\sin(\theta(x))| \diff \mu(x)$$
where $V_d$ is the volume of the unit ball and $\theta(x)$ is the angle at the centre of the ball between $x$ and any fixed co-ordinate axis.

Similarly, for $m>1$, we obtain
\[
s^{(m)}_d = \frac{1}{V_d^m} \int_{(B^d)^m} \left|\frac{u \wedge v_1 \wedge \cdots \wedge v_m}{|v_1| \cdots |v_m|} \right| \, \diff \mu(v_1) \, \diff \mu(v_2) \cdots \diff \mu(v_m) = \frac{\Gamma(\frac{d}{2})^{m+1}}{\Gamma(\frac{d+1}{2})^{m} \Gamma(\frac{d-m}{2})}
\]
where $u$ is a unit vector along a chosen coordinate axis.

When simplex edges are short, the precise position of the centroid has a major influence on the skewness.
To mitigate this, the empirical estimate $\hat s^{(m)}$ is given by a weighted mean of the skewnesses of each simplex, where the weights are the products of the edge lengths incident to the centroid.
Since the $ s^{(m)}_d$ increase monotonically to $1$ as $d\rightarrow\infty$, by comparing the skewness of simplices in the data to the expected simplex skewness for balls of varying dimensions $d$, a dimension estimate can be obtained by linear interpolation~\cite{Johnsson2015LowSkewness}.

The alternative method \essb takes the unit vector in each edge direction and estimates the dimension from the projection on each edge to to the others, using a similar process.
A simplex-based approach was also used in~\cite{Cheng2009DimensionSlivers}.

\subsubsection{Other Parametric methods}

The \fishers method~\cite{Albergante2019EstimatingAnalysis} introduces an estimation technique that leverages the principle of linear Fisher separability. The underlying insight is that separability becomes more pronounced in higher-dimensional spaces. A key aspect of this method is its independence from the manifold hypothesis. It is also important to note that irregularities in data sampling can result in a reduction of the estimated intrinsic dimensionality. To estimate the dimension, the probability that a randomly selected element of the processed set is not separable is estimated. Then the obtained value is compared with the values for the equidistributions on a ball, a sphere, or the Gaussian distribution. The method of~\cite{Sutton2023RelativeLearning} also relies on this separability property.

Other estimators are based on distances in the graph at intermediate scale, rather than at the nearest neighbour scale~\cite{Granata2016AccurateDatasets} or on comparing the Wasserstein distance between samples of different size~\cite{Block2022IntrinsicDistance}.

\subsection{Estimators using topological and metric invariants}\label{sec:graph_estimators}

In this section, we discussion three related dimension estimators. The $k$-nearest neighbour estimator \knnestimator and the minimum spanning tree estimator -- a.k.a. zeroth-persistent homology estimator \phzero{} -- are both based on the theory of Euclidean functionals~\cite{Yukich1998ProbabilityProblems}. 
Both \phzero and $\mathsf{mag}$ are based on invariants of metric spaces. We refer the reader to~\cite{Govc2021PersistentMagnitude,OMalley2023AlphaMagnitude} about connections between persistent homology and magnitude. We also note a related concept to magnitude dimension which we do not cover here, the spread dimension estimator~\cite{Dunne2023MetricHypothesis}.

Due to their similarities, we devote \Cref{app:ph_vs_knn} to a comparison of the empirical performances of \phzero and \knn on the benchmark datasets which we describe in \Cref{ssec:datasets}. In \Cref{app:mag}, we consider the empirical performance of on spheres of different dimensions, which demonstrate limitations of $\mathsf{mag}$ in estimating dimension from finitely many point samples. Here we outline the theoretical foundations of such methods.

\subsubsection{MST and PH dimension}    
Given a finite point set $\bbx \subset X$ in a metric space, the minimum spanning tree $\mst(\bbx)$ of $\bbx$ is a tree with $\bbx$ as vertex set, such that the sum of $\sum_{(x_i, x_j) \in \mst(\bbx)} d(x_i, x_j)$ over edges $E$ of the tree is minimal. For $\alpha > 0$, we define the $\alpha$-weight of $\bbx$ to be
\begin{equation}
    E_\alpha(\bbx) := \sum_{(x_i, x_j) \in \mst(\bbx)} d(x_i, x_j)^\alpha
\end{equation}
Given the relationship between zeroth-dimension persistent homology and minimum spanning trees, it is natural to generalise this notion to higher dimension homological features captured by $\bbx$. We let $\PHDim_i(\bbx)$ denote the \emph{finite} part of the persistence diagram of the $i$-th Vietoris-Rips or \v{C}ech persistent homology of $\bbx$.  We let the $\alpha$-\emph{total persistence} of the persistence diagram be
\begin{equation}
    E^i_\alpha(\bbx) = \sum_{(b,d) \in \PHDim_i(\bbx)} (d-b)^\alpha
\end{equation}
In particular if we consider dimension $i = 0$, we have $E_\alpha(\bbx) := E^0_\alpha(\bbx) $.
Some relationships also hold when $i>0$; see~\cite{Schweinhart2020FractalComplexes,Schweinhart2021PersistentDimension} for details.

In~\cite{Costa2004GeodesicLearning,Birdal2021IntrinsicNetworks,Adams2020AHomology}, it is proposed that the intrinsic dimension of data $\bbx$ be estimated by taking various subsamples $\bbx' \subset \bbx$, and computing the slope $m$ of the curve  $\log (E^0_\alpha(\bbx'))$ as a function of $\log(|\bbx'|)$. The inferred dimension is then given by
\begin{equation} \label{eq:phdim_slope}
    \hat{d} = \frac{\alpha}{1-m}.
\end{equation}
There is considerable literature on the study of minimum spanning trees that justifies this estimator. In the study of minimum spanning trees on point sets in $[0,1]^d$, it has been a folklore theorem that the growth rate of $E_\alpha$ with the number of points is at most $O(n^{\max(1-\alpha/d,0)})$ for $\alpha \in (0,d)$~\cite{Kozma2005TheDimension}. In other words, for higher dimensional cubes $d > \alpha$, the $\alpha$-weight $E_\alpha$ can grow sublinearly with $n$, while for lower dimensional cubes $ d \leq \alpha$, the $\alpha$-weight $E_\alpha$ does not grow with $n$ at all. This motivated the definition of the \emph{minimum spanning tree dimension} of a bounded metric space $X$ in~\cite{Kozma2005TheDimension} as 
\begin{equation} \label{eq:mstdim_kozma}
    \mstdim(X) := \inf \{\alpha > 0 \ : \ E_\alpha(\bbx) \text{ finite } \ \forall \text{ finite subsets } \bbx \subset X\}.
\end{equation}
Subsequently, the \emph{persistent homology dimension} was analogously defined~\cite{Birdal2021IntrinsicNetworks} to be 
\begin{equation} \label{eq:phdim_birdal}
    \phdim^i(X) := \inf \{\alpha > 0 \ : \ E_\alpha^i(\bbx) \text{ finite } \ \forall \text{ finite subsets } \bbx \subset X\}.
\end{equation}
By definition it follows that $\mstdim(X)  = \phdim^0(X) $. In Theorem 2 of~\cite{Kozma2005TheDimension}, it was shown that for any bounded metric space $X$, the minimum spanning tree dimension recovers the upper box counting dimension
\begin{equation}
    \mstdim(X) = \boxdim(X).
\end{equation}
Abstractly defined, neither~\cref{eq:mstdim_kozma,eq:phdim_birdal} are amenable to computation. Adams et al.~\cite{Adams2020AHomology}  obtained a lower bound on $ \mstdim(X)$ based on results of Schweinhart~\cite{Schweinhart2020FractalComplexes} by extending the $O(n^{1-\alpha/d})$ bound on the growth of $E_\alpha$ to any bounded metric space $X$. For $\alpha \in (0,  \mstdim(X))$, and any $\{x_1, \ldots, x_n\} \subset X$, there is some constant $C_{\alpha,d}$ such that 
\begin{equation} \label{eq:birdal}
\log E_\alpha(\{x_1, \ldots, x_n\})  \leq 
     \left(1 - \frac{\alpha}{ \mstdim(X)} \right) \log n + C_{\alpha,d}
\end{equation}
This implies for any sequence $(x_i)$ of distinct elements of $X$, we can obtain a lower bound on the upper box dimension and minimum spanning tree dimension by
\begin{equation} \label{eq:birdal_lb}
    \mstdim(X) \geq \frac{\gamma}{1-\beta} \quad \text{where} \quad \beta = \limsup_{n \to \infty} \frac{\log(E_\gamma(\{x_1, \ldots,x_n \})}{\log n}.
\end{equation}
The probabilistic properties of the lower bound $\gamma/(1-\beta)$ in~\cref{eq:birdal_lb} have also been widely studied. Given a probability measure $\mu$ with bounded support on a metric space $X$,~\cite{Schweinhart2020FractalComplexes} defines the persistent homology dimension of the metric measure space $(X, \mu)$ as
\begin{equation}
\phdim^i(\mu; \gamma) = \frac{\gamma}{1-\beta} \quad \text{where} \quad \beta = \limsup_{n \to \infty} \frac{\log \mathbb{E}(E_\gamma^i(\{x_1, \ldots,x_n \})}{\log n},
\end{equation}
given a parameter $\gamma > 0$. We recall the classic result for probability measures on Euclidean space by Beardwood, Halton and Hammersley~\cite{Beardwood1959ThePoints}, and Steele~\cite{Steele1988GrowthEdges}, which implies $\phdim^0(\mu; \gamma) = d$ for $\mu$ an absolutely continuous measure on $\bbr^d$, and $\gamma \in (0,d)$. 
\begin{theorem} \label{thm:steele_mst}
    Let $\mu$ be a probability measure on $\bbr^d$ with compact support, and $f$ be the density of its absolutely continuous part.  If $x_1, \ldots, x_n$ are i.i.d. samples from $\mu$,  then with probability 1, 
    \begin{equation} \label{eq:steele}
        \lim_{n \to \infty}     \frac{E_\alpha(\{x_1, \ldots, x_n\}) }{n^{1 - \alpha/d}}\to c(\alpha, d)\int_{\bbr^d} f(x)^{1-\alpha/d}\ \diff x,
    \end{equation}
    for $\alpha \in (0,d)$. Here $c(\alpha, d) > 0$ is a constant that only depends on $\alpha, d$. 
\end{theorem} 
Refinements of~\Cref{thm:steele_mst} have been proved for special cases. If $\mu$ is the uniform distribution on the $d$-dimensional unit cube,  Kesten and Lee~\cite{Kesten1996THEPOINTS} proved a central limit theorem for ${E_\alpha}$: for $\alpha > 0$, we have a convergence in distribution
\begin{equation}
    \sqrt{n}\left(    \frac{E_\alpha(\{x_1, \ldots, x_n\}) }{n^{1 - \alpha/d}} - \mu\right)\to N(0, \sigma_{\alpha,d}^2).
\end{equation}
Costa and Hero~\cite{Costa2015LearningDatasets} generalise~\Cref{thm:steele_mst} from $\rd$ to compact Riemannian manifolds of dimension $d$, with $f$ being a bounded density relative to the density of the manifold. In~\cite{Costa2004GeodesicLearning} they also observe that the limit right hand side of~\cref{eq:steele} is in fact related to R\'enyi entropy. 

Recently,~\cite{Schweinhart2020FractalComplexes} proved a generalisation of \Cref{thm:steele_mst} for $d$-Ahlfors regular measures on a metric space. A measure $\mu$ is \emph{$d$-Ahlfors regular} if there is are real, positive constants $c,\delta_0 > 0$ such that for all $\delta\in (0, \delta_0)$, and $x \in X$, the measure of the open ball of radius $\delta$ at $x$ is bounded by 
\begin{equation}
    \frac{1}{c} \delta^d \leq \mu(B_\delta(x)) \leq c \delta^d.
\end{equation}
This condition is satisfied, for example, for uniform measures on a $d$-dimensional manifold. Theorem 3 of~\cite{Schweinhart2020FractalComplexes} implies for $\gamma \in (0,d)$, the persistent homology dimension of a $d$-Ahlfors regular measure recovers the dimension parameter $d$:
\begin{equation*}
   \phdim^0(\mu; \gamma) = d.\end{equation*}

\begin{remark} We note that other aspects of the minimum spanning tree are related to dimension: for samples from a $d$-Ahlfors measure,~\cite{Kozma2010OnSpaces} showed the the maximum distance along an edge of its minimum spanning tree scales as $((\log n)/n)^{1/d}$.
\end{remark}
\begin{remark}
    For data sampled from a submanifold,~\cite{Costa2004GeodesicLearning} proposed using an Isomap based geodesic distance distance estimation for the construction of the minimum spanning tree. 

In practice, since constructing a minimum spanning tree on $n$ points with full metric data is computationally onerous, a simplification can be made by taking the $k$-nearest neighbour graph and computing the minimum spanning tree of the \knn graph instead to speed up the computation. 
\end{remark}

\subsubsection{The \knn graph}
The $k$-nearest neighbour graph $G_k(\bbx)$ on a finite subset $\bbx$ of a metric space has for every point in $\bbx$ an edge between itself and its $q$\textsuperscript{th} nearest neighbour in $\bbx$ for $q \leq k$.
\begin{equation}
    L_\alpha^k(\bbx) = \sum_{(x,y) \in G_k(\bbx)} d(x,y)^\alpha
\end{equation}
be the total edge length of the \knn graph. Like the total weight of its minimum spanning tree $E^\alpha$, the total edge length of the \knn graph $L_1^k$ satisfies suitable additive properties such that the Euclidean functional theory as described in~\cite{Yukich1998ProbabilityProblems} applies. Thus, a similar asymptotic result can be derived~\cite[Theorem 8.3]{Yukich1998ProbabilityProblems}:
\begin{theorem} \label{thm:yukich}
    Let $\mu$ be a probability measure on $\bbr^d$ with compact support where $d \geq 2$, and $f$ be the density of its absolutely continuous part.  If $x_1, \ldots, x_n$ are i.i.d. samples from $\mu$,  then
    \begin{equation} \label{eq:steele}
        \lim_{n \to \infty}     \frac{L_1^k(\{x_1, \ldots, x_n\}) }{n^{1 - 1/d}}\xrightarrow{P} c(k, d)\int_{\bbr^d} f(x)^{1-1/d}\ \diff x,
    \end{equation}
  Here $c(k, d)$ is a constant that only depends on $k, d$. 
\end{theorem} 
Given this theoretical characterisation~\cite{Costa2003EntropicLearning} proposes an estimator \knnestimator using a similar strategy to the estimation of $\mstdim$ and $\phdim^0$, as expressed in \cref{eq:phdim_slope}. Given data $
\bbx$, we compute the total length $L^k_1(\bbx')$ of $k$-nearest neighbours graphs of subsamples $\bbx' \subset \bbx$, and infer the slope $m$ of the curve $\log(L^k_1(\bbx'))$ as a function of $\log(|\bbx'|)$:
\begin{equation} \label{eq:knn_slope}
    \hat{d} = \frac{1}{1-m}.
\end{equation}
\begin{remark}[Error propagation] \label{rmk:ph_knn_error}
    Both $\hat{d}_{\knn}$ and $\hat{d}_{\mathrm{PH}}$ are subject to instability when the intrinsic dimension is high. For high dimensions, the regressed slope $m$ is just under one $m = 1-1/d$; any small error $\epsilon$ in the slope can cause a large change in the estimated dimension
    \begin{equation*}
        m \mapsto m + \epsilon \implies \hat{d} \mapsto \frac{d}{1 - \epsilon d}
    \end{equation*}
    The dimension estimate will be impacted if $|\epsilon| > 1/d$; for datasets with high intrinsic dimension the error tolerance in the inference of the slope becomes smaller. In particular if the slope is over-estimated $\epsilon d  \gtrsim 1$,  the dimension estimate can even become very negative. This is reflected in the performance of $\knnestimator$ on M10d Cubic with intrinsic dimension 70 (see \Cref{app:ph_vs_knn}).  
\end{remark}
\begin{remark}[Robustness to outliers] \label{rmk:ph_knn_outlier} Given a probability measure on an embedded manifold, we can regard the absolutely continuous part of the measure (w.r.t. the density of the underlying manifold) as the `signal', while the singular part can be regarded as a model for outliers. While the minimum spanning tree and k-nearest neighbour graphs are individually sensitive to outliers in data, \Cref{thm:steele_mst,thm:yukich} suggests that the scaling of the total edge lengths of such objects with increasing sample size should be robust against outliers, as the scaling should only depend on the absolutely continuous part of the measure. Nonetheless, in practice, if the measure of the absolutely part is small (i.e. the number of outliers dominate the sample), then it may take a large number of samples for the asymptotic scaling behaviour of $E_\alpha$ and $L^k$ to emerge. 
\end{remark}
\subsubsection{Magnitude dimension}
The \emph{magnitude} of a metric space is a way of measuring the ``effective cardinality' of the space~\cite{Leinster2013OnSpace,Meckes2013PositiveSpaces,Meckes2015MagnitudeSpaces,OMalley2023AlphaMagnitude}. For simplicity of exposition, we only consider compact subsets of Euclidean space here and refer the reader to~\cite[\S 3]{Meckes2015MagnitudeSpaces} for a general definition of magnitude for compact metric spaces. For any finite metric space $X = \{x_1, \ldots, x_n\}$, its magnitude is defined to be
\begin{equation*}
    |X| = \sum_{ij} ({Z^{-1}})_{ij},\quad \text{where}\quad Z_{ij} = e^{-d(x_i,x_j)}.
\end{equation*}
If $X$ is a compact subset of $\bbr^n$, then its magnitude is given by (see~\cite[\S 1.3]{Leinster2013OnSpace})
\begin{equation}
    |X| = \sup\{|\bbx|,\ \bbx \subseteq X \text{ finite} \}.
\end{equation}
If a sequence of finite subsets $\bbx_k \subset \bbr^n$ converges fo a compact subset $X$ of $\bbr^n$ in Hausdorff distance, then $|\bbx_k| \to |X|$~\cite[Corollary 2.7]{Meckes2013PositiveSpaces}. We note that in the general case, the approximation of magnitude with finite subsets, and the stability of magnitude with respect to Gromov--Hausdorff distance between metric spaces is a nuanced topic: see~\cite{Katsumasa2025IsSpaces}.

The magnitude dimension of a compact subset of $\bbr^n$ is derived from how magnitude behaves as the scale of the space is varied. We let  $tX$ denote the metric space with metric rescaled as $td(x,y)$ for $t>0$. The \emph{magnitude function} is then the map $t \mapsto |tX|$, which is continuous for $X$ a compact subset of $\bbr^n$~\cite[Corollary 5.5]{Meckes2015MagnitudeSpaces}. The magnitude function captures the effective cardinality of a metric space at different scales and its limiting behaviour at large scales recovers the dimension. For $X$ a compact subset of $\bbr^n$, its \emph{upper magnitude dimension} is defined to be 
\begin{equation}
\overline{\magdim}(X) = \limsup_{t \to \infty} \frac{\log|tX|}{\log t}.
\end{equation}
The lower magnitude dimension $\underline{\magdim}(X)$ is similarly defined where the $\limsup$ is replaced by $\liminf$. For $X$ a compact subset of $\bbr^n$,~\cite[Corollary 7.4]{Meckes2015MagnitudeSpaces} states that the upper (lower) magnitude dimension coincides with the upper (lower) Minkowski dimension. When the limit exists and the lower and upper magnitude dimensions coincide, we use that as the definition of the magnitude dimension $\magdim(X)$.

Given finite samples $\bbx \subset X$ of a compact subset of $\bbr^n$, it was proposed in~\cite{Willerton2009HeuristicSpaces,Andreeva2023MetricNetworks} that we consider the magnitude function of $\bbx$ and use the slope of the line of best fit to $\log t$ against $\log|t\mathbb{X}|$ for $t \gg 0$ as a heuristic estimation of $\magdim(X)$.

\subsubsection{Analysis}
In our experiments, \phzero dimension performs comparably with other estimators, yet is susceptible to over-estimation in the presence of noise (see discussion in \Cref{sec:noise}). However, magnitude dimension requires a lot of points for even spheres of moderate dimension (see \Cref{tab:mag}), making it an unreliable estimator.

\subsection{Other estimators}
Other approaches have been developed but are not considered here. One common theme has been the use of diffusion geometry, normalising flows, and convolution with Gaussians to recover information about the manifold. Work in this area includes~\cite{Horvat2022IntrinsicFlows, Jones2024ManifoldDimension, Kamkari2024AModels, Stanczuk2024DiffusionManifolds, Tempczyk2022LIDL:Likelihood, Yeats2023AdversarialMaps}.

One can also consider the trade-off between reducing dimension and minimising metric distortion of dimension reduction methods such as~\cite{Balasubramanian2002TheStability.,VanDerMaaten2008VisualizingT-SNE} to estimate dimension. 

A quantum cognition machine learning approach was used in~\cite{Candelori2025RobustLearning}, where excellent performance against noisy datasets is reported.

\subsection{Local and global estimators}
\begin{table}[h]
\centering 
\begin{tabular}{|l|l|}\hline 
       & Estimators                                                                                                                                 \\ \hline 
Local  & \lpca, \mle, \wodcap, \ess                                                     \\\hline  
Global & \phzero, $\mathsf{mag}$, \knnestimator, \gride, \twonn, \danco, \mindml, \corrint \\ \hline 
\end{tabular}
\caption{A classification of estimators considered in our experiments into global and local estimators. }
\end{table}
An alternative categorisation of estimators divides them into local and global methods, with local estimators providing local dimension estimates which are then combined, while global estimates provide a single value.

In truth, any reliable estimator for the dimension of a manifold is likely to first obtain local information of some sort, aggregate this information, and then convert it into a dimension estimator. The estimators usually known as ``local estimators'' are, from this point of view, those for which the local information is already a dimension estimate.

In this section, we review some of the estimators already discussed above which might be described as ``global estimators''.

The method of Fan \etal~\cite{Fan2010IntrinsicAnalysis} identifies different eigenvalues for each neighbourhood from $\lpca$. These local eigenvalues are then combined before the thresholding method is applied to the combined data. In this way, despite the method being a variant of ``local \pca'', it does not directly produce local dimensions, and so is strictly speaking a global method. The architecture of the \scikitdimension package does not easily allow for implementation of this method and so the results as presented do not reflect the original method as designed by Fan \etal. Instead the thresholding method is applied locally, converting it into a local method. Note that any \lpca method could be adapted to function in the same way, with thresholding being applied to these ``global eigenvalues''.

\corrint operates in a similar spirit. A simple local estimate of dimension at a point $p \in \bbx$ is given by $\frac{\log N(p, r_2) - \log N(p,r_1)}{\log r_2 - \log r_1}$ for two radii $r_1 < r_2$, under the assumption that $N(p,r)$ is proportionate to the volume of the ball of radius $r$, and that this grows as $r^d$. However, for \corrint, rather than combining these local values of $d$, the volumes are combined over all balls, and a single estimate of $d$ is then calculated.

\twonn considers the entire distribution of local statistics to obtain a global value. At each point, the ratios $\rho_{2,1}$ are considered. A local dimension estimate could be obtained from this ratio, but instead the empirical distribution of these ratios across all $p \in \bbx$ is considered and compared to the theoretical distribution, which is dependent on $d$. This approach seems to result in a relatively high standard deviation compared to other estimators.

The estimators described in \Cref{sec:graph_estimators} might appear to be global in that a global geometric object, usually a graph of some kind, is constructed, and then a single metric invariant is extracted from it. However, even here, the analysis of the estimators to demonstrate their convergence turns out to rely on an understanding of how they behave locally, which can then be used to infer the global behaviour via some additivity condition.

Truly global estimators generally assume some global structure in the data. For example, a direct application of \pca can be used to estimate dimension, with the assumption here being that the data lies in an affine subspace. The \mindkl estimators are also truly global, making the global assumption that the distribution of distances behaves similarly to that in a ball.

The \wodcap method stands somewhat alone, in that it is essentially a bilocal estimator. Two points are used as the centres of intersecting balls, and the resulting estimate of dimension is an estimate at this \emph{pair of points}.

\subsection{Underestimation}\label{sec:underestimation}

Underestimation of dimension has been widely reported, especially for higher-dimensional datasets, and is observed again in this survey.
This is commonly attributed two possible causes: the ``boundary effect' or ``edge effect'', with observations of this dating back at least to~\cite{Pettis1979AnInformation}, and a shortage of samples. 

However, caution is warranted when considering how well this observation generalises beyond the benchmarking set. We find that, when using datasets sampled from $SO(n)$, overestimation is common.

\subsubsection{Boundary effects}
For points lying near the boundary of a manifold, many of the nearest neighbours are, in a sense, missing.
As a result, the distribution of nearest neighbour distance is far from that which holds for points in the interior.

For a $d$-dimensional ball, or $d$-dimensional cube, as $d$ grows the proportion of the volume contained in a small neighbourhood of the boundary approaches $1$. For example, in a unit ball, for any $0 < r < 1$, a proportion $r^d$ of the volume is within a distance $r$ of the centre, and as $d \to \infty$ this converges to 0. As a result, many points in the sample will lie near the boundary and not satisfy the hypotheses of the estimator.

From the point of view of probability measures, the uniform measure on the ball or cube is therefore not very different from a distribution on the boundary -- the Wasserstein 2-distance between a ball and a sphere is given by
\[ \left( \int_0^1 (1-r)^2 \frac{2\pi^{d/2}}{\Gamma(d/2)} r^{d-1} \diff r \right)^{1/2} = \left(\frac{4\pi^{d/2}}{\Gamma(d/2)d(d+1)(d+2)}\right)^{1/2}\] which converges to 0 exponentially fast as $d \to \infty$.
As a result, even the highest quality dimension estimator will have a tendency to underestimate the dimensions of balls and cubes.

In fact, the proportion of the volume of a sphere contained near an equator also approaches $1$ as $d$ grows, and so spheres of different dimensions are also difficult to distinguish~\cite{Ledoux2001ThePhenomenon}

For this reason, pure probabilistic approaches to dimension estimation on balls and spheres will generally end in underestimation.
The assumption that underestimation is inevitable is a generalisation from this observation to a belief that the negative bias induced by the presence of boundary will always grow with dimension.
However, this phenomenon whereby measure is concentrated near the boundary is not one which holds universally. It depends on the family of manifolds being considered.
For example, the $d$-dimensional manifolds given by $S^{d-1} \times [-1, 1]$ will have a proportion $\delta$ of the volume lying within $\delta$ of the boundary. This value is completely independent of $d$.
\begin{remark}
    The observations here restrict the performance of local dimension estimators for datasets with high intrinsic dimension. 
    Local dimension estimators perform dimension estimation on local neighbourhoods of points $p$ in the dataset, often modelled by intersecting the datasets with an $\epsilon$-neighbourhood ball (equivalently a $k$-nearest neighbourhood can be expressed as an $\epsilon$-ball with $\epsilon$ being the distance to the $k$-nearest neighbour). Assuming that the local neighbourhood is sufficiently small,  the data that a local estimator sees is drawn from an approximately uniform distribution on a ball. 

    As the reference point is conditioned to lie at the centre of this ball, the impact may not be so severe. However, the concentration phenomenon will still make it difficult to distinguish distributions of distances, as these are concentrated around the radius of the ball.
\end{remark}
\subsubsection{Sample size effects}

Another claimed source of negative bias is the insufficiency of data.
For example, it has been claimed that the original version of the \corrint estimator~\cite{Grassberger1983MeasuringAttractors} can only provide estimates $\hat d < \frac{2 \log N}{\log{\diam(\bbx)-\log \varepsilon}}$, where $\varepsilon \ll \diam (\bbx)$ is the smallest radius considered~\cite{Eckmann1992FundamentalSystems}. However, this calculation is based on an assumption that the volume of balls grows as $r^d$ for all values of $r$ up to $\diam (\bbx)$, which does not generally hold.

Assuming that a given estimator is asymptotically correct, it may be possible to use the given sample to estimate the value it would take on an arbitrarily large sample. The \idea estimator~\cite{Rozza2011IDEA:Algorithm} attempts this by using a jackknife subsampling method and fitting the estimates for a subsamples to a curve with a horizontal asymptote. 

For small sample sizes, in high dimensions, most points will be linearly separable from the rest of the data. This means that the underlying geometric hypotheses do not hold at most points, so that it is reasonable to expect significant difficulties in estimating dimension. Sensitivity to sample size will be discussed in the analysis of our expermental results in \Cref{sec:comparisons}.

\subsubsection{Empirical correction}

Some estimators (e.g. \danco and the Camastra--Vinciarelli adaptation of \corrint) seek to use empirical corrections of the perceived negative bias. They start with a previously documented estimator (\mindml or \corrint, respectively) which is believed to have a negative bias, run it on a family of datasets to get empirical results, and then use this to produce a correction factor. 

However, this relies both on the dataset under study being similar to the dataset used for correction and on the underestimation being generally true.
For example, \corrint performs accurately on $SO(n)$ and so providing an empirical correction would, in fact, reduce accuracy.

\section{Experiments: Results and Discussion}

\subsection{The \scikitdimension package}

We carry out our experiments using an augmented version of the \scikitdimension package~\cite{Bac2021Scikit-Dimension:Estimation}. This Python package provides a variety of dimension estimation methods equipped with default parameters as suggested by the authors of individual methods.

Additional methods have been added to the package by the present authors, including amendments to the architecture to allow neighbourhoods to be selected by radius rather than nearest neighbours.

The new methods incorporated \geomle, \gride, \cdim, \wodcap, Camastra \& Vinciarelli's extension to the Grassberger Procaccia algorithm for \corrint, the packing-number based estimator, and the magnitude and \phzero estimators. New functionality is added for \lpca and \mle, where users can choose $\epsilon$-neighbourhoods in addition to $\knn$ neighbourhoods. We also added a probabilistic thresholding method for \pca~\cite{Minka2001AutomaticPCA}.  We recommend that new and old estimators are implemented into this framework to give practitioners a focal point to begin. The code for the new estimators can be found on this  \href{https://github.com/kmyim/scikit-dimension}{Github fork of the \scikitdimension package}.

\subsection{Datasets}
\label{ssec:datasets}
We use a now standard collection of datasets for benchmarking purposes~\cite{HeinMatthias2005IntrinsicRd,Campadelli2015IntrinsicFramework,Rozza2012NovelEstimators}, with a small number of additions. These datasets are readily generated in \texttt{scikit-dimension}.
The datasets encompass a large range of dimensions (1 to 70), codimensions (0 to 72) and geometries (flat, constant curvature, variable curvature). We should note that not all of the datasets are drawn from uniform distributions on their manifolds.
Each underlying manifold is diffeomorphic to either a sphere or a cube.
We give a brief description of each in \Cref{tab:BM_desc}. 

\begin{table}
\footnotesize

\begin{tabular}{| l || l | l | l |}\hline 
Dataset & $d$ &  $D$ &  Description \\ \hline 
M1\_Sphere & 10 & 11 & Uniform distribution on round sphere \\ \hline

M2\_Affine\_3to5 and M9\_Affine & 3, 20 & 5, 20 & Affine subspaces \\ \hline

M3\_Nonlinear\_4to6 & 4 & 6 & Nonlinear manifold, could be mistaken to be 3d.
\\ \hline

M4, M6 and M8\_Nonlinear & 4, 6, 12 & 8, 36, 72  & Nonlinear manifolds generated from the same function
\\ \hline

M5a\_Helix1d & 1 & 3 & A 1d helix
\\ \hline

M5b\_Helix2d & 2 & 3 & Helicoid 
\\ \hline

M7\_Roll & 2 &3& Classic swiss roll
\\ \hline

M10a,b,c,d\_Cubic &10, 17, 24, 70 &11, 18, 24, 72 & Hypercubes
\\ \hline

M11\_Moebius & 2& 3 & The 10 times twisted Moebius band
\\ \hline

M12\_Norm & 20 & 20 & Isotropic multivariate Gaussian
\\ \hline

M13a\_Scurve& 2 & 3 & Surface in the shape of an ``S"
\\ \hline

M13b\_Spiral & 1& 13& Helix curve in 13 dimensions.
\\ \hline

 Mbeta &10& 40& Generated with a smooth nonuniform pdf
\\ \hline

Mn1 and Mn2\_Nonlinear & 18, 24 & 72, 96 &  Nonlinearly embedded manifolds of dimensions 
\\ \hline

Mp1, Mp2 and Mp3\_Paraboloid & 3, 6,, 9 & 12, 21, 30 & Paraboloids nonlinearly embedded of dimensions
\\ \hline

\end{tabular}
\caption{Description of the benchmark manifolds from \scikitdimension~\cite{Bac2021Scikit-Dimension:Estimation}.}
\label{tab:BM_desc}

\end{table}

For some purposes we have also considered the standard matrix embedding of $SO(n)$ in $\bbr^{n \times n}$, though we have not fully benchmarked this dataset. This produces a homogeneous manifold of dimension $\frac{n(n-1)}{2}$, not lying in any affine subspace, and having a new topology compared to the other benchmark datasets. We feel that these datasets would be good additions to the benchmark manifolds of~\cite{Bac2021Scikit-Dimension:Estimation}, as including manifolds with known geometric and topological information will increase knowledge of where specific estimators work well. 

For our experiments investigating the effects of noise, curvature, we consider a collection of datasets where we have a good control and understanding of their dimensions and geometry. These include a torus of revolution in $\bbr^3$ (which has no boundary) as well as families of paraboloids with varying curvature shown in Figures \ref{fig:toruspn}, \ref{fig:EP} and \ref{fig:HPara}.

\subsection{Comparisons of estimators on the benchmark datasets}\label{sec:comparisons}

For each estimator in the cohort listed in \Cref{tab:assessment_on_bm}, we investigate various potential challenges which can identify areas of weaknesses. The areas of focus are:
\begin{itemize}
    \item Dependency on a tailored choice of hyperparameters to dataset to achieve near optimal performance;
    \item Whether the estimator requires many point samples to perform well;
    \item Performance on high dimensional datasets with flat geometry, such as M9 Affine, M10b,c,d, Cubic;
    \item High variance in performance between different random samples from the same measure;
    \item Poor performance on datasets that are non-linear, such as those with high curvature or uneven sampling density. These include M6 Nonlinear, M12Norm, Mbeta, and the Mn and Mp Nonlinear and Paraboloid datasets. 
\end{itemize}
Note that an estimator performing well against an individual criterion above does not make it a ``good'' estimator. A trivially bad estimator that outputs the same dimension for any dataset is both insensitive to the choice of hyperparameter (none) and has zero variance. 

In \Cref{app:bmtables}, we detail the experimental procedure to assess a list of estimators against the benchmark dataset and give the numerical results of those experiments. We defer discussions of each estimator's performance on the benchmark datasets with respect to the metrics above to the captions of the results tables in \Cref{app:bmtables}. In this section we give a qualitative summary of the observations on each estimator in \Cref{tab:assessment_on_bm} and an overview of how the estimators perform against the criteria. 

\begin{remark}
    Not all of the estimators discussed in \Cref{sec:estimator_description} are included in our full experiments. Magnitude dimension suffers from being sample hungry and performs poorly on even spheres of moderate dimensions. The conical dimension and packing number estimators are computationally inefficient (at least in our possibly naive implementations) and take too long to compute to generate sufficient data for calculations of standard deviations.  We recommend further research into how to improve the efficiency of those estimators. 
\end{remark}

\paragraph{Dependency on a tailored choice of hyperparameters} In our experiments, we varied the hyperparameters of estimators around those used in the original papers, or the defaults in the \texttt{scikit-dimension} implementation, as reported in  \Cref{app:hyperpm}.  We compared the performance of the best possible estimate an estimator can achieve with optimal hyperparameters within our range, and the estimate achieved by choosing hyperparameters that ensures good performance on most of the datasets in the benchmark set (this hyperparameter is defined precisely in \Cref{app:bmtables}). A large discrepancy in results between these two choices indicates that an estimator needs tailored choices of hyperparameters to achieve optimal results. This is an undesirable effect, 

For the collection of ``slope-inference'' based global estimators considered here -- \knnestimator, \phzero, \twonn, and \gride{} -- this discrepancy is often small, especially on low dimensional datasets without complicated nonlinearities. This may be due to the fact that there are only one or two hyperparameters for these estimators, far fewer compared to others on our list. There seems to be a locality bias for these estimators. For \knnestimator, the close to optimal performance (on low dimensional datasets without complicated nonlinearities) can be reached by choosing the nearest neighbour parameter $k$ to be close to one (see \Cref{tab:knn_bmhp}, and also \Cref{fig:knn_bm} for an illustration). For $\phzero$, we can also guarantee reasonable performance with the choice $\alpha = 0.5$, which emphasises the contribution of small distances over large ones across edges of the minimum spanning tree (\Cref{tab:ph_bmhp}). Choosing the hyperparameters that reduce \gride to \mle (with input from the distance to the two nearest neighbour distances) is often effective (\Cref{tab:gride_bmhp}). We also note that given such hyperparameters, the performances of \gride and \twonn are often similar in \Cref{tab:gride_bm,tab:twonn_bm}.

We defer discussions about other estimators and their need to tune hyperparameters to specific data in the captions of the benchmark results tables in \Cref{app:bmtables}. For local estimators, the specific issue of tuning the neighbourhood size and aggregation method is analysed in greater detail in \Cref{sec:NbhdSize,sec:aggregation}. 
\begin{remark}
    We emphasise that our empirical study is subject to our particular choices of hyperparameter ranges, which cannot encompass all possible hyperparameter combinations due to finite computational resources. We refer the reader to \Cref{app:hyperpm} for the range of hyperparameters chosen for each estimator, which was guided by the literature. 
\end{remark}

\begin{figure}
    \centering
    \begin{subfigure}{0.85\textwidth}
        \centering
        \includegraphics[width=\linewidth]{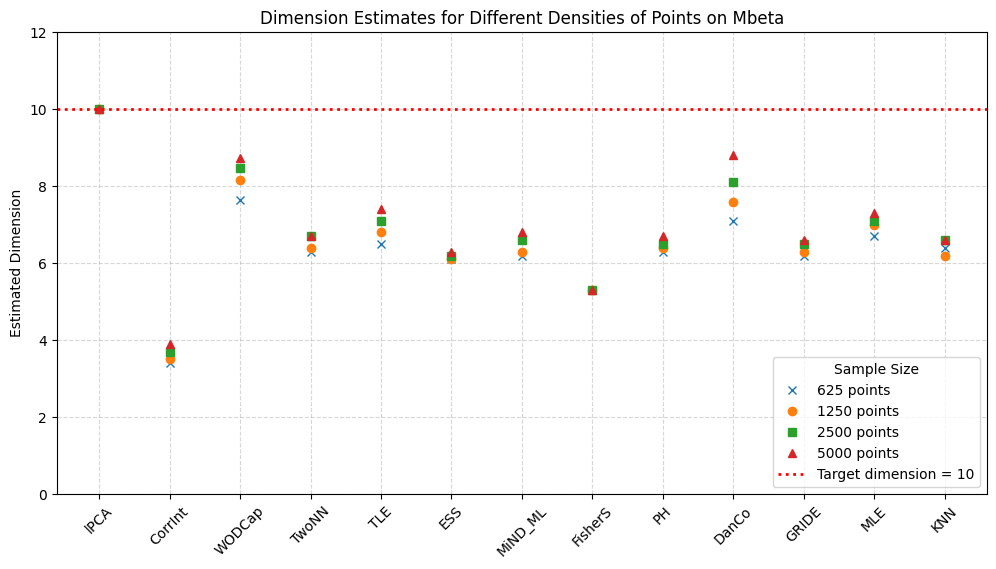}
        \caption{}
        \label{fig:first}
    \end{subfigure}
    \hfill
    \begin{subfigure}{0.85\textwidth}
        \centering
        \includegraphics[width=\linewidth]{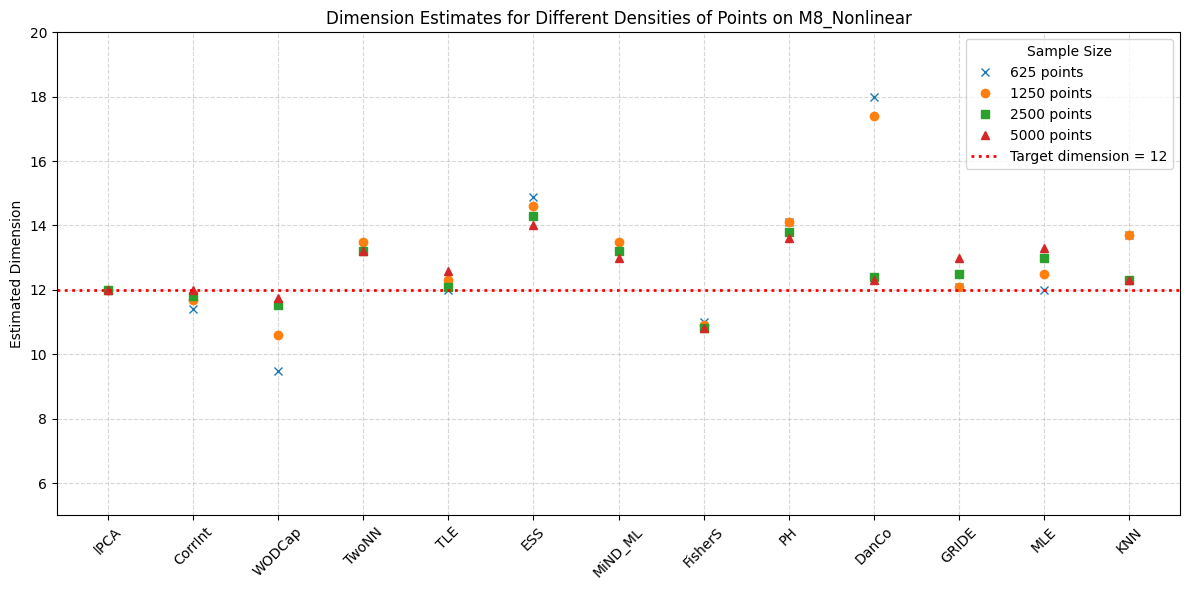}
        \caption{}
        \label{fig:second}
    \end{subfigure}
    \caption{The best estimates from our benchmark tables for 625, 1250, 2500 and 5000 points for each estimator on two different datasets, Mbeta and M8Nonlinear. As a general rule, increasing sample size improves accuracy. However, to return a correct estimate would require a lot more than 5000 points from these datasets for most estimators. Some estimators have a clear bias which does not reduce with increasing sample sizes. Also seen is a differing level of responses to changing sample size, which appear broadly consistent across the two datasets. For example, \wodcap underestimates, becoming much better as sample size increases, while on the other hand \fishers and \lpca do not change significantly from 625 points to 5000. }
    \label{fig:Dense}
\end{figure}

\paragraph{Sample Economy} We restrict our assessment here to how an estimator performs on low dimensional datasets (dimension $< 6$), as most estimators face challenges with even moderately high dimensional datasets. One surprising observation is the slow increase in most estimators' accuracy with the number of samples, at least in the regime of $N \in \{625, 1250, 2500, 5000\}$ being tested. As summarised in \Cref{tab:assessment_on_bm}, and detailed in \Cref{app:bmtables}, this is often true with global estimators such as \phzero, \knnestimator, \mindml, \gride, and \twonn where $N \in  \{625,1250\}$ often results in accurate dimensions estimation on low dimensional datasets.  On the other hand, while this is also observed on some local estimators such as \ess, \tle, other local estimators such as \lpca, \mle only recover comparable performance in the $N \in  \{2500,5000\}$ regime, suggesting that they are more sample hungry. Given fewer samples, \knn neighbourhoods have a larger effective radius and so  non-flat geometries in the dataset can further bias the estimation. \Cref{fig:Dense} demonstrates the behaviour of all estimators on two datasets.

\paragraph{Accuracy on high dimensional datasets}
Overall, most estimators tend to underestimate, even on their best hyperparameters. This is often especially acute on high dimensional datasets, and some nonlinear datasets.
One exception is \lpca, which appears to give the correct dimension every time. However, as we will discuss below, this is mainly due to our ability to tune the hyperparameter to give the correct result; with another hyperparameter the estimate could be substantially incorrect (see \Cref{fig:FO}). 

As we would expect, most estimators are very good on low dimensional data, and struggle when the dimension increases beyond 6. There are exceptions to this rule, with \ess and \danco performing very well on the high dimensional datasets. We note that, as the sample size increases, most estimators improve. However, there are exceptions, with \danco, \ess and \fishers not changing substantially on most datasets as sample size increases.

\paragraph{Variance of dimension estimates}
Local estimators, such as \lpca, \mle, \mindml, \wodcap, and \tle, tend to have a low variance. These estimators aggregate many local dimension estimates into a global dimension estimate. The standard deviation on the mean, median, or harmonic mean of the local estimates decreases with the number of local estimates.  On the other hand, estimators such as \phzero and \knnestimator suffer from higher error sensitivity in high dimensions which can increase variance (see \Cref{rmk:ph_knn_error}).  

\begin{figure}[h]
    \centering
\includegraphics[width=0.8\linewidth]{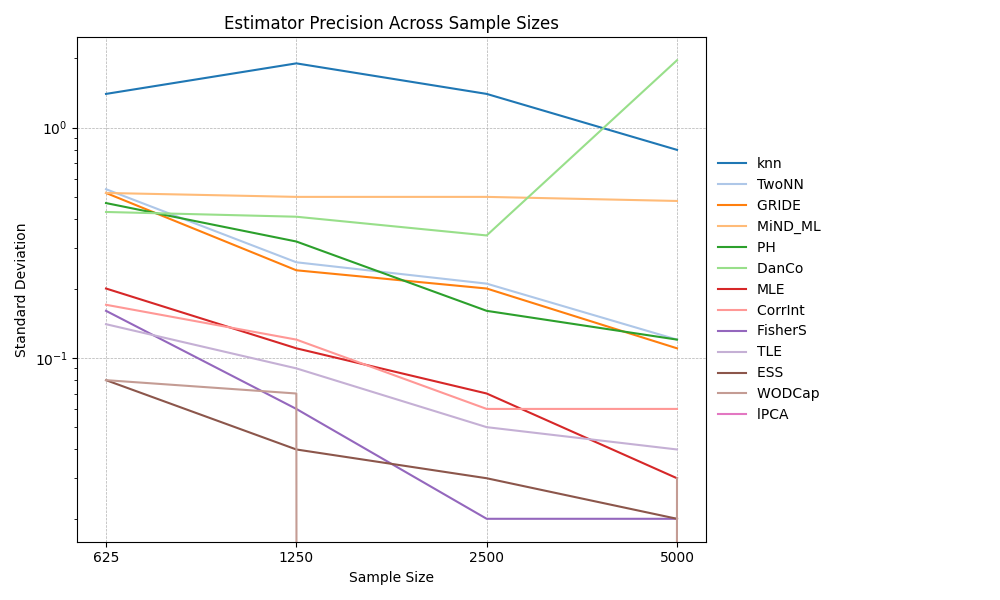}
    \caption{Standard deviation of dimension estimates of M1Sphere (uniform samples on $S^{10} \subset \bbr^{11}$) over 20 runs, as shown in the tables in \Cref{app:bmtables}, using the hyperparameters which provide the best estimate. This plot has a logarithmic scale on both axes. \lpca has standard deviation zero, as does \wodcap for $N=2500$, and so these points are not shown.}
    \label{fig:1011stdev}
\end{figure}

In \Cref{fig:1011stdev}, we visualise the effect of increasing sample size on the variance of the estimate on M1Sphere (uniform samples on $S^{10} \subset \bbr^{11}$). We note that the standard deviation of \mindml, \knnestimator and \danco decreases at a slower rate than the other estimators as the number of samples increases. Indeed, the standard deviation of \danco increases, which may be due to an unusual choice of hyperparameters yielding the best mean estimate.

The variance might be expected to decay with increasing sample size with rate $N^{-1}$.
However, we observe that for $N=625$ the variance is higher than would be expected.
As $N$ increases from 625 to 1250, 2500, and 5000, we might expect the variance to be $8$, $4$ and $2$ times greater than its final value at $N=5000$. 
In fact, taking the median of these ratios over the estimators (discounting \lpca, which has 0 variance, and \danco, due to the unusual behaviour at $N=5000$), we observe $15.3$, $5.1$ and $1.8$.
This suggests that, for many estimators, small sample sizes which result in larger neighbourhoods that are less well approximated by flat spaces create additional variance.

\paragraph{Nonlinear datasets}
Most estimators struggle quite a lot with these datasets. An exception is \lpca though, as discussed, hyperparameter selection can ensure good performance for \lpca. We conjecture that since \lpca infers the dimension of the tangent plane, nonlinear features such as variable sampling density generate a smaller bias than with parametric estimators such as \mle. 

It is interesting to compare which estimators work well on which datasets. For example, \mle, \phzero, \knnestimator, \gride, \twonn, \mindml, and \corrint tend to perform well on M3,4,8 Nonlinear and struggle with M12 Norm and Mn1,2 Nonlinear; yet it is the reverse with \lpca, \danco,  and \ess.  One particularly difficult dataset is Mbeta. As a 10-dimensional manifold with ambient dimension 40, the dimension and codimension are both relatively high. Furthermore, density and curvature are variable.

Some estimators, such as \phzero, \knnestimator, and \corrint, are meant to capture a notion of dimension of a metric \emph{measure} space, which can be distinct from the dimension of the support of the measure. Hence, for datasets such as samples from the normal distributions (M12 Norm), the estimated dimension of such estimators can be different to the dimension of the support. 

\begin{table}[h]
\centering
\begin{tabular}{|l|l|l|l|l|l|}
\hline
        & No tailoring of params & Sample economy & High dim ($>6$) & Low variance & Nonlinear \\ \hline
lPCA    &                          &                &                      & $\checkmark$ &                       \\ \hline
MLE     &                          &                &                      & $\checkmark$ &                       \\ \hline
PH      & $\checkmark$             & $\checkmark$   &                      &              &                       \\ \hline
KNN     &  $\checkmark$                        & $\checkmark$   &                      &              &                       \\ \hline
WODCap  &                          &                &                      & $\checkmark$ &                       \\ \hline
GRIDE   & $\checkmark$             & $\checkmark$   &                      &              &                       \\ \hline
TwoNN   & $\checkmark$             & $\checkmark$   &                      &              &                       \\ \hline
DANCo   &                          &                & $\checkmark$         &              &                       \\ \hline
MiND ML &                          & $\checkmark$   &                      & $\checkmark$ &                       \\ \hline
CorrInt & $\checkmark$             &                &                      &              &                       \\ \hline
ESS     & $\checkmark$             & $\checkmark$   & $\checkmark$         &  $\checkmark$              &                       \\ \hline
FisherS &                          &                &                      & $\checkmark$ &                       \\ \hline
TLE     &                          & $\checkmark$   &                      & $\checkmark$ &                       \\  \hline 
\end{tabular}
\caption{\label{tab:assessment_on_bm} Qualitative assessment of performances of estimators on the benchmark dataset. None of the estimators consistently perform well on the nonlinear datasets in the benchmark.}
\end{table}

\subsection{Influence of hyperparameter choices on estimated dimensions}\label{sec:hyperparameter_choice}
\begin{figure}[h]
    \centering
    \includegraphics[width=0.9\linewidth]{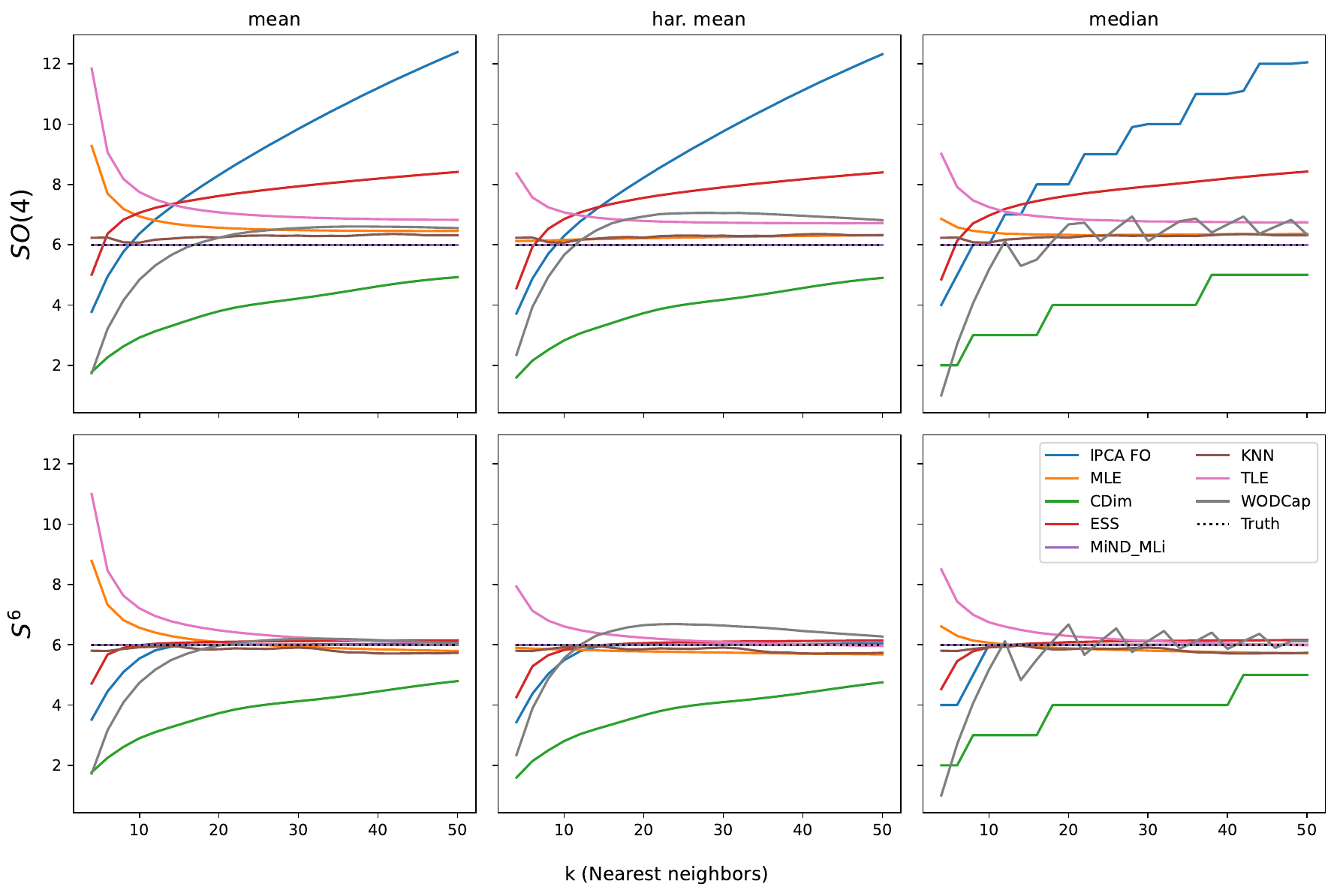}
    \caption{Comparison of dimension estimates of $SO(4)$ and $S^6$ for estimators with a $k$-nearest neighbour parameter. The input data consists of 2500 points uniformly sampled on the manifolds, which have intrinsic dimension 6. The $k$ parameter is varied from 4 to 50 in steps of 2. For local methods -- all on the figure, apart from \mindml and \knnestimator{} -- we vary the method of aggregation over local estimates. Note that \wodcap does not aggregate the local dimension estimates, but rather the estimated volume fraction of spherical caps. We also remark that \mindml, which is hard to distinguish on the plot, consistently returns an estimate of 6. }
    \label{fig:nbhd}
\end{figure}

\subsubsection{Aggregation }\label{sec:aggregation}
With local estimators, a dimension estimate is produced on different local neighbourhoods of query points, and a global dimension estimate is then obtained by aggregation. The choice of aggregation method can be considered to be a hyperparameter passed to the estimator. We considered varying the aggregation method between the arithmetic mean, median, and harmonic mean.

While taking the mean or the median of the different pointwise estimates may be obvious choices, this may lead to biases in the estimates for probabilistic methods. MacKay and Ghahramani suggested the use of the harmonic mean of local dimension estimates to produce a global dimension estimate~\cite{MacKay2005Comments2004}.
In this particular case, the suggestion is based on the fact that this is the maximum likelihood estimator for the global dimension, under the simplifying assumption that the estimates at each point are independent.
The harmonic mean is frequently used when averaging rates and its utility here is suggestive of dimension as a rate at which volume opens up with radius~\cite{Schumacher2020EstimateLID}.

\Cref{fig:nbhd} makes clear the benefit of using alternative aggregation methods, especially when using small neighbourhood sizes. It shows how dimension estimation varies when we change the neighbourhood size parameter for $n = 2500$ samples of $S^6$ and $SO(4)$. If we use the mean, \mle and \tle show divergent results for low values of $k$. These are much improved by using the harmonic mean instead, 
\ess, on the other hand, performs better for low $k$ when the arithmetic mean is used.

For \lpca, \cdim, and \wodcap, their estimation process involves enumerating objects, and the output range of local estimates is discrete  (in the case of \lpca and \cdim the actual estimate is an integer dimension). The distribution of local estimates across the data can fall on a very small support $\ll |\bbx|$. We observe that the median aggregated dimension can be sensitive to shifts in the distribution; for example, if half of the local estimates are 6 and the other half are 7, a very small perturbation to the dataset can skew the distribution to output a median dimension estimate of either 6 or 7. Similarly, a change in the hyperparameter (in this case, the $k$-nearest neighbour parameter) can skew the relative proportion of estimates over the discrete set of outputs, and the output can be unstable with respect to the change in hyperparameter. We see that this instability is an artefact introduced by the median aggregation, since the mean and harmonic mean vary continuously with $k$. 

\subsubsection{Local Neighbourhood Size}\label{sec:NbhdSize}

\paragraph{Bias}
Figure \ref{fig:nbhd} shows how estimators behave as the parameter $k$, controlling the size of \knn neighbourhoods, changes. We consider two sample datasets, $S^6$ and $SO(4)$, both of which are 6-dimensional. As noted in Section \ref{sec:aggregation}, the harmonic mean provides the best estimates in general and so we will focus on this method.

The estimators obtained from distributions of nearest neighbour distances (\mle and \tle, excluding the integer-valued MIND MLi) appear to converge to a common value as $k$ grows, which for $SO(4)$ is an overestimate but for $S^6$ is an underestimate.
For smaller values of $k$, both \mle and \tle tend to overestimate. 
\mle, with harmonic mean, performs best for small values of $k$.

One clear potential source of bias for parametric estimators of this type is the failure of the underlying hypotheses to hold. 
In general, these are, firstly, that a sample from a ball centred on $p \in \bbx$ yields points with the same distance from $p$ as a sample with uniform density from a Euclidean ball would and, secondly, that the distances $r_i(p)$ for each $p \in \bbx$ are independently and identically distributed.

The failure of the first assumption can be caused by the failure of the embedding to be totally geodesic, by the existence of non-manifold points phenomena, including the presence of boundary, and by variable sampling density.

The second assumption is not true: for points $p, q \in \bbx$ which are close to each other the statistic $r_i$ is positively correlated. Furthermore, since $\bbx$ arises from a binomial point process rather than a Poisson point process, the existence of a densely sampled region with low values for $r_i$ necessarily implies that the remainder of $M$ is more sparsely sampled, so that for points $p, q \in \bbx$ which are far from each other, $r_i$ will be negatively correlated.
As argued in~\cite{Levina2004MaximumDimension}, it seems likely that the long-range effects are much weaker, and it is shown in~\cite{Denti2022TheEstimator} that distance ratios are independent as long as they come from disjoint neighbourhoods.
Trunk~\cite{Trunk1968StatisticalCollections} found spatial correlations were not significant by comparing empirical distributions to the theoretical distribution which arises from the assumption of independence and using a Kolmogorov--Smirnov test. However, the exact experimental procedure is unclear: from context it seems likely that at best $d \leq 4$ is considered.

All of these errors will tend to become more evident as neighbourhood size increases, so that it is reasonable to anticipate that parametric methods will experience a bias that increases with $k$, as \mle appears to.

The remaining estimators, \lpca and \cdim, which seek an approximating affine subspace, and \ess, demonstrate a tendency to increase with $k$. For \lpca and \cdim, the estimate relies on how well the collection of $k+1$ points approximates an affine subspace. The underlying hypotheses are less delicate, with the failure to be totally geodesic and the existence of non-manifold points being the sources of error. However, boundary points need not be an issue. As noted below, a good estimate requires $k$ to be sufficiently large. However, the data for $SO(4)$ demonstrate how, in a high codimension setting, \lpca can significantly overestimate if $k$ is too high.

\begin{figure}
    \centering
    \includegraphics[width=0.85\linewidth]{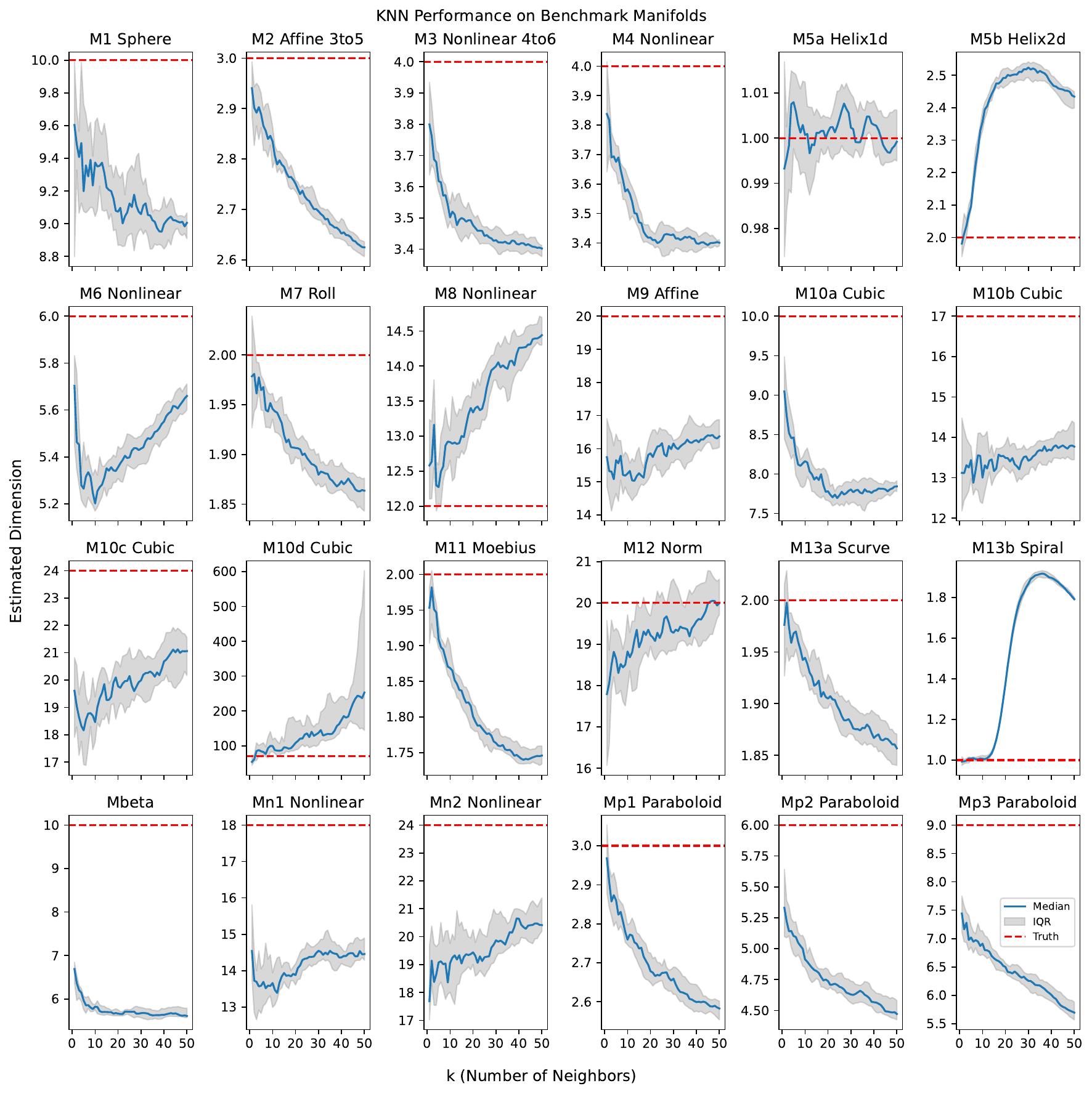}
    \caption{Performance of $\knnestimator$-estimator vs $k$ on benchmark manifolds, for 5000 samples on each manifold. We show the empirical median dimension estimate, and the interquartile range, over 20 sets of random samples. Note the different scaling of the $y$-axis from dataset to dataset. The high variance for M10d Cubic is due to the sensitivity to errors in high dimensions of the method; see \Cref{rmk:ph_knn_error}.}
    \label{fig:knn_bm}
\end{figure}

\paragraph{Throttling}

The number of points in a neighbourhood can ``throttle'' the dimension estimator, so that it is impossible for it to return an estimate above a certain value.

\begin{figure}
    \centering
\includegraphics[width=0.8\linewidth]{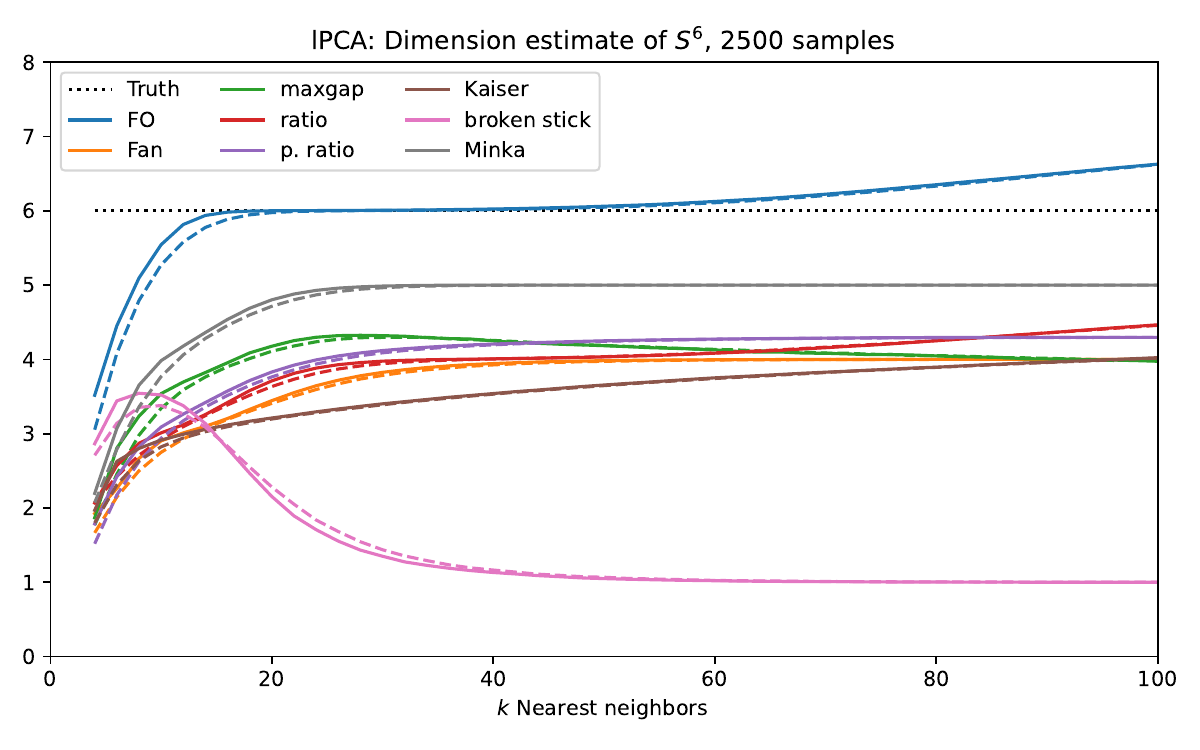}
    \caption{Dimension estimates from \lpca when neighbourhood sizes are varied from 4 to 100, for $S^6 \subset \mathbb{R}^7$. Colours indicate different threshold methods. Solid curves correspond to estimates using \knn neighbourhoods,  dashed curves correspond to using $\epsilon$-neighbourhoods, where for each nearest neighbour value $k$, the radius $\epsilon$ is chosen to be the median \knn distances}
    \label{fig:S6nbhdlPCA}
\end{figure}

\begin{figure}
    \centering
\includegraphics[width=0.8\linewidth]{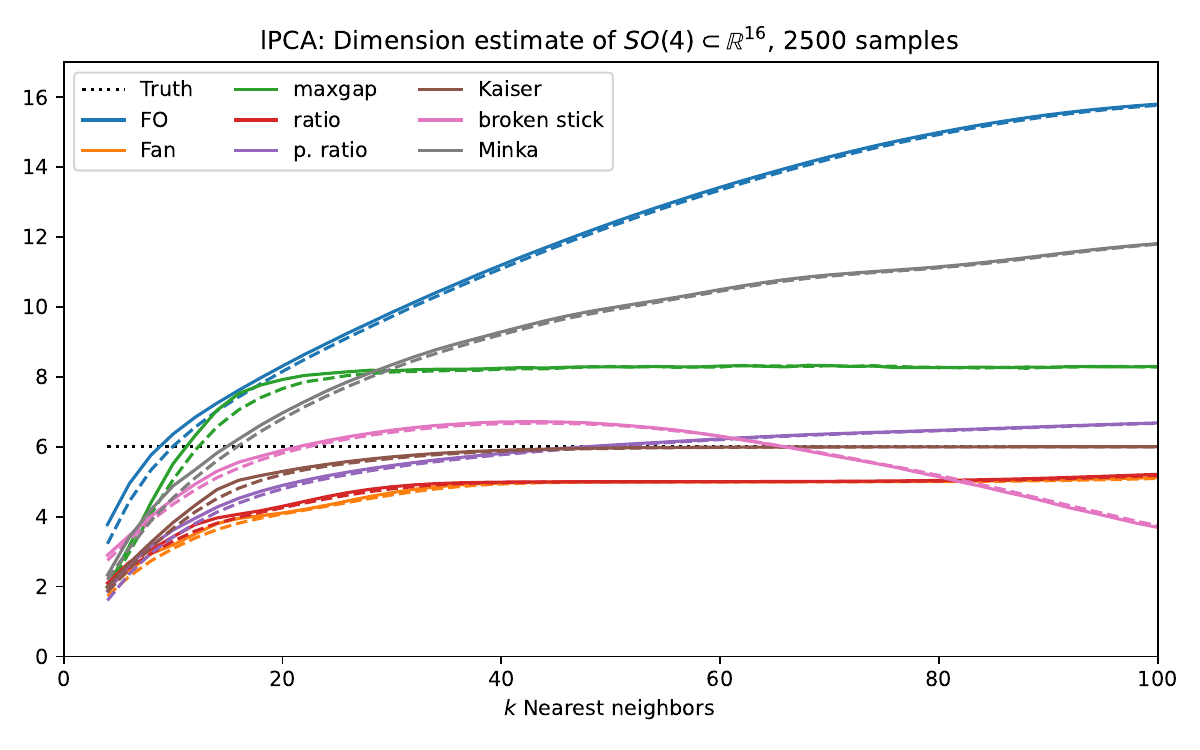}
    \caption{Dimension estimates from \lpca when neighbourhood sizes are varied from 4 to 100, for $SO(4) \subset \mathbb{R}^{16}$. Colours indicate different threshold methods. Solid curves correspond to estimates using \knn neighbourhoods, dashed curves correspond to using $\epsilon$-neighbourhoods, where for each nearest neighbour value $k$, the radius $\epsilon$ is chosen to be the median \knn distances.}
    \label{fig:SO4nbhdlPCA}
\end{figure}

For example, if $\knn$ neighbourhoods are used, then it is clear that $\lpca$ can only give $k$ non-zero eigenvalues and so the dimension estimate will never exceed $k$. We can describe this as linear throttling -- for an accurate estimate the parameter $k$ must grow linearly in $d$. This phenomenon is clear in \Cref{fig:S6nbhdlPCA} and \Cref{fig:SO4nbhdlPCA} where throttling occurs for $4 \leq k \leq 6$

However, there are also estimators with exponential throttling, where $k$ must grow exponentially in $d$.
For example, the estimates of~\cite{Kleindessner2015DimensionalityDistances} suffer from their discrete nature.
Volumes are approximated by numbers of points in balls built from the \knn graph. 
The bounded valence of the graph is what generates throttling in this instance.
The doubling property estimator, for example, can never exceed $\log_2 (k+ \frac{1}{k+1})$.
This is because $|B_{\mathrm{e}}(i,1)| =k+1$ always and, since each \knn of $x_i$ has at most $k+1$ unique nearest neighbours, the maximal value of $|B_{\mathrm{e}}(i,2)| $ is $k(k+1)+1$. 

The maximum value which can be returned by \wodcap is $S^{-1}(\frac{2}{k+1})$ where 
\[S(\hat d) = I_{\frac34} \left(\frac{\hat d + 1}{2} , \frac{1}{2} \right). \]

Given that we have $S(d) = I_{\frac{3}{4}}(\frac{d+1}{2},\frac{1}{2}) \approx \frac{2}{k+1}$. We can write the regularized incomplete beta function in terms of the Gamma function and an integral: $$S(d) = \frac{\Gamma(\frac{d+1}{2},\frac{1}{2})}{\Gamma(\frac{d+1}{2}) \Gamma(\frac{1}{2})} \int_{0}^{\frac{3}{4}} t^{\frac{d-1}{2}}(1-t)^{-\frac{1}{2}} \diff t$$

Then by using Gamma function identities, if $d$ is even, we arrive at $$ k \approx \frac{2}{\int_{0}^{\frac{3}{4}} t^{\frac{d-1}{2}}(1-t)^{-\frac{1}{2}}dt} \frac{\pi d!}{2^{d-2}((\frac{d}{2})!)^2} -1.$$

The right term of the product on the right hand side goes to $0$ as $d \rightarrow \infty$; however, it is completely dominated by the reciprocal of the integral. As if we plug in $d = 2$, $20$ and $200$, we get  $ k \approx 19$, $528$ and $2.6 \times 10^{14}$ respectively. From this it appears that as $d$ grows $k$ needs to grow exponentially. Since $k$ is intended to define a small neighbourhood, the number of points of the sample required for the method to estimate $d$ becomes completely unfeasible.

Probabilistic phenomena can cause throttling as well, in the sense that $k$ must grow at a certain rate for a correct estimator to be returned with a given probability. 
An example occurs with $\cdim$, where Figure \ref{fig:nbhd} is suggestive of throttling.
Note that the theoretical guarantees for \cdim already require $k$ to grow exponentially with $d$.
While this sample size is sufficient for the estimator to work, it need not be necessary.
The algorithm to compute the estimate finds the largest possible subset of directions to nearest neighbours where all pairwise angles are at least $\pi/2$.
In a large dimensional space, the angle between any two directions is very likely to be close to $\pi/2$, so that for any given pair it is approximately equally likely that the angle is greater than or less than $\pi/2$.
For a subset of length $\hat{d}$, the probability that any given additional vector can be added to it is approximately $2^{-\hat d}$.
Since there are $k-\hat d$ neighbours to check, the probability that the subset cannot be enlarged is \[\left(\frac{2^{\hat d}-1}{2^{\hat d}}\right)^{k - \hat d}.\]
Once $\hat d$ is relatively large, it is therefore necessary to have a very large value of $k$ in order to obtain a new direction which is obtuse to the existing $\hat d$ directions with a given probability.
In fact, using Markov's inequality and considering the matrix of inner product signs as in~\cite{Sun2008OnMatrix}, we can see that the probability of finding $\hat d$ such directions is $\binom{k}{\hat d} 2^{-\hat d (\hat d - 1)/2}$, and so, by Stirling's formula, for large $d$ the probability of finding $\hat d$ directions is bounded above by $C \frac{2^{k - \hat d (\hat d - 1)/2}}{\sqrt{k}} $. This indicates that $\cdim$ suffers at least quadratic throttling.

All the tangential estimators described in this survey are vulnerable to at least linear throttling. This appears inevitable, since the tangent space is the linear structure of the manifolds, the affine space which approximates the data. Unless $k \geq d$ we cannot hope to accurately recover the entire affine space.

This consideration makes clear that, for very high-dimensional datasets, the use of a tangential estimator cannot be recommended. The presence of throttling can be detected in \lpca by comparing the estimated dimension to the hyperparameter $k$ and we recommend that implementations of \lpca warn users when throttling is occurring.

\paragraph{Noise}

The use of smaller neighbourhoods is dangerous in the presence of noise, where it will tend to produce overestimates while in the presence of curvature, larger neighbourhoods are more vulnerable. The influence of noise is discussed in \Cref{sec:noise}

\subsubsection{Influence of thresholding methods on \lpca}
\input{tables/lpca_thresh_n_5000_k_80_mean}

In \Cref{tab:lpca_thresh}, we observe how different choices of thresholding methods for \lpca have a significant effect on the dimension estimate. On our benchmark datasets, while the FO thresholding method works better than most, there are datasets such as M4 Nonlinear, M5b and Helix2d where it is less accurate than other thresholding methods. The range of dimension estimates displayed in \Cref{tab:lpca_thresh} for different thresholding methods across many datasets demonstrates the influence of the thresholding method. 

\begin{figure}
    \centering
   
    \includegraphics[width=0.8\linewidth]{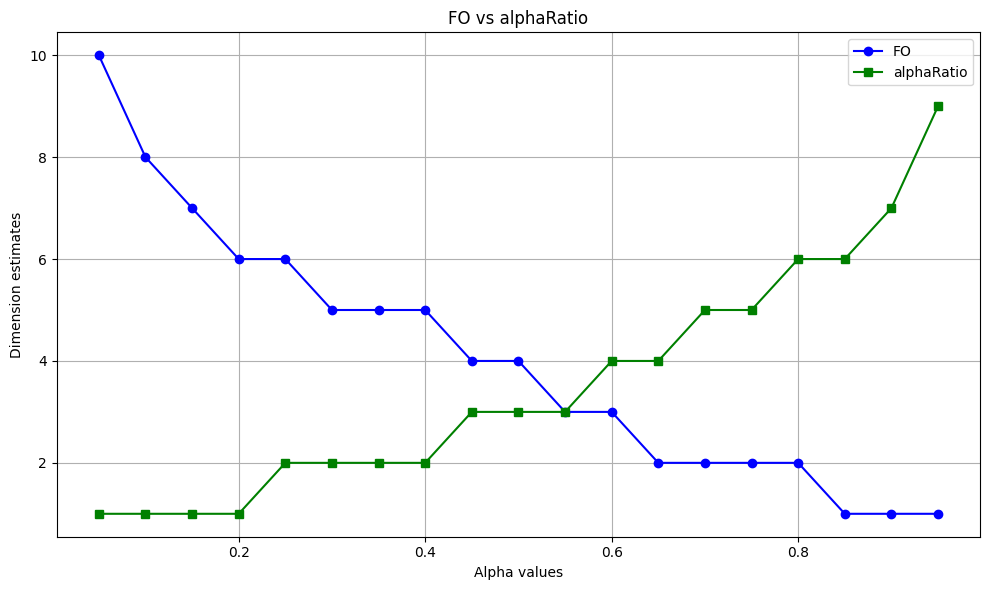}

    \caption{We consider 5000 points on M6\_Nonlinear and fix the neighbourhood size at 50. Searching a range of values for the hyperparameter $\alpha$ in both alphaFO and alphaRatio, we observe that the methods mirror each other. Both can produce the correct dimension, 6, but only for a narrow bandwidth of the hyperparameters. Outside of these bandwidths the estimates can vary significantly. The comparison shows that although $\alpha=0.05$ is the default for both as the ``out of the box'' value in \scikitdimension, the results are very different. Hence, a thorough understanding of what the hyperparameters represent and of their potential importance is essential. }
    \label{fig:FO}
\end{figure}

The choice of hyperparameters for thresholding methods also greatly affects estimates. We demonstrate this sensitivity by varying the $\alpha$ parameter for FO and alphaRatio on the M6 Nonlinear dataset, the outcomes which we show in \Cref{fig:FO}. We note that the out of the box choices for $\alpha$ for each version in \texttt{scikit-dimension} are both 0.05, producing a huge discrepancy between the two methods. However, both methods are capable of producing the correct answer for the perfectly tuned choice of hyperparameter.
It is crucial that practitioners understand the potential sensitivity in the operation of estimators before applying them.

\subsection{Robustness of estimators to noise}\label{sec:noise}

For the noise experiments, datasets were constructed by drawing 2500 points from the dataset detailed in Table \ref{tab:noise_ds}. These points were subsequently corrupted with either Gaussian noise or outliers. The methodology for outlier generation is fully described in Algorithm~\ref{alg:add_outliers}. Hyperparameter selection for the estimators is based on values previously optimized for the $S^{10}$ dataset under noise-free conditions. The final results, presented in the tables, represent medians calculated from 20 experimental repetitions.

\begin{algorithm}
\caption{Generate data set with outliers}
\label{alg:add_outliers}
\begin{algorithmic}[1]
\Procedure{AddOutliers}{$D, n_{out}$}
    \State \textbf{Input:} Clean dataset $D$ with $n$ observation of size $d$.
    \State \hspace{\algorithmicindent} \hspace{\algorithmicindent}  Number of outliers $n_{out}$
    \State \textbf{Output:} Datas set with outliers
    
    \State $out\_indices \gets random\_choice([1,...,n], n_{out})$ \Comment{Sample indices for outliers}
    \For{\textbf{each} $index$ \textbf{in} $out\_indices$}
        \State $point \gets D[index]$
        \For{$i \gets 1$ \textbf{to} $d$}
            \State $m \gets random(3,6)$  \Comment{Get random multiplier between 3 and 6}
            \State $point[i] \gets point[i] \cdot m $
        \EndFor
        \State $D[index] \gets point$
    \EndFor
    \State \textbf{return} $D$
\EndProcedure
\end{algorithmic}
\end{algorithm}

\begin{figure}
    \centering
    \begin{subfigure}{0.42\textwidth}
        \centering
        
    \includegraphics[width=\linewidth]{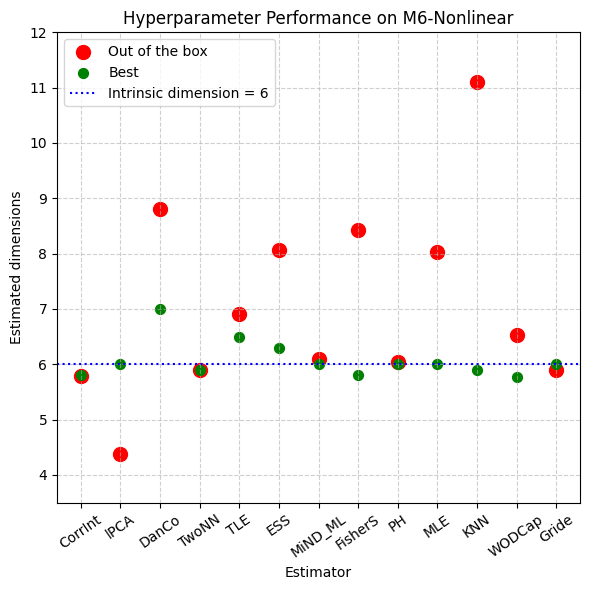}
        
        \label{fig:first}
    \end{subfigure}
    \hfill
    \begin{subfigure}{0.42\textwidth}
        \centering
        \vspace{5mm}
        
    \includegraphics[width=\linewidth]{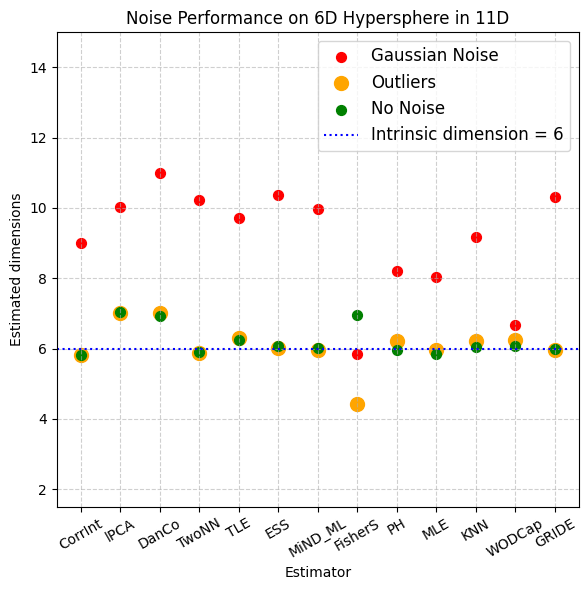}
        \label{fig:second}
    \end{subfigure}
    \vspace{1em}
    \caption{Left: We compare the average performance over 20 runs for the best performing hyperparamenters against the ``out of the box'' hyperparameters for each estimator on the M6 Nonlinear data set with 5000 points. Right: We compare the average performance over 10 runs across two different types of noise, gaussian noise with a standard deviation of 0.1, 125 outliers as described in Algorithm 1 and baseline of no noise. The sample size of 2500 points on $S^{6} \subset \bbr^{11}$ was used. The hyperparameters used were the best hyperparameters for $S^{10} \subset \mathbb{R}^{11}$. Most can handle outliers well but struggle with Gaussian noise. For \fishers this is reversed.}
    \label{fig:both_cn}
\end{figure}

The complete results are presented in Tables \ref{tab:611_gauss}–\ref{tab:1011_out}. Some of the most notable findings are illustrated in \Cref{fig:both_cn}.

The estimators' behavior in the presence of noise varies significantly, depending on both the type and magnitude of the applied noise. Furthermore, a comparison of the \lpca and \mindml estimators reveals that an estimator's robustness cannot be solely judged by its performance on a single dataset. While all three estimators perform exceptionally well for a 10-dimensional hypersphere (both with and without noise), a small amount of ambient Gaussian noise causes a significant overestimation for a 6-dimensional hypersphere.

All tested estimators exhibit susceptibility to ambient Gaussian noise. This is particularly evident when there is a substantial difference between the intrinsic dimension and the embedding dimension. In such instances, even a very small noise level (with a standard deviation of 0.01) significantly alters the results for some estimators (e.g., \lpca and \mindml in Table \ref{tab:611_gauss}). 

It is crucial to note that the parameters used for these experiments were optimized for the uncorrupted data. 
As some methods, such as \corrint and \gride, have parameters which are intended to prevent selecting neighbours on the noise scale, adjusting these parameters could potentially improve results for noisy data. 
Of the tested estimators, \corrint demonstrates the highest resistance to ambient Gaussian noise. 
Interestingly, estimates from \fishers decrease as the standard deviation of the disturbances increases, even though the disturbances are of a higher dimension than the initial dataset.

Most estimators demonstrated robustness to outliers, with the exceptions being \fishers. For the \fishers estimator, the addition of outliers led to a significant reduction in dimension estimates.

\begin{table}[]
    \centering
    \begin{tabular}{|c|c|c|}
         \hline
         Dataset & d & n  \\ \hline
         $S^6$& 6 & 11 \\ \hline
         $S^{10}$ & 10 & 11 \\ \hline
         $SO(4)$ & 6  & 16 \\ \hline
    \end{tabular}
    \caption{Datasets used for noise experiments. Intrinsic dimension of dataset and embedding dimensions are marked respectively by $d$ and $n$.}
    \label{tab:noise_ds}
\end{table}

\subsection{Varying responses of estimators to curvature}

\begin{figure}
    \centering
    \includegraphics[width=0.75\linewidth]{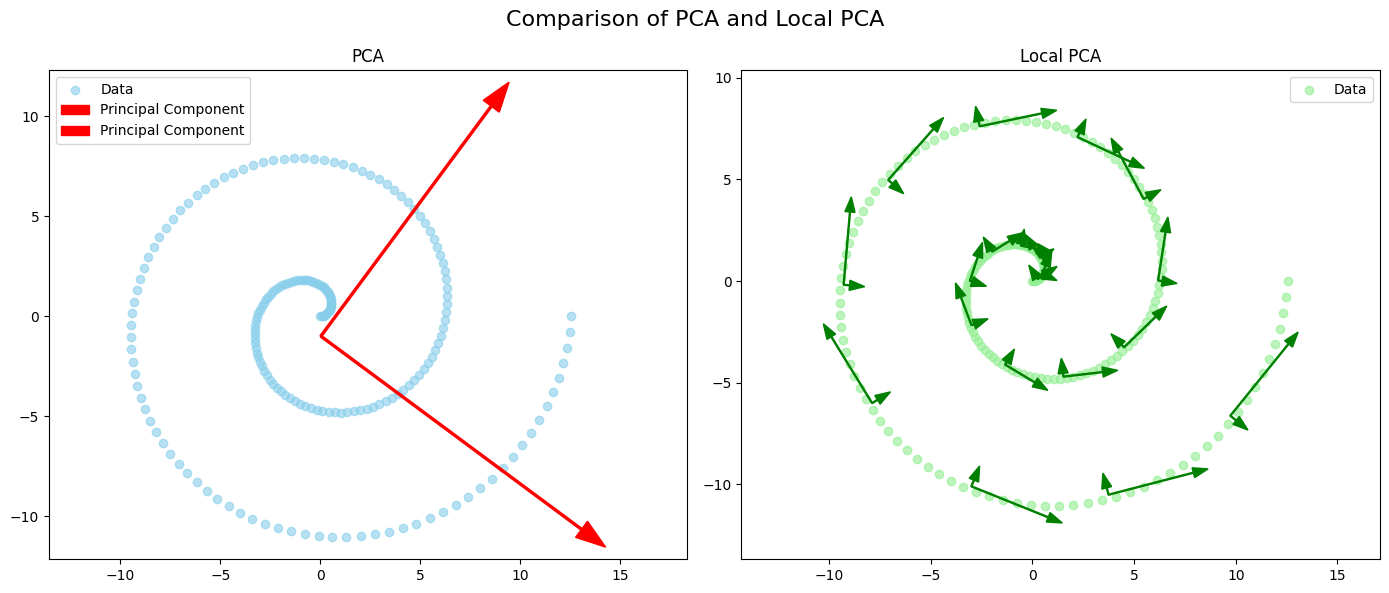}
    \caption{Demonstration of how curvature affects \pca when points are drawn from a non-linear curve in $\mathbb{R}^2$. By comparison, using \lpca mitigates the effect of curvature. If the neighbourhoods are chosen too large \lpca will also begin to struggle.}
    \label{fig:PCAvslPCA}
\end{figure}

The underlying manifold of data is unlikely to be entirely flat -- if it is, then \pca will be a sufficient dimension estimator.
It is therefore important to understand how how curvature can affect dimension estimation. 
Consider, for example, the performance of \pca and of \lpca on points on drawn from a non-linear curve in $\mathbb{R}^2$ as shown in \Cref{fig:PCAvslPCA}.

\begin{figure}
    \centering
    \includegraphics[width=0.5\linewidth]{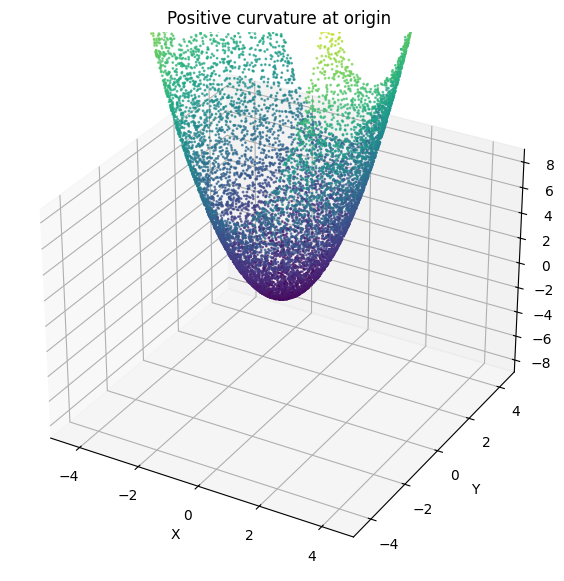}
    \caption{Points drawn from an elliptic paraboloid. The curvature at the origin is positive.}
    \label{fig:EP}
\end{figure}

\begin{figure}
    \centering
    \includegraphics[width=0.5\linewidth]{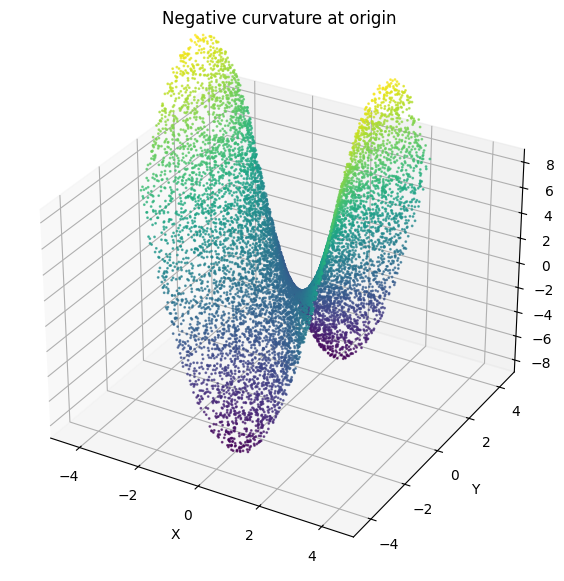}
    \caption{Points drawn from a hyperbolic paraboloid. The curvature at the origin is negative.}
    \label{fig:HPara}
\end{figure}

\begin{figure}
    \centering
    \includegraphics[width=0.5\linewidth]{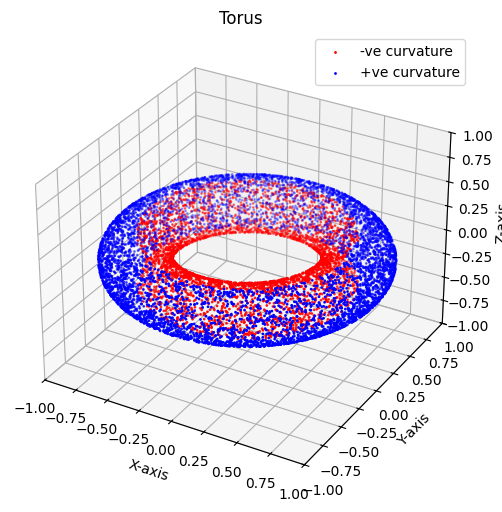}
    \caption{Sample from a torus. Red points having negative Gauss curvature, blue points have positive curvature.}
    \label{fig:toruspn}
\end{figure}

We investigate the effect of curvature on two pointwise dimension estimators, \lpca and \mle using a selection of paraboloids as well as the standard embedding of a torus in $\mathbb{R}^3$ as illustrative examples. These datasets are shown in \Cref{fig:EP,fig:HPara,fig:toruspn}.

We consider a family of elliptic and hyperbolic paraboloids given by the equations $2 x^2 \pm \frac{y^2}{b^2} = z$.
One principal curvature is fixed, while the second varies.
We estimate the pointwise dimension at the point $(0,0,0)$. 
This is repeated 1500 times for each surface (sampled uniformly with the same density, so that when $b=1$ we sample 10000 points).
The dimension is estimated using \lpca with a range of nearest neighbour values (20-165 increasing in 5's to give us 30 values of $k$).
We count how many values of $k$ give an estimate of 2 and also the largest value of $k$ that gives a dimension 2 estimate. 

\begin{figure}
    \centering
    \begin{subfigure}{0.45\textwidth}
        \centering
        \includegraphics[width=\linewidth]{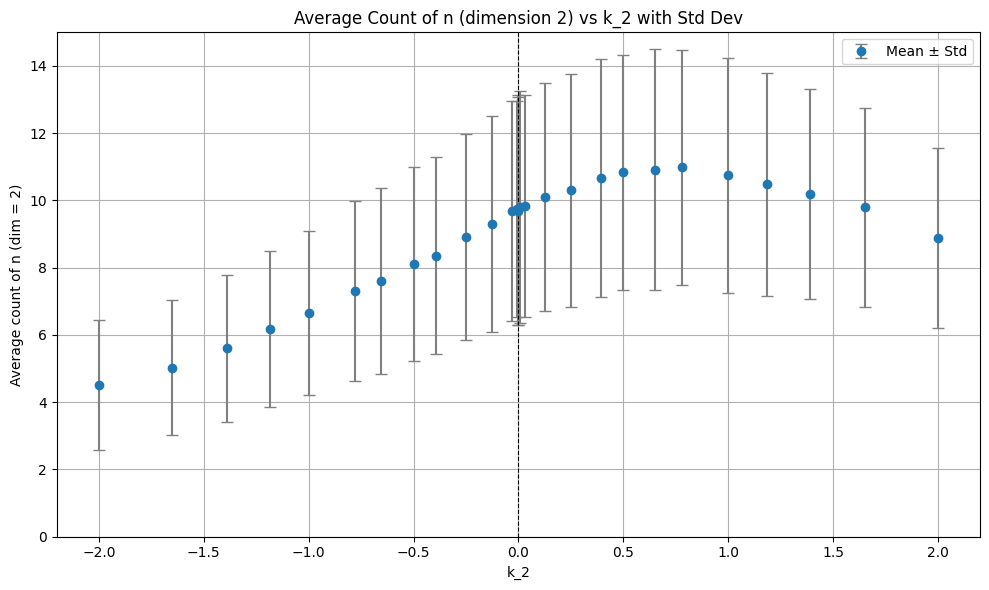}
        \caption{Number of choices of $k$ giving an estimate of 2}
        \label{fig:first}
    \end{subfigure}
    \hfill
    \begin{subfigure}{0.45\textwidth}
        \centering
        \includegraphics[width=\linewidth]{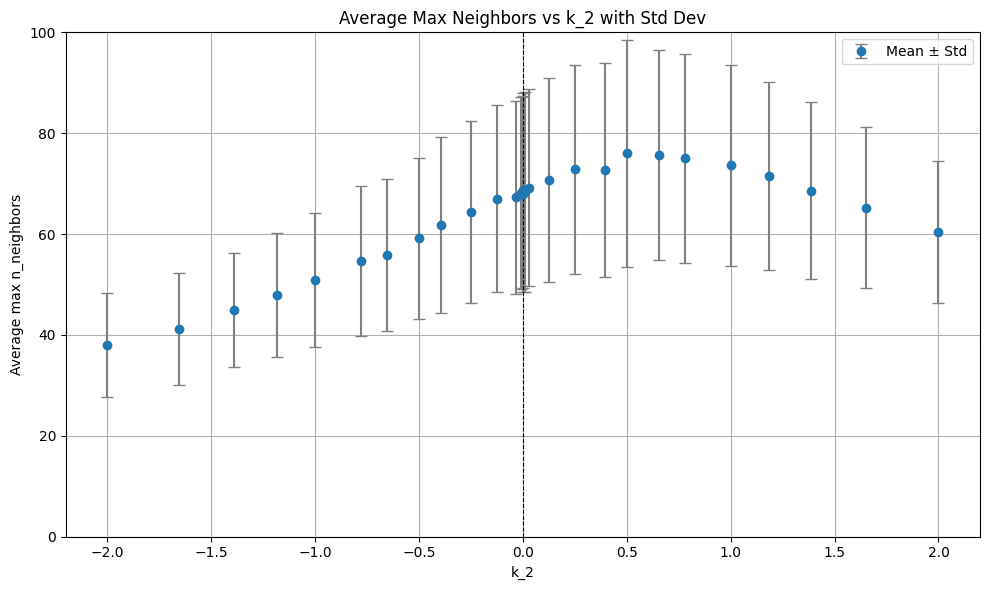}
        \caption{Largest choice of $k$ giving an estimate of 2}
        \label{fig:second1}
    \end{subfigure}
    \caption{Both figures show the same trend. As curvature increases from negative to positive, the number of choices of $k$ and the largest choice of $k$ giving an estimate of 2 both increase. This trend continues past Gauss curvature 0. At $\kappa_2 \simeq 0.6$ this trend reverses. In the case of the torus below, $\kappa_2 \leq 1$ making this reversal hard to detect.}
    \label{fig:both1}
\end{figure}

In \Cref{fig:both1} we observe a clear improvement (more $k$ values giving dimension 2 and larger values giving dimension 2) as the Gauss curvature increases from being negative to being zero. 
This improvement continues until the second principal curvature is $\kappa_2 \simeq 0.6$. 
This suggests that for \lpca using ``FO'' and $\alpha=0.05$, negative curvature creates an upward bias. 
If $\kappa_1$  has a sign, then performance is best when $\kappa_2 \simeq \frac{\kappa_1}{8}$, rather than $\kappa_2 = 0$. 
It would be informative to see if this relationship holds for different values of $\kappa_1$. 
We must note that the standard deviations are large but the trend is very clear and visible in both statistics. 

\begin{figure}
    \centering
\includegraphics[width=0.8\linewidth]{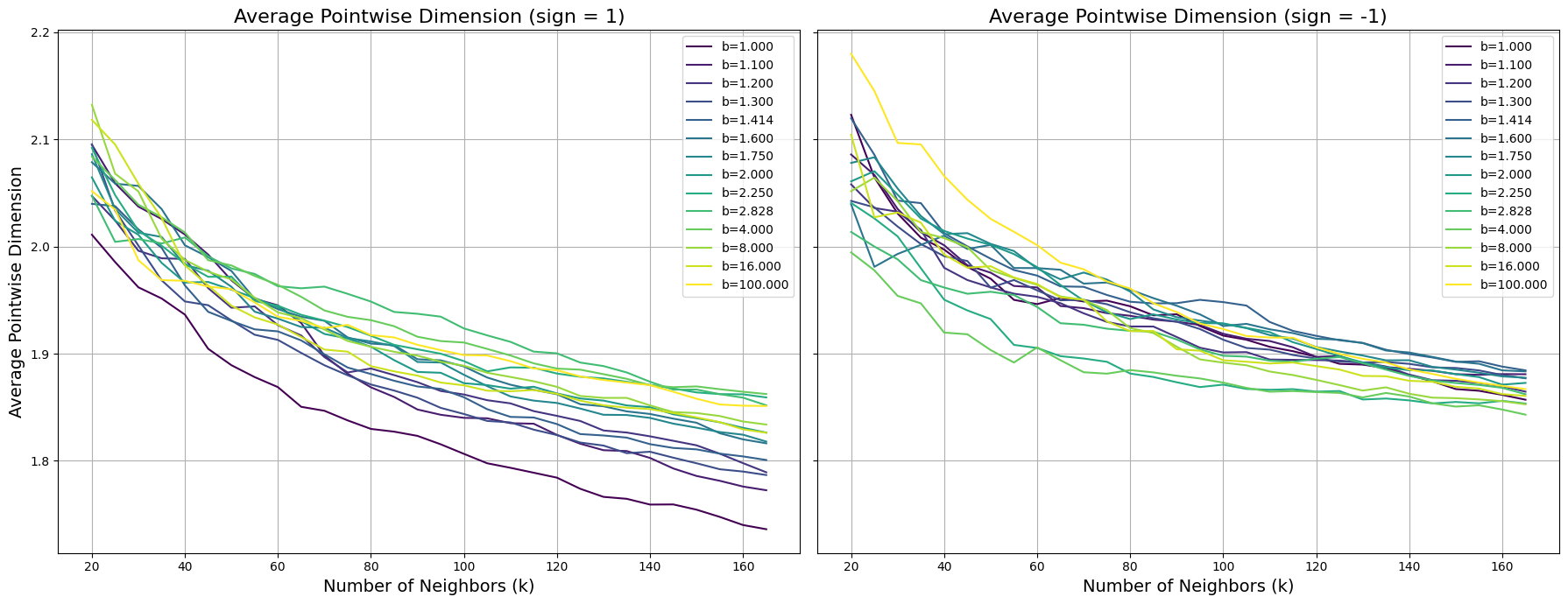}
    \caption{Average pointwise dimension at (0,0,0) for \mle for different for data from the positively curved $2x^2 + \frac{y^2}{b^2} = z$ on the left and from the negatively curved $2x^2 - \frac{y^2}{b^2} = z$ on the right. The same trend appears in both, that as $k$ is increased the pointwise estimates decrease.}
    \label{fig:mlecurv}
\end{figure}

On the other hand, for \mle the results are quite different. The largest value of $k$ considered always gives a dimension estimate of 2. The amount of values of $k$ yielding dimension 2 estimates on average is 27.5 to 29 out of 30. 
We study the effect of curvature here by instead averaging the pointwise dimension of $(0,0,0)$ over 150 runs for each $k$. This is shown in \Cref{fig:mlecurv}.

\begin{figure}
    \centering
    \includegraphics[width=0.8\linewidth]{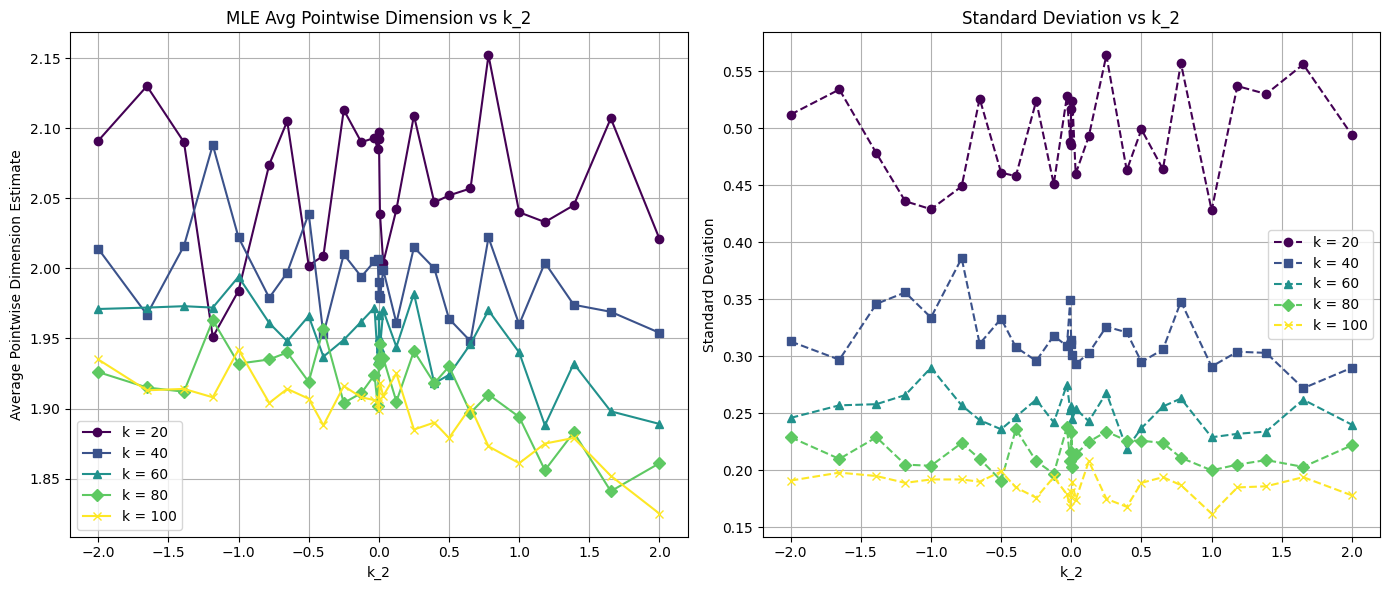}
    \caption{For \mle there is a trend to underestimate as either $\kappa_2$ or the neighbourhood size $k$ is increased. The standard deviation stays reasonably constant for changes of $\kappa_2$, while it decreases with $k$.}
    \label{fig:curvmle}
\end{figure}

\begin{figure}
    \centering
    \includegraphics[width=0.8\linewidth]{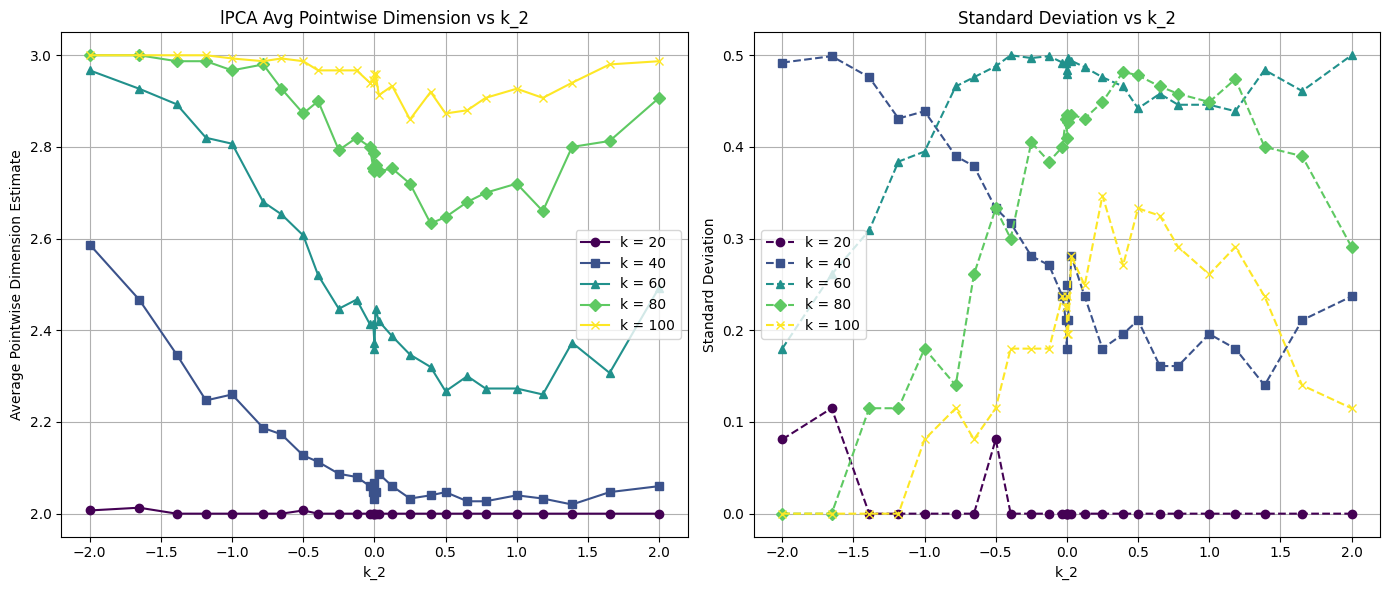}
    \caption{For \lpca we recover the same trend in the estimate with changes in $\kappa_2$ which we saw in Figure \ref{fig:both1} for $k\geq 40$. We also see the high standard deviations in the estimate which occurs as we are only considering one point for each dataset.}
    \label{curvlpca}
\end{figure}

In \Cref{fig:curvmle,curvlpca} we plot the average pointwise dimension estimate for a fixed $k$ against $\kappa_2$. We find that for $\mle$, for all $k$, there is a trend to slightly underestimate the dimension as curvature increases. As $k$ increases, the estimate also decreases, as does the standard deviation which is not dependent on curvature. However, for $\lpca$ we see that for increasing values of $k$ the estimates get worse, and for a fixed $k$ we see the same trend as in figure \Cref{fig:both1}, that the best estimates are made at a positive curvature. For $\lpca$ the standard deviations much more than for $\mle$.

\begin{figure}
    \centering
    \begin{subfigure}{0.4\textwidth}
        \centering
        \includegraphics[width=\linewidth]{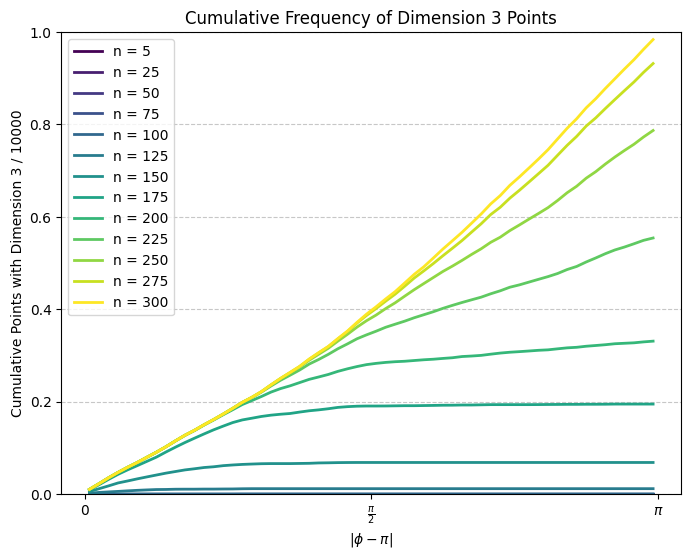}
        \caption{\lpca}
        \label{fig:first}
    \end{subfigure}
    \hfill
    \begin{subfigure}{0.4\textwidth}
        \centering
        \includegraphics[width=\linewidth]{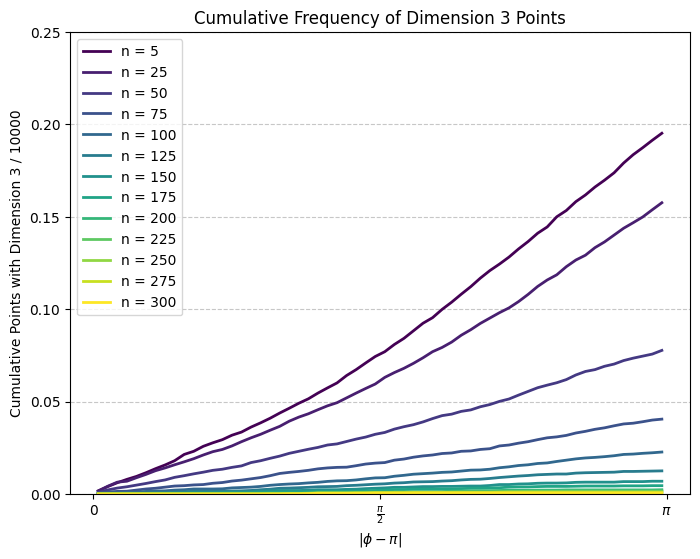}
        \caption{\mle}
        \label{fig:second}
    \end{subfigure}
    \caption{Using a torus embedded in $\bbr^3$ as an example, we investigate the tendency to overestimate dimension as the local geometry is varied. \lpca and \mle exhibit divergent responses. The $x$-axis represents the position of points on the torus, with $0$ representing the innermost latitude (negative curvature), $\pi/2$ the top and bottom of the torus (flat), and $\pi$ the outermost latitude (positive curvature). 
    Larger neighbourhood size $n$ results in overestimation by \lpca, but better accuracy for \mle.
    For \lpca this phenomenon is first apparent in negative curvature and, as $n$ grows, it spreads to the positively curved region.
    In the worst case \mle performs significantly better than \lpca. 
    The figure also emphasises the great importance of hyperparameter choice.}
    \label{fig:both}
\end{figure}

 To study the impact of curvature on the torus, we sampled 10000 points uniformly from the torus in $\mathbb{R}^3$. 
 The degree of overestimation of each estimator is measured by counting the number of points with pointwise estimated dimension 3 (rounded to the nearest integer). 
 To examine how the estimate varies with curvature, we plot the cumulative distribution of overestimated points against $|\phi - \pi|$, where $\phi$ is the angle of the inner circle from $0$ to $2 \pi$.
 In these co-ordinates, $|\phi - \pi|=0$ corresponds to the inside of the torus (negative curvature), $\pi/2$ is the top and bottom of the torus (zero curvature) and $\pi$ represents the outside of the torus (positive curvature. 
 In particular,  we observe that for \lpca, as $k$ increases and the neighbourhood captures more curvature, that pointwise dimension estimates of 3 appear on the inside of the torus (points of negative curvature) and decrease in frequency towards the outside of the torus. 
 This is captured in the decrease in gradient in \Cref{fig:both} (a).
 
 However, the \mle estimator does not show this distinction. 
 There is an even spread of points with overestimated dimension regardless of the curvature (note the gradient is increasing in the figure as there are more points on the outer part of the torus).
 As $k$ increases and neighbourhoods become larger, and less like flat discs, we in fact see a reduction in the number of overestimates.

One potential explanation for \mle's outperformance of \lpca at this task is that \lpca, by design, seeks out an affine subspace which approximates the manifold. 
The better the manifold is approximated by a linear embedding, the better \lpca will perform, and curvature is directly detrimental to this. 
This suggests that, in general, any tangential estimator is likely to be vulnerable to the effects of curvature.
\mle, on the other hand, depends only on radial distances from a single sampled point. Curvature has a tendency to reduce those distances (regardless of the sign of curvature). This may explain why the precise nature of the curvature at different points does not seems to have a significant impact on the results. Again, it appears likely that any estimator based on radial distances, as most of the parametric estimators are, will be more robust to curvature than tangential estimators.

\section{Conclusion and Future Work}
Our survey illustrates the difficulty in designing good dimension estimators that are robust to noise and accommodate the geometries of different spaces. A practical problem is the appropriate choice of hyperparameters, which can significantly bias the dimension estimate. There is no single estimator or set of hyperparameters that can perform well across all settings. 
Therefore, we recommend that practitioners use a range of estimators and hyperparameters to acknowledge the effect of prior assumptions on the inferred dimension. 
Dimension estimators should also be used alongside other tools, such as curvature and topological summaries, to contextualise the extent to which the dimension estimates are reliable. 
We believe future research on techniques to guide the choice of estimators and hyperparameters would be extremely valuable.

Where limited data is an issue, we find that global estimators are generally more suitable. Local estimators need more samples, as each neighbourhood must have a sufficient number of sampled points.
However, global estimators are more susceptible to randomness in sampling, tending to have higher variance.

Our classification of estimators into broad families -- tangential, parametric and those using topological and metric invariants -- demonstrates that there are a variety of perspectives from which dimension estimation can be approached.
We find that the best estimators tend to lie in the parametric family.
Persistent homology provides a reasonably successful estimator from a topological perspective.

A tendency towards underestimation for datasets of higher dimensions is confirmed. 
However, we caution against generalisation. Experiments on data drawn from $SO(n)$ often demonstrate overestimation.
We believe the cause of underestimations comes principally from concentration of measure near the boundary and from finite size effects.
Empirical corrections attempt to reverse this underestimation tendency but, if the dataset under study is not similar to the ones used to calculate the correction factor, this can lead to large errors.

The range of datasets used allowed us to assess the estimators over a wide variety of desirable criteria. 
We find that no estimators provide a satisfactory level of performance on non-linear datasets with dimension above six.
However, \ess performs strongly on all other criteria.
We strongly recommend that future researchers include \ess as a comparison method to benchmark performance.

Our additions to the methods of \scikitdimension: \geomle, \gride, \cdim, \wodcap, Camastra \& Vinciarelli's extension to the Grassberger--Procaccia \corrint algorithm, the packing-number based estimator, and the magnitude and \phzero estimators, expand the methods practitioners can draw from in an easily accessible place. We have also added new functionality for \lpca and \mle, so that users can choose $\epsilon$-neighbourhoods in addition to $\knn$ neighbourhoods, giving greater freedom to practitioners. We have also added a probabilistic thresholding method for \pca~\cite{Minka2001AutomaticPCA}.

Given the increasing use of \phzero and $\mathsf{mag}$ dimension estimators in the applied topology community, especially on machine learning problems~\cite{Andreeva2023MetricNetworks,Limbeck2025MetricRepresentations}, we give a more detailed investigation and benchmarking of performance on these estimators. While \phzero performs comparably to other estimators investigated here, the estimation of magnitude dimension suffers from finite size effects, which is detailed in \Cref{app:mag}.

We demonstrate that the choice of hyperparameters is crucial and that is is essential for practitioners to understand the role they play for each estimator. Our recommendation is that a range of hyperparameters be used to build confidence in the result. We also recommend developers of dimension estimators consider theoretical limits that hyperparameter choices on the range of dimension estimates it can obtain, as we have identified a throttling phenomenon that results from poor choices of hyperparameters across several estimators.

Gaussian noise presents an issue for most estimators. However, local estimators can overcome certain types of outliers through aggregation.

We provide evidence that the role of curvature plays an important role in estimations. We confirming the known negative effects of curvature on \lpca, but find that, for at least one aggregation method, slightly positively curved surfaces can be easier to estimate that those with zero Gauss curvature. The effects of curvature \mle are much lower that the effect of \lpca, which may generalise to a statement that parametric estimators are more robust to curvature than tangential estimators.

Areas that require major progress within this field are estimator performance on non-linear manifolds and high dimensional manifolds, as well as the development of practical ways to guide hyperparameter choice. Alternatively, it would be worth investigating adaptive methods, automating the choice of hyperparameters using features of the input data. 
The limited and varied results shown here on the role of curvature clearly justify a systematic approach to determining to what extent curvature has an impact on dimension estimators.

\section*{Acknowledgements}
JB: This work was supported by the Additional Funding Programme for Mathematical Sciences, delivered by EPSRC (EP/V521917/1) and the Heilbronn Institute for Mathematical Research. PD and JM: This work was supported by Dioscuri program initiated by the Max Planck Society, jointly managed with the National Science Centre (Poland), and mutually funded by the Polish Ministry of Science and Higher Education and the
German Federal Ministry of Education and Research. JH and KMY: This work was supported by a UKRI Future Leaders Fellowship [grant number MR/W01176X/1; PI J Harvey]. This material is based in part upon work supported by the National Science Foundation under Grant No. DMS-1928930, while JH was in residence at the Simons Laufer Mathematical Sciences Institute in Berkeley, California, during Fall 2024. 

\clearpage
\bibliography{refs.bib}
\bibliographystyle{plain}

\appendix
\renewcommand{\thetable}{\Alph{section}\arabic{table}}
\renewcommand{\thefigure}{\Alph{section}\arabic{figure}}

\clearpage
\section{Comparison of PH and KNN}
\setcounter{table}{0}
\setcounter{figure}{0}
\label{app:ph_vs_knn}
Since \phzero and \knnestimator are derived from the common theory of Euclidean functionals, and are similar in construction, we highlight a comparison in their performance on the benchmark set of datasets. We first discuss some theoretical advantages of \phzero. The key difference between the two estimators is that \phzero further processes the distance information by using the minimum spanning tree (or zeroth dimensional persistent homology) of the point set. The minimum spanning tree takes the global connectivity into account; in comparison, \knnestimator considers edge lengths along the \knn graph, which is a much coarser organisation of the connectivity of the point cloud compared to \phzero. One key benefit of \phzero is the stability property with respect to the perturbation of points conferred to the $\alpha$-weight of the minimum spanning tree~\cite{Skraba2025WassersteinDiagrams}. In addition, the minimum spanning tree of \phzero does not rely on assumptions about a suitable local neighbourhood size, which is required as a hyperparameter in \knnestimator in the construction of the \knn graph. 

In our empirical observations on the benchmark dataset, while the mean estimates of both estimators are comparable, we see that \knnestimator is more susceptible to the randomness of the point sample, and can have much larger variance, especially for higher dimensional datasets. 

To investigate the effects of hyperparameter choice, in \Cref{tab:ph_alpha}, we show how estimates of spheres for different dimensions vary for different choices of $\alpha$. In the main text, \Cref{fig:knn_bm} shows the performance of \knnestimator as $k$ is varied on the benchmark dataset. 

\input{tables/hyperparameters/ph_alpha}

\begin{figure}[h]
    \centering  \includegraphics[width=0.85\linewidth]{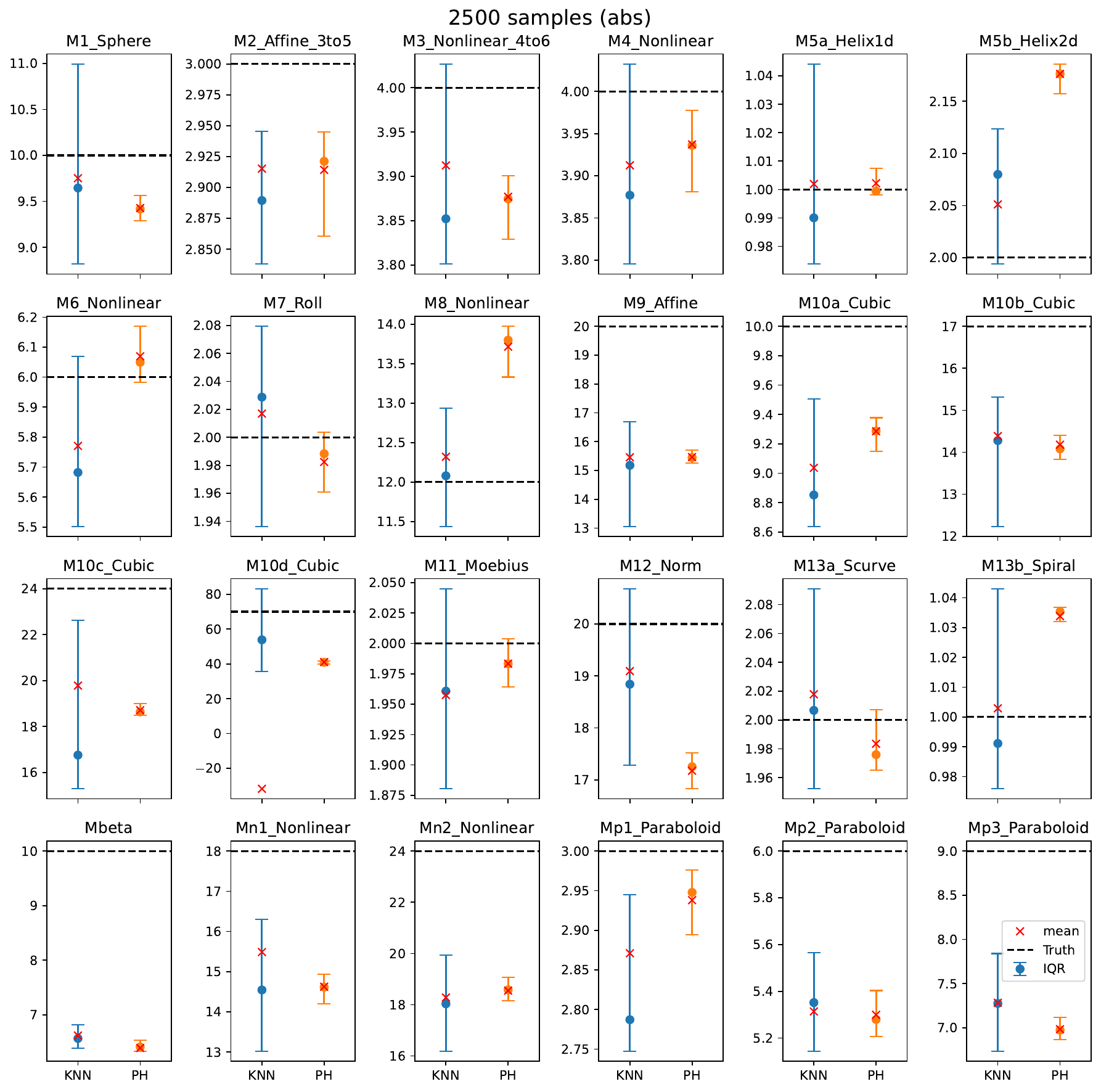}
    \caption{Performance of \phzero and \knnestimator estimators on benchmark manifolds with 2500 samples. The hyperparameters chosen here are fixed across all datasets, with $\alpha = 0.5$ and $k = 1$ for \phzero and \knnestimator respectively. Both choices are the ones that minimises the median (as taken across the benchmark manifolds) absolute and relative error in the mean dimension estimate. The error bars indicate the interquartile range and median dimension estimate. The red cross indicates the mean dimension estimate, and the dashed line indicates the true dimension.  We note that the \knnestimator estimator consistently has greater variance when compared to PH, and can output extreme outliers in high dimensional cases such as M10d Cubic.}
    \label{fig:PH_KNN_abs_2500}
\end{figure}

\begin{figure}
    \centering  \includegraphics[width=0.85\linewidth]{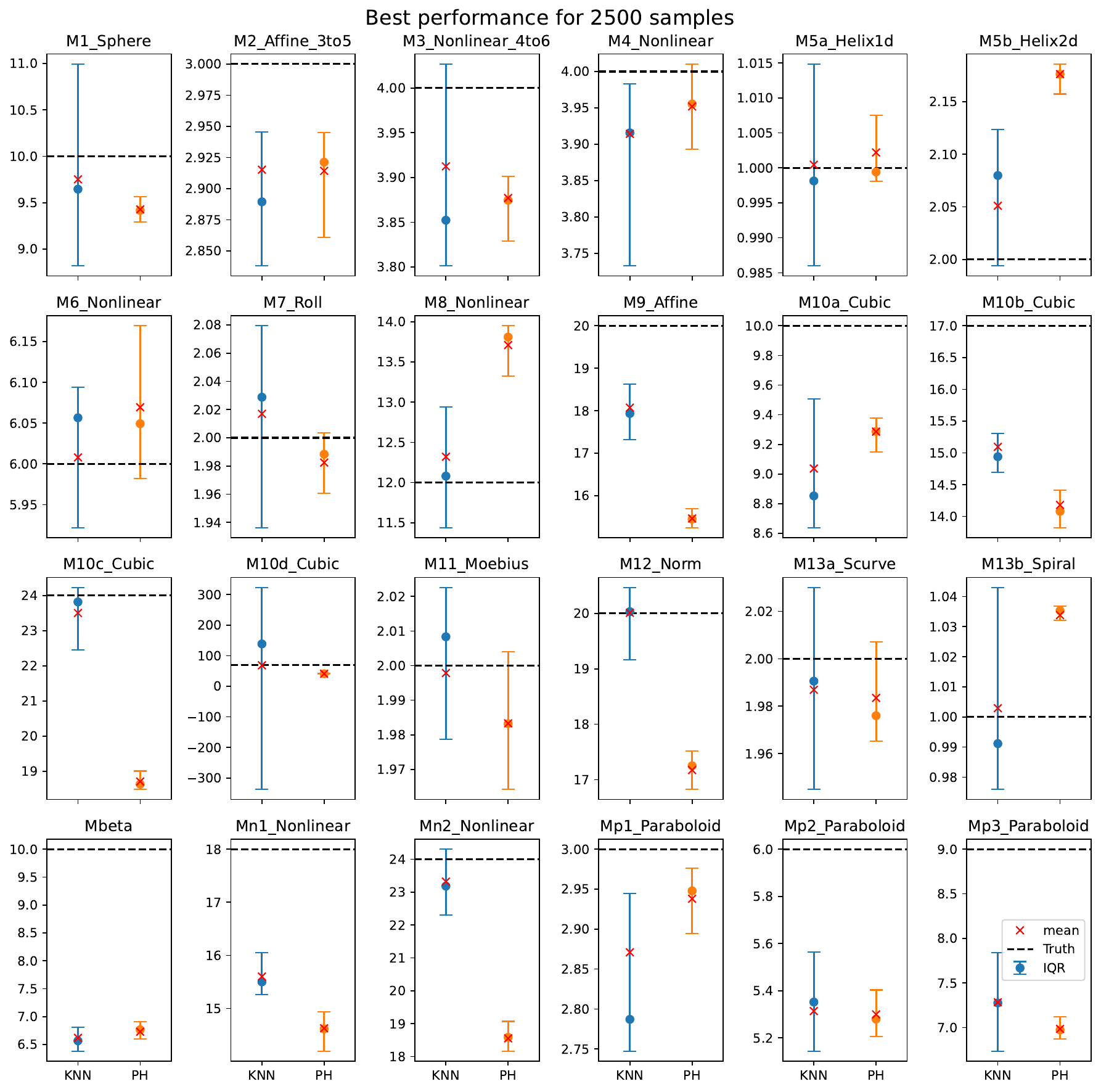}
    \caption{Performance of \phzero and \knnestimator estimators on benchmark manifolds with 2500 samples.  The hyperparameters chosen here are different across the datasets, and are chosen to be the the ones minimises the difference between the mean dimension estimate and ground truth. The error bars indicate the interquartile range and median dimension estimate. The red cross indicates the mean dimension estimate, and the dashed line indicates the true dimension.  We note that the \knnestimator estimator consistently has greater variance when compared to \phzero, though there it is often more accurate given the right hyperparameter choice. }
    \label{fig:PH_KNN_best_2500}
\end{figure}

\clearpage
\section{Finite size issues of magnitude dimension}
\setcounter{table}{0}
\setcounter{figure}{0}
\label{app:mag}

 Focussing on the practicalities, finite size issues can affect the inference of magnitude dimension from finite samples, since  $\magdim(\bbx) =0$. Because $|t\mathbb{X}| \to |\mathbb{X}|$ as $t \to \infty$, the slope of the line approaches zero. In practice, while the range of $t$ over which we fit the curve must be large enough to approximate the limit, it cannot be so large that the finite size effect occurs. If there are too few points sampled from $\mathbb{X}$, then the finite size effect takes over before $t$ can be large enough for the asymptotic behaviour to emerge. This means the number of points may need to be quite large for the dimension to be read off the empirical curve $\log|t\mathbb{X}|$.

\begin{figure}[h]
    \centering
    \includegraphics[width=0.65\linewidth]{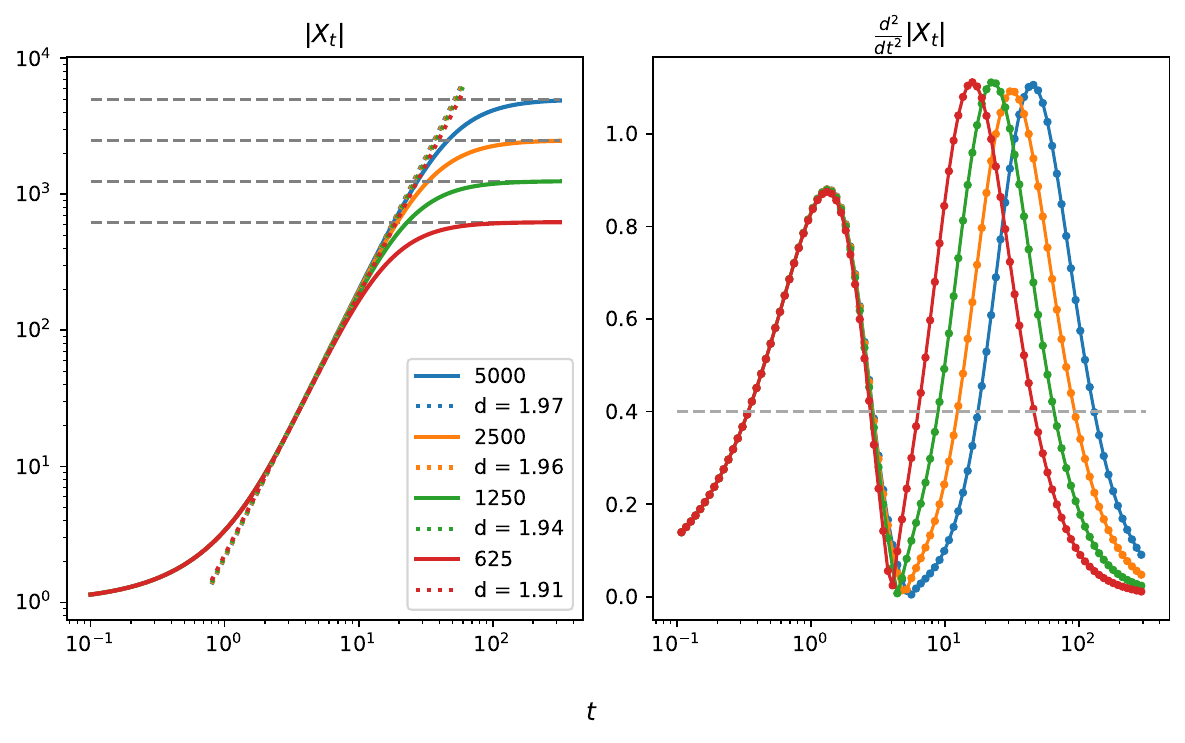}
    \caption{Magnitude functions of random samples of $N = 625, 1250, 2500, 5000$ points on $\mathbb{S}^2 \subset \mathbb{R}^3$. We observe that, as $N$ increases, the finite cap on $|t\bbx|$ arrives at a larger value of $t$, and a larger part of the linear region of the curve is preserved. We use the slope of the linear part of the curve as the magnitude dimension estimate.  On the right we plot the magnitude of the second derivative of the curve, approximated by finite difference. The linear portion of the curve is selected to be the part of the curve whose second derivative lies below the threshold value indicated in the right hand panel.  }
\label{fig:mag_sphere_2}
\end{figure}

We demonstrate this in an experiment with uniform random samples from $S^d \subset \bbr^{d+1}$. For example, for $d = 2$,  \Cref{fig:mag_sphere_2} displays $\log t$ vs $\log|t\bbx|$ for $|\bbx| = 625, 1250, 2500,5000$. For small $t$, the curves are identical for different number of samples, yet as $t$ increases and $|t\bbx|$ grows, the finite size effect takes hold and the curves plateau at $|t\bbx| \to |\bbx|$. In \Cref{tab:mag}, we show the magnitude dimension estimates for $d = 2,4,8,16$ and varying number of samples. Even for modest dimensions and high number of samples, the dimension estimates are far below the actual dimension, as the finite size effect prevents the emergence of asymptotic growth of the magnitude curve. The magnitude curves for higher dimensions are illustrated in \Cref{app:mag}. 

\begin{table}[h]
\centering
\begin{tabular}{|l|llll|}
\hline
n               & 625  & 1250 & 2500 & 5000 \\ \hline
$S^2$  & 1.91 & 1.94 & 1.96 & 1.97 \\
$S^4$  & 2.92 & 3.15 & 3.33 & 3.47 \\
$S^{8}$ & 3.55 & 3.97 & 4.37 & 4.75 \\ 
$S^{16}$ & 3.79 & 4.33 & 4.82 & 5.34 \\ \hline
\end{tabular}
\vspace{1em}
\caption{Magnitude dimension estimates of the dimension of spheres $S^m$ where $m = 2,4,8, 16$, given $N$ uniform i.i.d. samples. We note that even for moderately high dimensions, the magnitude dimension estimator can fail to recover the intrinsic dimension of the unit sphere even for $N = 5000$, where other estimators succeed. }
\label{tab:mag}
\end{table}

\begin{figure}[h]
    \centering
    \includegraphics[width=0.65\linewidth]{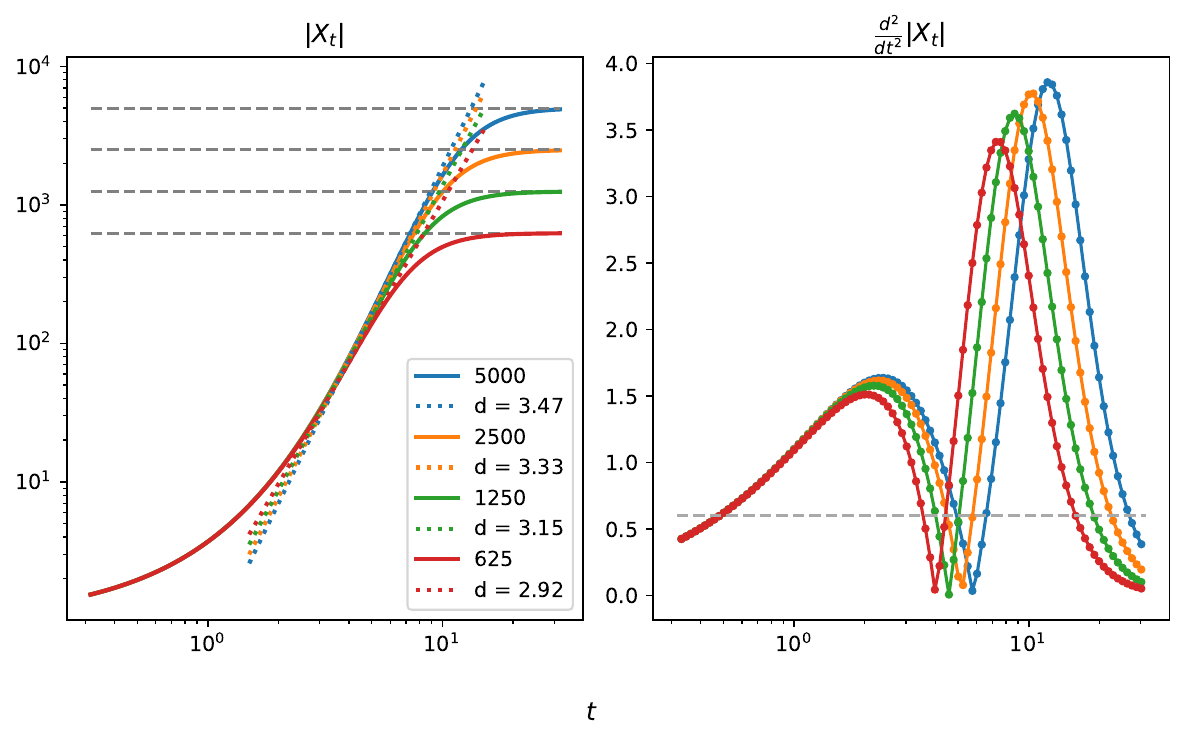}
    \caption{Magnitude functions of random samples of $N = 625, 1250, 2500, 5000$ points on $\mathbb{S}^4 \subset \mathbb{R}^5$. We observe that, as $N$ increases, the finite cap on $|t\bbx|$ arrives at a larger value of $t$, and a larger part of the linear region of the curve is preserved. We use the slope of the linear part of the curve as the magnitude dimension estimate.  On the right we plot the magnitude of the second derivative of the curve, approximated by finite difference. The linear portion of the curve is selected to be the part of the curve whose second derivative lies below the threshold value indicated in the right hand panel.  }
    \label{fig:mag_sphere_4}
\end{figure}
\begin{figure}[h]
    \centering
    \includegraphics[width=0.65\linewidth]{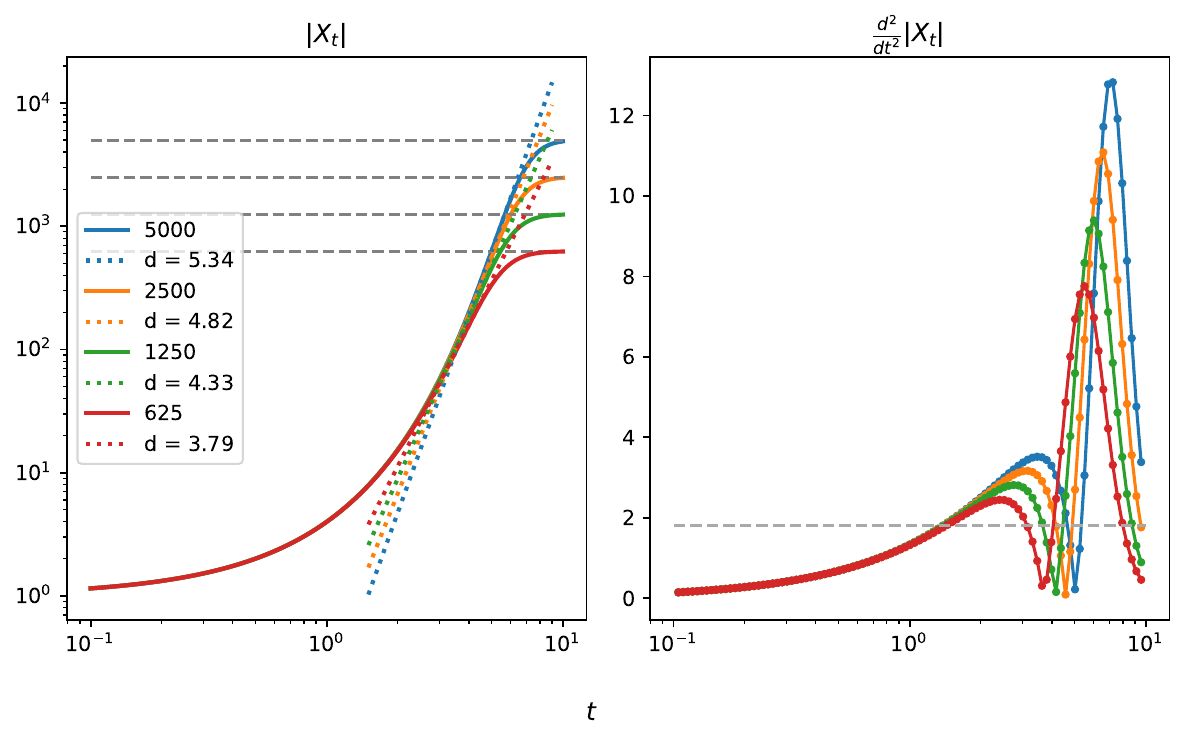}
    \caption{Magnitude functions of random samples of $N = 625, 1250, 2500, 5000$ points on $\mathbb{S}^{16} \subset \mathbb{R}^{17}$. We observe that, as $N$ increases, the finite cap on $|t\bbx|$ arrives at a larger value of $t$, and a larger part of the linear region of the curve is preserved. We use the slope of the linear part of the curve as the magnitude dimension estimate.  On the right we plot the magnitude of the second derivative of the curve, approximated by finite difference. The linear portion of the curve is selected to be the part of the curve whose second derivative lies below the threshold value indicated in the right hand panel.  }
    \label{fig:mag_sphere_16
}
\end{figure}

\clearpage
\section{Performance of Estimators on Benchmark Datasets} \label{app:bmtables}
\setcounter{table}{0}
\setcounter{figure}{0}

We assess the estimators with the following experimental procedure.
Let $\mathcal M$ be the set of benchmark manifolds, $\mathcal E$ the set of estimators, and $\mathcal H_E$ the set of hyperparameters for some estimator $E \in \mathcal E$.
For each triplet $(M,E,H)$ representing a manifold, an estimator and a hyperparameter choice, we evaluated the performance of the estimator over 20 randomly sampled point sets from $M$. We record the empirical mean, which we denote by $\hat d (E, M, H)$, and the standard deviation of the dimension estimates. For each dataset type in the list of benchmark datasets, we varied the number of samples from 625, 1250, 2500, and 5000, examining the performance as the number of points is successively doubled. 

We varied the choice of hyperparameters over a range specified in \Cref{app:hyperpm}.
Across the local estimators we varied the number of nearest neighbours. We aggregate the performance across hyperparameters in the tables in \Cref{app:bmtables}, where we show dimension estimates on three different types of hyperparameters:
\begin{enumerate}
    \item The hyperparameter that minimises the difference between the estimated and intrinsic dimension
    \begin{equation*}
        \hat{d}_\text{best} (E,M) = \min_{H \in \mathcal H_E} |\hat{d}(E, M, H) - d(M)|
    \end{equation*}
    \item The performance of the estimator with a fixed choice of hyperparameter; this is either
    \begin{enumerate}
        \item The hyperparameter that minimises the median \emph{absolute} error across the benchmark manifolds:
    \begin{align*}
    \hat{d}_\text{abs} (E, M) &=  \hat{d} (E, M, H_{\text{abs}}(E))\\
        {H}_{\text{abs}}(E) &= \argmin_{H \in \mathcal H_E} \underset{M \in \mathcal M} {\text{median}} |\hat{d}(E, M, H) - d(M)|
    \end{align*}
    \item The one that minimises the median \emph{relative} error across the benchmark manifolds:
    \begin{align*}
        \hat{d}_\text{rel} (E, M) &=  \hat{d} (E, M, {H}_{\text{rel}}(E))\\
        {H}_{\text{rel}}(E) &= \argmin_{H \in \mathcal H_E} \underset{M \in \mathcal M} {\text{median}} \frac{|\hat{d}(E, M, H) - d(M)|}{d(M)}
    \end{align*}
    \end{enumerate}
\end{enumerate}
The hyperparameter choices ${H}_{\text{abs}}$ and ${H}_{\text{rel}}$ guarantee a reasonable performance of the estimator on most of the datasets in the benchmark. While a choice of hyperparameter might be optimal for a certain dataset, the same hyperparameter may lead to a poor performance on another. The difference between between $\hat{d}_\text{best}$ and  $\hat{d}_\text{abs}$ or $\hat{d}_\text{rel}$ gives an indication of how dependent the estimator is on the hyperparameter choice in order to perform well on a particular dataset.

\setcounter{table}{0}
\setcounter{figure}{0}

\input{tables/lPCA_benchmark}

\input{tables/PHnew}
\input{tables/KNNb_benchmark}

\input{tables/DanCoNEW}
\input{tables/MLE_benchmark}
\input{tables/MINDML_benchmark}
\input{tables/GRIDE_benchmark}

\input{tables/CorrInt_benchmark}
\input{tables/WODCap_benchmark}
\input{tables/TwoNN_benchmark}
\input{tables/TLE_benchmark}
\input{tables/ESS_benchmark}
\input{tables/FISHERS_benchmark}
\input{tables/Cdim_benchmark1}

\clearpage
\section{Hyperparameter choices}
\label{app:hyperpm}
\setcounter{table}{0}
\setcounter{figure}{0}

This appendix contains the hyperparameters for the benchmarking experiments in \Cref{app:bmtables}.

Across all local estimators, and all sampling regimes, we varied between $\epsilon$ and $k$-nearest neighbour neighbourhoods; the $k$-nearest neighbour parameter is varied only between $k \in  \{10,20,40,80\}$. We also varied the $\epsilon$-neighbourhood across four settings. For consistency across datasets, given a point set, the $\epsilon$-parameter is chosen to be the median $k$-nearest neighbour distance of that point set, where  $k \in  \{10,20,40,80\}$ as well.

\begin{table}[h]
    \centering
    \begin{tabular}{| l || l | }\hline 
\lpca &  \\ \hline 
thresholding version & FO($\alpha = 0.05$) \\
 &  Fan($\alpha = 10$, $\beta=0.8$, $P = 0.95$) \\
 & Maxgap \\
 & Ratio ($\alpha = 0.05$) \\
 & Participation ratio \\
 & Kaiser \\
 & broken stick \\
 & Minka \\ \hline 
\end{tabular}
    \caption{\lpca thresholding methods being varied in our benchmark experiments.}
    \label{tab:lpca_hprange}
\end{table}
\begin{table}[h]
    \centering
    \footnotesize
\begin{tabular}{| l || l l | l l |l l |l l |}\hline 
\lpca & 625 &  & 1250 & & 2500 & & 5000 & \\ \hline 
    & med abs  & med rel & med abs  & med rel & med abs  & med rel & med abs  & med rel  \\ \hline 
nbhd type & eps & knn & knn & knn & eps & eps & knn & knn   \\ 
n neighbours & 20 & 40 & 40 & 40 & 80 & 80 & 80 & 80   \\ 
comb & hmean & median & mean & median & median & median & median & median   \\ 
ver & FO & FO & FO & FO & FO & FO & FO & FO   \\ 
\hline 
\end{tabular}
\caption{Hyperparameters for the dimension estimates in \Cref{tab:lPCA_bm}. }
\label{tab:lPCA_bmhp}
\end{table}

\begin{table}[h]
\centering
\begin{tabular}{| l || l | }\hline 
\phzero &  \\ \hline 
alpha &  0.5, 1.0, 1.5,  2.0\\ \hline 
n range min & 0.75 \\ \hline 
n range max & 1.0\\ \hline 
range type & fraction\\ \hline 
subsamples & 10\\ \hline 
nsteps & 10\\ \hline 
n neighbours & 30\\ \hline 
\end{tabular}
\caption{\phzero hyperparameter range in benchmark experiments.}
\end{table}
\begin{table}[h]
    \centering
    \begin{tabular}{| l || l l | l l |l l |l l |}\hline 
\phzero & 625 &  & 1250 & & 2500 & & 5000 & \\ \hline 
    & med abs  & med rel & med abs  & med rel & med abs  & med rel & med abs  & med rel  \\ \hline 
alpha & 0.5 & 0.5 & 0.5 & 0.5 & 0.5 & 0.5 & 0.5 & 0.5   \\ 
\hline 
\end{tabular}
\caption{\phzero hyperparameters in \Cref{tab:ph_bm}.}
    \label{tab:ph_bmhp}
\end{table}

\begin{table}[h]
\centering
\begin{tabular}{| l || l | }\hline 
\knnestimator & \\ \hline 
k & 1,\ldots,80\\ \hline 
n range min & 0.75\\ \hline 
n range max & 1.0\\ \hline 
range type & fraction\\ \hline 
subsamples & 10\\ \hline 
nsteps & 10\\ \hline 
\end{tabular}
\caption{\knnestimator hyperparameter range in benchmark experiments. Note that unlike other estimators with nearest neighbour $k$ hyperparameter, in the implementation it incurs an almost trivial cost to compute all $k$'s up to a specified maximum $k$.  }
\end{table}

\begin{table}[h]
    \centering
    \begin{tabular}{| l || l l | l l |l l |l l |}\hline 
\knnestimator & 625 &  & 1250 & & 2500 & & 5000 & \\ \hline 
    & med abs  & med rel & med abs  & med rel & med abs  & med rel & med abs  & med rel  \\ \hline 
k & 1 & 3 & 1 & 1 & 1 & 1 & 4 & 2  \\ 
\hline 
\end{tabular}
\caption{Hyperparameters for estimates in \Cref{tab:knn_bm}.}
\label{tab:knn_bmhp}
\end{table}

\begin{table}[h]
    \centering
    \begin{tabular}{| l || l | }\hline 
\gride &  \\ \hline 
n1 & 1, 2, 4, 8, 16, 32\\ \hline 
multiplier & 2, 3, 5\\ \hline 
d0 & 1\\ \hline 
d1 & 150\\ \hline 
\end{tabular}
\caption{\gride hyperparameter ranges in our experiments.}
\end{table}

\begin{table}[h]
\centering
\begin{tabular}{| l || l l | l l |l l |l l |}\hline 
\gride & 625 &  & 1250 & & 2500 & & 5000 & \\ \hline 
    & med abs  & med rel & med abs  & med rel & med abs  & med rel & med abs  & med rel  \\ \hline 
n1 & 1 & 1 & 1 & 1 & 1 & 1 & 1 & 1   \\ 
multiplier & 2 & 2 & 2 & 2 & 2 & 2 & 2 & 2   \\  
\hline 
\end{tabular}
\caption{Hyperparameters for the estimates in \Cref{tab:gride_bm}. Note that these choices make the estimator identical to \mle with input from  two nearest neighbours.  }
\label{tab:gride_bmhp}
\end{table}

\begin{table}
\centering
\begin{tabular}{| l || l | }\hline 
\corrint & \\ \hline 
k1 & 2, 4, 6, 8, 10, 12\\ \hline 
k2 & 12, 14, 16, 18, 20\\ \hline

\end{tabular}
\caption{\corrint hyperparameters range}
\end{table}

\begin{table}
\centering
\begin{tabular}{| l || l l | l l | l l | l l |}\hline 
\corrint & 625 & & 1250 & & 2500 & & 5000 & \\ \hline 
 & med abs & med rel & med abs & med rel & med abs & med rel & med abs & med rel  \\ \hline 
k1 & 2 & 10 & 2 & 12 & 2 & 2 & 2 & 2 \\ 
k2 & 12 & 12 & 12 & 18 & 12 & 12 & 12 & 12 \\ 
\hline 
\end{tabular}
\caption{Hyperparameters for the estimates in \Cref{tab:corrint_bm} }
\end{table}

\begin{table}
\centering

\begin{tabular}{| l || l | }\hline 
\danco & \\ \hline
k & 10, 20, 25, 30, 35, 40, 80\\ \hline 
ver & DANCo\\ \hline 
\end{tabular}

\caption{\danco hyperparameters range}

\end{table}

\begin{table}

\centering
\begin{tabular}{| l || l l | l l | l l | l l |}\hline 
\danco & 625 & & 1250 & & 2500 & & 5000 & \\ \hline 
 & med abs & med rel & med abs & med rel & med abs & med rel & med abs & med rel  \\ \hline 
k & 20 & 40 & 20 & 30 & 20 & 20 & 20 & 30 \\ 
ver & DANCo & DANCo & DANCo & DANCo & DANCo & DANCo & DANCo & DANCo \\ 
\hline 
\end{tabular}
\caption{Hyperparameters for the estimates in \Cref{tab:danco_bm}}
\end{table}

\begin{table}
\centering
\begin{tabular}{| l || l | }\hline 
\ess & \\ \hline
ver & a, b\\ \hline 
d & 1\\ \hline 
\end{tabular}
\caption{\ess hyperparameters range}
\end{table}

\begin{table}
\centering
\begin{tabular}{| l || l l | l l | l l | l l |}\hline 
\ess & 625 & & 1250 & & 2500 & & 5000 & \\ \hline 
 & med abs & med rel & med abs & med rel & med abs & med rel & med abs & med rel  \\ \hline 
nbnd type & knn & knn & knn & knn & knn & knn & knn & knn \\ 
n neighbours & 40 & 40 & 40 & 40 & 40 & 80 & 40 & 80 \\ 
ver & b & a & b & b & b & b & a & b \\ 
d & 1 & 1 & 1 & 1 & 1 & 1 & 1 & 1 \\ 
comb & hmean & median & hmean & median & hmean & median & mean & median \\ 
\hline 
\end{tabular}
\caption{Hyperparameters for the estimates in \Cref{tab:ess_bm}}
\end{table}

\begin{table}
\centering
\begin{tabular}{| l || l | }\hline 
\fishers & \\ \hline 
conditional number & 5, 6, 8, 10, 12, 13\\ \hline 
project on sphere & 0, 1\\ \hline 
limit maxdim & 0, 1\\ \hline 
\end{tabular}
\caption{\fishers hyperparameters range}
\end{table}

\begin{table}
\centering

\begin{tabular}{| l || l l | l l | l l | l l |}\hline 
\fishers & 625 & & 1250 & & 2500 & & 5000 & \\ \hline 
 & med abs & med rel & med abs & med rel & med abs & med rel & med abs & med rel  \\ \hline 
conditional number & 5 & 5 & 5 & 5 & 5 & 5 & 5 & 5 \\ 
project on sphere & 0 & 0 & 0 & 0 & 0 & 0 & 0 & 0 \\ 
produce plots & 0 & 0 & 0 & 0 & 0 & 0 & 0 & 0 \\ 
verbose & 0 & 0 & 0 & 0 & 0 & 0 & 0 & 0 \\ 
limit maxdim & 0 & 0 & 0 & 0 & 0 & 0 & 0 & 0 \\ 
\hline 
\end{tabular}
\caption{Hyperparameters for the estimates in \Cref{tab:fishers_bm}}
\end{table}

\begin{table}
\centering
\begin{tabular}{| l || l | }\hline 
\mindml & \\ \hline 
k & 1, 2, 3, 4, 5, 10, 15, 20\\ \hline 
ver & MLk, MLi\\ \hline 
D & 10, 20, 30, 40, 50, 60, 70\\ \hline 
\end{tabular}
\caption{\mindml hyperparameters range}
\end{table}

\begin{table}
\centering
\begin{tabular}{| l || l l | l l | l l | l l |}\hline 
\mindml & 625 & & 1250 & & 2500 & & 5000 & \\ \hline 
 & med abs & med rel & med abs & med rel & med abs & med rel & med abs & med rel  \\ \hline 
k & 1 & 1 & 1 & 1 & 1 & 1 & 1 & 1 \\ 
ver & MLk & MLi & MLk & MLk & MLk & MLk & MLi & MLk \\ 
D & 20 & 20 & 10 & 20 & 30 & 10 & 10 & 10 \\ 
\hline 
\end{tabular}
\caption{Hyperparameters for the estimates in \Cref{tab:mind_bm} }
\end{table}

\begin{table}
\centering
\begin{tabular}{| l || l | }\hline 
\mle & \\ \hline
k & 10, 20, 40, 80\\ \hline 
comb & mean, hmean, median\\ \hline  
\end{tabular}
\caption{\mle hyperparameters range}
\end{table}

\begin{table}
\centering
\begin{tabular}{| l || l l | l l | l l | l l |}\hline 
\mle & 625 & & 1250 & & 2500 & & 5000 & \\ \hline 
 & med abs & med rel & med abs & med rel & med abs & med rel & med abs & med rel  \\ \hline 
nbnd type & knn & knn & knn & knn & knn & knn & knn & knn \\ 
n neighbours & 10 & 10 & 10 & 10 & 10 & 10 & 10 & 10 \\ 
comb & median & mean & median & mean & median & mean & median & median \\ 
\hline 
\end{tabular}
\caption{Hyperparameters for the estimates in \Cref{tab:mle_bm}}
\end{table}

\begin{table}
\centering
\begin{tabular}{| l || l | }\hline 
\tle & \\ \hline
n neighbours & 10, 20, 40, 80 \\  \hline 
epsilon & 1e-06, 2e-06, 3e-06, 5e-06\\ \hline 
\end{tabular}
\caption{\tle hyperparameters range}
\end{table}

\begin{table}
\centering
\begin{tabular}{| l || l l | l l | l l | l l |}\hline 
TLE & 625 & & 1250 & & 2500 & & 5000 & \\ \hline 
 & med abs & med rel & med abs & med rel & med abs & med rel & med abs & med rel  \\ \hline 
nbnd type & knn & knn & knn & knn & knn & knn & knn & knn \\ 
n neighbours & 20 & 10 & 10 & 10 & 10 & 20 & 10 & 10 \\ 
epsilon & 5e-06 & 5e-06 & 5e-06 & 5e-06 & 5e-06 & 5e-06 & 5e-06 & 5e-06 \\ 
comb & hmean & mean & hmean & mean & hmean & median & hmean & hmean \\ 
\hline 
\end{tabular}
\caption{Hyperparameters for the estimates in \Cref{tab:tle_bm} }

\end{table}

\begin{table}
\centering
\begin{tabular}{| l || l | }\hline 
\twonn & \\ \hline
discard fraction & 0.05, 0.06, 0.08, 0.09, 0.1, 0.2, 0.3, 0.4, 0.5, 0.6\\ \hline 
\end{tabular}
\caption{\twonn hyperparameters range}
\end{table}

\begin{table}
\centering
\begin{tabular}{| l || l l | l l | l l | l l |}\hline 
\twonn & 625 & & 1250 & & 2500 & & 5000 & \\ \hline 
 & med abs & med rel & med abs & med rel & med abs & med rel & med abs & med rel  \\ \hline 
discard fraction & 0.05 & 0.05 & 0.05 & 0.05 & 0.06 & 0.06 & 0.05 & 0.05 \\ 
dist & False & False & False & False & False & False & False & False \\ 
\hline 
\end{tabular}
\caption{Hyperparameters for the estimates in \Cref{tab:twonn_bm}}

\end{table}

\begin{table}[]
    \centering
    \begin{tabular}{| l || l | } \hline
\wodcap &  \\ \hline 
k & 10, 20, 40, 80\\ \hline 
comb & mean, hmean, median\\ \hline 
\end{tabular}
\caption{Hyperparameter range for \wodcap}
    \label{tab:wodcap_hp}
\end{table}

\begin{table}[]
    \centering
    \begin{tabular}{| l || l l | l l |l l |l l |}\hline 
\wodcap & 625 &  & 1250 & & 2500 & & 5000 & \\ \hline 
    & med abs  & med rel & med abs  & med rel & med abs  & med rel & med abs  & med rel  \\ \hline 
comb & hmean & median & median & mean & median & mean & median & mean   \\ 
k & 20 & 20 & 20 & 20 & 20 & 20 & 20 & 20   \\ 
\hline 
\end{tabular}
 \caption{Hyperparameters for the estimates in \Cref{tab:wodcap_bm}}
\label{tab:wodcap_bmhp}
\end{table}

\clearpage
\section{Noise experiments}
\setcounter{table}{0}
\setcounter{figure}{0}

\input{tables/noise/6_11_gaussian}
\input{tables/noise/6_11_outliers}
\input{tables/noise/10_11_gaussian}
\input{tables/noise/10_11_outliers}
\input{tables/noise/6_16_so_gaussian}
\input{tables/noise/6_16_so_outliers}

\end{document}

%% file: tables/lpca_thresh_n_5000_k_80_mean.tex
\begin{table}[h!]
\centering
\tiny
\begin{tabular}{|l|ll||l|l|l|l|l|l|l|l|} \hline 
 & &  & FO & Fan & maxgap & ratio & part. ratio & Kaiser & broken stick & Minka \\  \hline  
M1 Sphere & 10 & 11 & 10.5 (0.0) & 6.1 (0.0) & 8.0 (0.0) & 6.7 (0.0) & 7.5 (0.0) & 5.5 (0.0) & 1.0 (0.0) & 8.9 (0.0)\\  
M2 Affine 3to5 & 3 & 5 & 3.0 (0.0) & 2.0 (0.0) & 1.7 (0.0) & 3.1 (0.0) & 1.6 (0.0) & 2.7 (0.0) & 3.8 (0.0) & 2.0 (0.0)\\  
M3 Nonlinear 4to6 & 4 & 6 & 5.0 (0.0) & 2.7 (0.0) & 3.1 (0.0) & 2.9 (0.0) & 2.9 (0.0) & 2.9 (0.0) & 2.9 (0.0) & 3.4 (0.0)\\  
M4 Nonlinear & 4 & 8 & 6.3 (0.0) & 3.1 (0.0) & 4.3 (0.0) & 3.3 (0.0) & 3.7 (0.0) & 3.6 (0.0) & 3.2 (0.0) & 4.4 (0.0)\\  
M5a Helix1d & 1 & 3 & 1.0 (0.0) & 1.8 (0.0) & 2.0 (0.0) & 3.0 (0.0) & 0.0 (0.0) & 1.0 (0.0) & 1.6 (0.0) & 1.6 (0.0)\\  
M5b Helix2d & 2 & 3 & 3.0 (0.0) & 1.9 (0.0) & 2.0 (0.0) & 1.9 (0.0) & 1.8 (0.0) & 1.6 (0.0) & 2.1 (0.0) & 2.0 (0.0)\\  
M6 Nonlinear & 6 & 36 & 11.1 (0.0) & 5.3 (0.0) & 9.0 (0.0) & 5.9 (0.0) & 6.4 (0.0) & 8.6 (0.0) & 8.1 (0.0) & 8.3 (0.0)\\  
M7 Roll & 2 & 3 & 2.0 (0.0) & 1.0 (0.0) & 1.0 (0.0) & 2.8 (0.0) & 0.6 (0.0) & 1.8 (0.0) & 2.9 (0.0) & 1.0 (0.0)\\  
M8 Nonlinear & 12 & 72 & 23.3 (0.0) & 11.1 (0.0) & 20.5 (0.1) & 12.3 (0.0) & 13.5 (0.0) & 18.9 (0.0) & 17.0 (0.0) & 18.2 (0.0)\\  
M9 Affine & 20 & 20 & 20.0 (0.0) & 9.7 (0.0) & 16.8 (0.0) & 9.9 (0.0) & 12.1 (0.0) & 8.9 (0.0) & 1.0 (0.0) & 16.6 (0.0)\\  
M10a Cubic & 10 & 11 & 11.0 (0.0) & 5.9 (0.0) & 8.5 (0.0) & 6.0 (0.0) & 7.0 (0.0) & 5.1 (0.0) & 1.0 (0.0) & 9.0 (0.0)\\  
M10b Cubic & 17 & 18 & 18.0 (0.0) & 8.9 (0.0) & 15.1 (0.0) & 9.0 (0.0) & 11.1 (0.0) & 8.1 (0.0) & 1.0 (0.0) & 15.0 (0.0)\\  
M10c Cubic & 24 & 25 & 25.0 (0.0) & 11.7 (0.0) & 21.2 (0.0) & 11.9 (0.0) & 14.6 (0.0) & 11.0 (0.0) & 1.0 (0.0) & 20.3 (0.0)\\  
M10d Cubic & 70 & 71 & 54.8 (0.0) & 25.3 (0.0) & 55.2 (0.0) & 26.3 (0.0) & 32.2 (0.0) & 28.2 (0.0) & 1.0 (0.0) & 45.4 (0.0)\\  
M11 Moebius & 2 & 3 & 2.4 (0.0) & 1.2 (0.0) & 1.2 (0.0) & 2.7 (0.0) & 0.9 (0.0) & 1.8 (0.0) & 2.6 (0.0) & 1.2 (0.0)\\  
M12 Norm & 20 & 20 & 20.0 (0.0) & 9.9 (0.0) & 17.0 (0.0) & 9.9 (0.0) & 12.2 (0.0) & 8.9 (0.0) & 1.0 (0.0) & 16.9 (0.0)\\  
M13a Scurve & 2 & 3 & 2.0 (0.0) & 1.0 (0.0) & 1.0 (0.0) & 2.8 (0.0) & 0.6 (0.0) & 1.8 (0.0) & 2.9 (0.0) & 1.0 (0.0)\\  
M13b Spiral & 1 & 13 & 2.0 (0.0) & 1.0 (0.0) & 1.0 (0.0) & 6.5 (0.1) & 0.7 (0.0) & 2.0 (0.0) & 3.0 (0.0) & 1.0 (0.0)\\  
Mbeta & 10 & 40 & 9.8 (0.0) & 4.5 (0.0) & 5.9 (0.1) & 5.2 (0.0) & 4.7 (0.0) & 9.0 (0.0) & 7.3 (0.0) & 7.7 (0.0)\\  
Mn1 Nonlinear & 18 & 72 & 18.3 (0.0) & 10.7 (0.0) & 20.5 (0.0) & 11.6 (0.0) & 13.1 (0.0) & 17.9 (0.0) & 17.4 (0.0) & 16.0 (0.0)\\  
Mn2 Nonlinear & 24 & 96 & 24.3 (0.0) & 13.3 (0.0) & 29.2 (0.0) & 14.6 (0.0) & 16.6 (0.0) & 22.3 (0.0) & 20.9 (0.0) & 21.1 (0.0)\\  
Mp1 Paraboloid & 3 & 12 & 3.0 (0.0) & 2.0 (0.0) & 1.6 (0.0) & 3.9 (0.0) & 1.6 (0.0) & 3.0 (0.0) & 3.9 (0.0) & 2.0 (0.0)\\  
Mp2 Paraboloid & 6 & 21 & 5.9 (0.0) & 3.8 (0.0) & 2.8 (0.0) & 4.1 (0.0) & 4.0 (0.0) & 5.9 (0.0) & 6.5 (0.0) & 5.0 (0.0)\\  
Mp3 Paraboloid & 9 & 30 & 8.8 (0.0) & 5.3 (0.0) & 4.5 (0.1) & 5.9 (0.0) & 5.4 (0.0) & 8.6 (0.0) & 8.8 (0.0) & 7.6 (0.0)\\  
\hline 
 \end{tabular}
 \caption{Performance of \lpca on benchmark datasets, each consisting of 5000 points, and we report the mean and standard deviation in the dimension estimate over 20 samples. The knn neighbourhood is fixed to be  $k = 80$, and the local dimension estimates are aggregated using the mean. Individual hyperparameters of the thresholding methods are \scikitdimension defaults, as stated in \Cref{tab:lpca_hprange}. }
 \label{tab:lpca_thresh}
\end{table}

%% file: tables/hyperparameters/ph_alpha.tex
\begin{table}[h]
\centering
\begin{tabular}{|l||l|l|l|l|l|l|l|}
\hline
$\alpha$ & 0.5        & 0.75       & 1.0        & 1.25       & 1.5        & 1.75       & 2.0        \\ \hline
$S^2$    & 2.02(0.04) & 2.01(0.04) & 2.01(0.04) & 2.01(0.04) & 2.01(0.04) & 2.00(0.04) & 2.00(0.04) \\ \hline
$S^4$    & 4.02(0.13) & 4.00(0.13) & 4.00(0.13) & 4.00(0.12) & 4.00(0.12) & 4.00(0.12) & 4.00(0.12) \\ \hline
$S^8$    & 7.69(0.18) & 7.63(0.17) & 7.60(0.17) & 7.58(0.17) & 7.56(0.17) & 7.55(0.17) & 7.53(0.17) \\ \hline
\end{tabular}\vspace{1em}
\caption{Dimension estimates from \phzero, given samples of 1000 points on spheres of different dimensions, for different choices of the power parameter $\alpha$.\label{tab:ph_alpha}}
\end{table}

%% file: tables/lPCA_benchmark.tex
\begin{sidewaystable}[h]
\centering
\tiny
\begin{tabular}{|l|ll||lll|lll|lll|lll|}\hline
               &    &    & lPCA       &            &            &            &            &            &            &            &            &            &            &            \\ \hline
n              &    &    & 625        &            &            & 1250       &            &            & 2500       &            &            & 5000       &            &            \\ \hline
               & d  & n  & Best       & med abs    & med rel    & Best       & med abs    & med rel    & Best       & med abs    & med rel    & Best       & med abs    & med rel    \\ \hline
M1 Sphere      & 10 & 11 & 10.0 (0.0) & 9.8 (0.1)  & 11.0 (0.0) & 10.0 (0.0) & 10.8 (0.0) & 11.0 (0.0) & 10.0 (0.0) & 11.0 (0.0) & 11.0 (0.0) & 10.0 (0.0) & 10.0 (0.0) & 10.0 (0.0) \\ \hline
M2 Affine      & 3  & 5  & 3.0 (0.0)  & 3.0 (0.0)  & 3.0 (0.0)  & 3.0 (0.0)  & 3.0 (0.0)  & 3.0 (0.0)  & 3.0 (0.0)  & 3.0 (0.0)  & 3.0 (0.0)  & 3.0 (0.0)  & 3.0 (0.0)  & 3.0 (0.0)  \\ \hline
M3 Nonlinear   & 4  & 6  & 4.0 (0.0)  & 4.0 (0.1)  & 5.0 (0.0)  & 4.0 (0.0)  & 5.0 (0.0)  & 5.0 (0.0)  & 4.0 (0.0)  & 5.0 (0.0)  & 5.0 (0.0)  & 4.0 (0.0)  & 5.0 (0.0)  & 5.0 (0.0)  \\ \hline
M4 Nonlinear   & 4  & 8  & 4.0 (0.0)  & 4.6 (0.2)  & 7.0 (0.0)  & 4.0 (0.0)  & 6.6 (0.0)  & 7.0 (0.0)  & 4.0 (0.0)  & 7.0 (0.0)  & 7.0 (0.0)  & 4.0 (0.0)  & 6.0 (0.0)  & 6.0 (0.0)  \\ \hline
M5a Helix1d    & 1  & 3  & 1.0 (0.0)  & 1.1 (0.0)  & 3.0 (0.0)  & 1.0 (0.0)  & 1.4 (0.0)  & 1.0 (0.0)  & 1.0 (0.0)  & 1.0 (0.0)  & 1.0 (0.0)  & 1.0 (0.0)  & 1.0 (0.0)  & 1.0 (0.0)  \\ \hline
M5b Helix2d    & 2  & 3  & 2.0 (0.0)  & 2.9 (0.0)  & 3.0 (0.0)  & 2.0 (0.0)  & 3.0 (0.0)  & 3.0 (0.0)  & 2.0 (0.0)  & 3.0 (0.0)  & 3.0 (0.0)  & 2.0 (0.0)  & 3.0 (0.0)  & 3.0 (0.0)  \\ \hline
M6 Nonlinear   & 6  & 36 & 6.0 (0.0)  & 6.0 (0.3)  & 11.0 (0.0) & 6.0 (0.0)  & 11.0 (0.1) & 11.0 (0.0) & 6.0 (0.0)  & 12.0 (0.2) & 12.0 (0.2) & 6.0 (0.0)  & 11.0 (0.0) & 11.0 (0.0) \\ \hline
M7 Roll        & 2  & 3  & 2.0 (0.0)  & 2.0 (0.0)  & 2.4 (0.5)  & 2.0 (0.0)  & 2.1 (0.0)  & 2.0 (0.0)  & 2.0 (0.0)  & 2.0 (0.0)  & 2.0 (0.0)  & 2.0 (0.0)  & 2.0 (0.0)  & 2.0 (0.0)  \\ \hline
M8 Nonlinear   & 12 & 72 & 12.0 (0.0) & 7.5 (0.4)  & 20.4 (0.5) & 12.0 (0.0) & 20.1 (0.1) & 20.0 (0.0) & 12.0 (0.0) & 24.0 (0.2) & 24.0 (0.2) & 12.0 (0.0) & 23.0 (0.0) & 23.0 (0.0) \\ \hline
M9 Affine      & 20 & 20 & 20.0 (0.0) & 10.4 (0.4) & 19.0 (0.0) & 20.0 (0.0) & 19.3 (0.0) & 19.0 (0.0) & 20.0 (0.0) & 20.0 (0.0) & 20.0 (0.0) & 20.0 (0.0) & 20.0 (0.0) & 20.0 (0.0) \\ \hline
M10a Cubic     & 10 & 11 & 10.0 (0.0) & 9.0 (0.1)  & 11.0 (0.0) & 10.0 (0.0) & 11.0 (0.0) & 11.0 (0.0) & 10.0 (0.0) & 11.0 (0.0) & 11.0 (0.0) & 10.0 (0.0) & 11.0 (0.0) & 11.0 (0.0) \\ \hline
M10b Cubic     & 17 & 18 & 16.5 (0.1) & 11.0 (0.3) & 18.0 (0.0) & 16.4 (0.1) & 17.8 (0.0) & 18.0 (0.0) & 17.6 (0.0) & 18.0 (0.0) & 18.0 (0.0) & 17.6 (0.0) & 18.0 (0.0) & 18.0 (0.0) \\ \hline
M10c Cubic     & 24 & 25 & 24.0 (0.1) & 11.6 (0.4) & 23.0 (0.0) & 24.0 (0.1) & 22.6 (0.0) & 23.0 (0.0) & 23.9 (0.0) & 25.0 (0.0) & 25.0 (0.0) & 23.7 (0.0) & 25.0 (0.0) & 25.0 (0.0) \\ \hline
M10d Cubic     & 70 & 72 & 55.2 (0.1) & 11.6 (0.6) & 37.0 (0.0) & 55.2 (0.0) & 37.4 (0.0) & 37.0 (0.0) & 55.2 (0.0) & 54.2 (0.4) & 54.2 (0.4) & 55.2 (0.0) & 55.0 (0.0) & 55.0 (0.0) \\ \hline
M11 Moebius    & 2  & 3  & 2.0 (0.0)  & 2.4 (0.0)  & 3.0 (0.0)  & 2.0 (0.0)  & 2.9 (0.0)  & 3.0 (0.0)  & 2.0 (0.0)  & 3.0 (0.0)  & 3.0 (0.0)  & 2.0 (0.0)  & 2.0 (0.0)  & 2.0 (0.0)  \\ \hline
M12 Norm       & 20 & 20 & 20.0 (0.0) & 6.9 (0.3)  & 19.0 (0.0) & 20.0 (0.0) & 19.3 (0.0) & 19.0 (0.0) & 20.0 (0.0) & 20.0 (0.0) & 20.0 (0.0) & 20.0 (0.0) & 20.0 (0.0) & 20.0 (0.0) \\ \hline
M13a Scurve    & 2  & 3  & 2.0 (0.0)  & 2.0 (0.0)  & 2.0 (0.0)  & 2.0 (0.0)  & 2.0 (0.0)  & 2.0 (0.0)  & 2.0 (0.0)  & 2.0 (0.0)  & 2.0 (0.0)  & 2.0 (0.0)  & 2.0 (0.0)  & 2.0 (0.0)  \\ \hline
M13b Spiral    & 1  & 13 & 1.0 (0.0)  & 2.0 (0.0)  & 2.0 (0.0)  & 1.0 (0.0)  & 2.0 (0.0)  & 2.0 (0.0)  & 1.0 (0.0)  & 2.0 (0.0)  & 2.0 (0.0)  & 1.0 (0.0)  & 2.0 (0.0)  & 2.0 (0.0)  \\ \hline
Mbeta          & 10 & 40 & 10.0 (0.0) & 4.5 (0.2)  & 9.0 (0.0)  & 10.0 (0.0) & 9.0 (0.1)  & 9.0 (0.0)  & 10.0 (0.0) & 10.0 (0.0) & 10.0 (0.0) & 10.0 (0.0) & 10.0 (0.0) & 10.0 (0.0) \\ \hline
Mn1 Nonlinear  & 18 & 72 & 18.0 (0.0) & 8.6 (0.5)  & 18.0 (0.0) & 18.0 (0.0) & 17.9 (0.0) & 18.0 (0.0) & 18.0 (0.0) & 18.2 (0.4) & 18.2 (0.4) & 18.0 (0.0) & 18.0 (0.0) & 18.0 (0.0) \\ \hline
Mn2 Nonlinear  & 24 & 96 & 24.0 (0.0) & 8.9 (0.5)  & 22.9 (0.3) & 24.0 (0.0) & 22.5 (0.0) & 22.5 (0.5) & 24.0 (0.0) & 24.0 (0.0) & 24.0 (0.0) & 24.0 (0.0) & 24.0 (0.0) & 24.0 (0.0) \\ \hline
Mp1 Paraboloid & 3  & 12 & 3.0 (0.0)  & 2.9 (0.0)  & 3.0 (0.0)  & 3.0 (0.0)  & 3.0 (0.0)  & 3.0 (0.0)  & 3.0 (0.0)  & 3.0 (0.0)  & 3.0 (0.0)  & 3.0 (0.0)  & 3.0 (0.0)  & 3.0 (0.0)  \\ \hline
Mp2 Paraboloid & 6  & 21 & 6.0 (0.0)  & 5.1 (0.1)  & 6.0 (0.0)  & 6.0 (0.0)  & 5.9 (0.0)  & 6.0 (0.0)  & 6.0 (0.0)  & 6.0 (0.0)  & 6.0 (0.0)  & 6.0 (0.0)  & 6.0 (0.0)  & 6.0 (0.0)  \\ \hline
Mp3 Paraboloid & 9  & 30 & 9.0 (0.0)  & 6.8 (0.2)  & 9.0 (0.0)  & 9.0 (0.0)  & 8.4 (0.0)  & 9.0 (0.0)  & 9.0 (0.0)  & 9.0 (0.0)  & 9.0 (0.0)  & 9.0 (0.0)  & 9.0 (0.0)  & 9.0 (0.0) \\ 
\hline
\end{tabular}
\caption{\label{tab:lPCA_bm} \textbf{Number of Samples}. With few samples, estimates can be sensitive to hyperparameter choices, such as in the case of M10b,c,d Cubic, and M12 Norm. However, the estimates become more accurate on more datasets as the number of samples increases. \textbf{High dimensional datasets}. With many samples, the only high dimensional dataset that troubles lPCA is M10d Cubic, which it significantly underestimates. \textbf{Variance}. lPCA has the lowest variance among all estimators. Note that the aggregation method over local dimension estimates is often the median in this table (see \Cref{tab:lPCA_bmhp}), which tends to produce an integer value; this effectively `rounded' aggregation also reduces the variance.  \textbf{Hyperparameter dependency}. While this is an issue if there are few samples for some datasets, for many samples, the estimates between the best performance and med rel, med abs are close. Notable exceptions include some non-linear datasets, such as M3, M4, M6, and M8 Nonlinear; and M5b Helix2d. \textbf{Nonlinear datasets}. With many point samples, lPCA performs well on Mn1,2 Nonlinear, and Mp1,2,3 Paraboloids, which many other estimators struggle with. However, some estimators perform much more consistently with regards to hyperparameter sensitivity on the other nonlinear datasets  M3, M4, M6,  M8 Nonlinear, M13b Spiral.  }
\end{sidewaystable}

%% file: tables/PHnew.tex
\begin{sidewaystable}[h]
\centering
\tiny
\begin{tabular}{|l|ll||lll|lll|lll|lll|}
\hline               &    &    & PH &              &              &              &              &              &              &              &              &              &              &              \\  \hline              &    &    &              &              &              &              &              &              &              &              &              &              &              &              \\  \hline              &    &    & 625          &              &              & 1250         &              &              & 2500         &              &              & 5000         &              &              \\  \hline               &    &    &              &              &              &              &              &              &              &              &              &              &              &              \\  \hline               & d  & n  & Best         & med abs      & med rel      & Best         & med abs      & med rel      & Best         & med abs      & med rel      & Best         & med abs      & med rel      \\  \hline               &    &    &              &              &              &              &              &              &              &              &              &              &              &              \\  \hline               M1Sphere & 10 & 11 & 9.2 (0.47) & 9.2 (0.47) & 9.2 (0.47)&9.2 (0.32) & 9.2 (0.32) & 9.2 (0.32)&9.4 (0.16) & 9.4 (0.16) & 9.4 (0.16)&9.5 (0.12) & 9.5 (0.12) & 9.5 (0.12)\\ \hline M2Affine3to5 & 3 & 5 & 2.9 (0.1) & 2.9 (0.1) & 2.9 (0.1)&2.9 (0.07) & 2.9 (0.07) & 2.9 (0.07)&2.9 (0.05) & 2.9 (0.05) & 2.9 (0.05)&2.9 (0.03) & 2.9 (0.03) & 2.9 (0.03)\\ \hline M3Nonlinear4to6 & 4 & 6 & 3.8 (0.18) & 3.8 (0.18) & 3.8 (0.18)&3.9 (0.12) & 3.9 (0.12) & 3.9 (0.12)&3.9 (0.07) & 3.9 (0.07) & 3.9 (0.07)&3.9 (0.08) & 3.9 (0.08) & 3.9 (0.08)\\ \hline M4Nonlinear & 4 & 8 & 4.0 (0.13) & 4.0 (0.13) & 4.0 (0.13)&4.0 (0.1) & 4.0 (0.1) & 4.0 (0.1)&3.9 (0.09) & 3.9 (0.09) & 3.9 (0.09)&3.9 (0.07) & 3.9 (0.07) & 3.9 (0.07)\\ \hline M5aHelix1d & 1 & 3 & 1.0 (0.02) & 1.0 (0.02) & 1.0 (0.02)&1.0 (0.01) & 1.0 (0.01) & 1.0 (0.01)&1.0 (0.01) & 1.0 (0.01) & 1.0 (0.01)&1.0 (0.0) & 1.0 (0.0) & 1.0 (0.0)\\ \hline M5bHelix2d & 2 & 3 & 2.8 (0.12) & 2.8 (0.12) & 2.8 (0.12)&2.6 (0.15) & 2.6 (0.15) & 2.6 (0.15)&2.3 (0.09) & 2.3 (0.09) & 2.3 (0.09)&2.0 (0.03) & 2.0 (0.03) & 2.0 (0.03)\\  \hline M6Nonlinear & 6 & 36 & 6.6 (0.32) & 6.6 (0.32) & 6.6 (0.32)&6.3 (0.3) & 6.3 (0.3) & 6.3 (0.3)&6.1 (0.15) & 6.1 (0.15) & 6.1 (0.15)&6.0 (0.11) & 6.0 (0.11) & 6.0 (0.11)\\ \hline M7Roll & 2 & 3 & 2.0 (0.06) & 2.0 (0.06) & 2.0 (0.06)&2.0 (0.05) & 2.0 (0.05) & 2.0 (0.05)&2.0 (0.03) & 2.0 (0.03) & 2.0 (0.03)&2.0 (0.01) & 2.0 (0.01) & 2.0 (0.01)\\ \hline M8Nonlinear & 12 & 72 & 14.1 (0.82) & 14.1 (0.82) & 14.1 (0.82)&14.1 (0.57) & 14.1 (0.57) & 14.1 (0.57)&13.8 (0.41) & 13.8 (0.41) & 13.8 (0.41)&13.6 (0.31) & 13.6 (0.31) & 13.6 (0.31)\\ \hline M9Affine & 20 & 20 & 15.6 (0.79) & 15.6 (0.79) & 15.6 (0.79)&15.3 (0.44) & 15.3 (0.44) & 15.3 (0.44)&15.3 (0.39) & 15.3 (0.39) & 15.3 (0.39)&15.7 (0.2) & 15.7 (0.2) & 15.7 (0.2)\\ \hline M10aCubic & 10 & 11 & 9.0 (0.4) & 9.0 (0.4) & 9.0 (0.4)&9.0 (0.27) & 9.0 (0.27) & 9.0 (0.27)&9.2 (0.18) & 9.2 (0.18) & 9.2 (0.18)&9.2 (0.12) & 9.2 (0.12) & 9.2 (0.12)\\ \hline M10bCubic & 17 & 18 & 13.7 (0.66) & 13.7 (0.66) & 13.7 (0.66)&13.7 (0.47) & 13.7 (0.47) & 13.7 (0.47)&14.1 (0.45) & 14.1 (0.45) & 14.1 (0.45)&14.2 (0.28) & 14.2 (0.28) & 14.2 (0.28)\\ \hline M10cCubic & 24 & 25 & 18.0 (0.82) & 18.0 (0.82) & 18.0 (0.82)&18.2 (0.58) & 18.2 (0.58) & 18.2 (0.58)&18.5 (0.5) & 18.5 (0.5) & 18.5 (0.5)&18.7 (0.34) & 18.7 (0.34) & 18.7 (0.34)\\ \hline M10dCubic & 70 & 72 & 40.3 (3.26) & 40.3 (3.26) & 40.3 (3.26)&39.6 (2.05) & 39.6 (2.05) & 39.6 (2.05)&40.3 (1.4) & 40.3 (1.4) & 40.3 (1.4)&41.6 (1.02) & 41.6 (1.02) & 41.6 (1.02)\\ \hline M11Moebius & 2 & 3 & 2.0 (0.06) & 2.0 (0.06) & 2.0 (0.06)&2.0 (0.05) & 2.0 (0.05) & 2.0 (0.05)&2.0 (0.03) & 2.0 (0.03) & 2.0 (0.03)&2.0 (0.02) & 2.0 (0.02) & 2.0 (0.02)\\ \hline M12Norm & 20 & 20 & 16.5 (0.87) & 16.5 (0.87) & 16.5 (0.87)&16.9 (0.78) & 16.9 (0.78) & 16.9 (0.78)&17.1 (0.53) & 17.1 (0.53) & 17.1 (0.53)&17.4 (0.36) & 17.4 (0.36) & 17.4 (0.36)\\ \hline M13aScurve & 2 & 3 & 2.0 (0.06) & 2.0 (0.06) & 2.0 (0.06)&2.0 (0.04) & 2.0 (0.04) & 2.0 (0.04)&2.0 (0.03) & 2.0 (0.03) & 2.0 (0.03)&2.0 (0.02) & 2.0 (0.02) & 2.0 (0.02)\\ \hline M13bSpiral & 1 & 13 & 1.8 (0.19) & 1.8 (0.19) & 1.8 (0.19)&1.5 (0.17) & 1.5 (0.17) & 1.5 (0.17)&1.1 (0.05) & 1.1 (0.05) & 1.1 (0.05)&1.0 (0.01) & 1.0 (0.01) & 1.0 (0.01)\\ \hline Mbeta & 10 & 40 & 6.3 (0.52) & 6.3 (0.52) & 6.3 (0.52)&6.4 (0.41) & 6.4 (0.41) & 6.4 (0.41)&6.5 (0.32) & 6.5 (0.32) & 6.5 (0.32)&6.7 (0.25) & 6.7 (0.25) & 6.7 (0.25)\\ \hline Mn1Nonlinear & 18 & 72 & 14.1 (0.75) & 14.1 (0.75) & 14.1 (0.75)&14.2 (0.48) & 14.2 (0.48) & 14.2 (0.48)&14.5 (0.45) & 14.5 (0.45) & 14.5 (0.45)&14.5 (0.3) & 14.5 (0.3) & 14.5 (0.3)\\ \hline Mn2Nonlinear & 24 & 96 & 18.0 (1.15) & 18.0 (1.15) & 18.0 (1.15)&18.3 (0.74) & 18.3 (0.74) & 18.3 (0.74)&18.4 (0.52) & 18.4 (0.52) & 18.4 (0.52)&18.6 (0.43) & 18.6 (0.43) & 18.6 (0.43)\\ \hline Mp1Paraboloid & 3 & 12 & 2.8 (0.11) & 2.8 (0.11) & 2.8 (0.11)&2.9 (0.09) & 2.9 (0.09) & 2.9 (0.09)&2.9 (0.07) & 2.9 (0.07) & 2.9 (0.07)&2.9 (0.05) & 2.9 (0.05) & 2.9 (0.05)\\ \hline Mp2Paraboloid & 6 & 21 & 4.7 (0.31) & 4.7 (0.31) & 4.7 (0.31)&5.0 (0.2) & 5.0 (0.2) & 5.0 (0.2)&5.3 (0.17) & 5.3 (0.17) & 5.3 (0.17)&5.4 (0.11) & 5.4 (0.11) & 5.4 (0.11)\\ \hline Mp3Paraboloid & 9 & 30 & 5.3 (0.68) & 5.3 (0.68) & 5.3 (0.68)&6.3 (0.52) & 6.3 (0.52) & 6.3 (0.52)&6.9 (0.35) & 6.9 (0.35) & 6.9 (0.35)&7.2 (0.16) & 7.2 (0.16) & 7.2 (0.16)\\ \hline    
\end{tabular}
\label{tab:ph_bm}
\caption{\footnotesize \textbf{Number of samples}. PH has good performance for low dimensional manifolds even with few points. \textbf{High dimensional datasets}. PH is prone to underestimate for high dimensional datasets, such as M9 Affine, and M10b,c,d, Cubic, even with many samples in the benchmark. \textbf{Variance}. The variance between different point samples is relatively high, especially for high dimensional datasets with few samples, such as M10d cubic. See \Cref{rmk:ph_knn_error} for discussion on this. \textbf{ Hyperparameter dependency}. Across three different point sampling regimes, and all three different ways of reporting dimension estimates, the choice of $\alpha = 0.5$ out of $\alpha \in \{0.5, 1.0, 1.5, 2.0\}$ consistently produces the best results. \textbf{ Nonlinear datasets} On Mbeta, Mn1,2 Nonlinear, the propensity to underestimate remain even after increasing the number of samples.  }
\end{sidewaystable}

%% file: tables/KNNb_benchmark.tex
\begin{sidewaystable}[h]
\centering
\tiny
\begin{tabular}{|l|ll||lll|lll|lll|lll|}
\hline
               &    &    & KNN &              &              &              &              &              &              &              &              &              &              &              \\  \hline
               &    &    &              &              &              &              &              &              &              &              &              &              &              &              \\  \hline
n              &    &    & 625          &              &              & 1250         &              &              & 2500         &              &              & 5000         &              &              \\  \hline
               &    &    &              &              &              &              &              &              &              &              &              &              &              &              \\  \hline
               & d  & n  & Best         & med abs      & med rel      & Best         & med abs      & med rel      & Best         & med abs      & med rel      & Best         & med abs      & med rel      \\  \hline
               &    &    &              &              &              &              &              &              &              &              &              &              &              &              \\  \hline 
               M1 Sphere & 10 & 11 & 9.6 (1.4) & 13.2 (13.7) & 9.5 (2.0)&9.9 (1.9) & 9.9 (1.9) & 9.9 (1.9)&9.8 (1.4) & 9.8 (1.4) & 9.8 (1.4)&9.7 (0.8) & 9.7 (0.8) & 9.5 (0.7)\\ \hline M2 Affine 3to5 & 3 & 5 & 2.9 (0.3) & 2.9 (0.3) & 2.8 (0.2)&2.9 (0.2) & 2.9 (0.2) & 2.9 (0.2)&2.9 (0.1) & 2.9 (0.1) & 2.9 (0.1)&2.9 (0.1) & 2.9 (0.1) & 2.9 (0.1)\\ \hline M3 Nonlinear 4to6 & 4 & 6 & 3.8 (0.4) & 3.8 (0.4) & 3.5 (0.2)&3.8 (0.4) & 3.8 (0.4) & 3.8 (0.4)&3.9 (0.2) & 3.9 (0.2) & 3.9 (0.2)&3.8 (0.2) & 3.7 (0.1) & 3.8 (0.2)\\ \hline M4 Nonlinear & 4 & 8 & 4.0 (0.2) & 3.9 (0.6) & 3.7 (0.3)&4.0 (0.1) & 3.9 (0.3) & 3.9 (0.3)&3.9 (0.3) & 3.9 (0.2) & 3.9 (0.2)&3.8 (0.2) & 3.7 (0.1) & 3.8 (0.1)\\ \hline M5a Helix1d & 1 & 3 & 1.0 (0.0) & 1.0 (0.1) & 1.0 (0.1)&1.0 (0.0) & 1.0 (0.1) & 1.0 (0.1)&1.0 (0.0) & 1.0 (0.0) & 1.0 (0.0)&1.0 (0.0) & 1.0 (0.0) & 1.0 (0.0)\\ \hline M5b Helix2d & 2 & 3 & 2.2 (0.1) & 2.6 (0.2) & 2.6 (0.2)&2.3 (0.1) & 2.3 (0.2) & 2.3 (0.2)&2.1 (0.1) & 2.1 (0.1) & 2.1 (0.1)&2.0 (0.1) & 2.1 (0.0) & 2.0 (0.1)\\ \hline M6 Nonlinear & 6 & 36 & 6.0 (0.5) & 6.2 (1.1) & 5.9 (0.7)&6.0 (0.2) & 5.6 (0.5) & 5.6 (0.5)&6.0 (0.2) & 5.8 (0.4) & 5.8 (0.4)&5.9 (0.1) & 5.3 (0.2) & 5.5 (0.2)\\ \hline M7 Roll & 2 & 3 & 2.0 (0.1) & 2.0 (0.2) & 1.9 (0.1)&2.0 (0.1) & 2.0 (0.1) & 2.0 (0.1)&2.0 (0.1) & 2.0 (0.1) & 2.0 (0.1)&2.0 (0.0) & 2.0 (0.0) & 2.0 (0.1)\\ \hline M8 Nonlinear & 12 & 72 & 13.7 (2.9) & 14.6 (7.1) & 14.4 (3.1)&13.7 (1.4) & 13.7 (2.7) & 13.7 (2.7)&12.3 (1.3) & 12.3 (1.3) & 12.3 (1.3)&12.3 (0.6) & 12.3 (0.6) & 12.7 (0.8)\\ \hline M9 Affine & 20 & 20 & 20.0 (2.5) & 23.8 (21.1) & 18.2 (4.8)&20.1 (1.5) & 16.3 (4.6) & 16.3 (4.6)&18.1 (0.8) & 15.5 (2.7) & 15.5 (2.7)&16.9 (0.5) & 15.2 (1.1) & 15.5 (1.6)\\ \hline M10a Cubic & 10 & 11 & 9.7 (2.8) & 9.7 (2.8) & 8.8 (1.5)&9.4 (1.4) & 9.4 (1.4) & 9.4 (1.4)&9.0 (1.0) & 9.0 (1.0) & 9.0 (1.0)&8.9 (0.7) & 8.5 (0.4) & 8.7 (0.6)\\ \hline M10b Cubic & 17 & 18 & 17.1 (2.5) & 18.2 (9.4) & 14.3 (3.0)&16.0 (0.9) & 12.8 (1.9) & 12.8 (1.9)&15.1 (0.5) & 14.4 (2.8) & 14.4 (2.8)&14.3 (0.4) & 13.3 (0.9) & 13.3 (1.2)\\ \hline M10c Cubic & 24 & 25 & 24.0 (5.6) & 30.7 (29.0) & 22.0 (7.6)&24.0 (4.2) & 19.3 (6.0) & 19.3 (6.0)&23.5 (1.6) & 19.8 (6.8) & 19.8 (6.8)&21.9 (1.2) & 18.7 (1.5) & 19.3 (1.5)\\ \hline M10d Cubic & 70 & 72 & 96.8 (323.5) & 166.3 (667.8) & 96.8 (323.5)&65.7 (1209.4) & -313.5 (1667.0) & -313.5 (1667.0)&68.5 (768.6) & -31.8 (471.5) & -31.8 (471.5)&68.4 (120.4) & 92.0 (36.6) & 67.8 (28.7)\\ \hline M11 Moebius & 2 & 3 & 2.0 (0.1) & 1.9 (0.1) & 1.9 (0.1)&2.0 (0.1) & 2.0 (0.2) & 2.0 (0.2)&2.0 (0.0) & 2.0 (0.1) & 2.0 (0.1)&2.0 (0.1) & 1.9 (0.0) & 2.0 (0.1)\\ \hline M12 Norm & 20 & 20 & 19.8 (4.3) & 17.3 (4.2) & 19.8 (4.3)&20.1 (2.5) & 19.0 (5.1) & 19.0 (5.1)&20.0 (1.1) & 19.1 (3.1) & 19.1 (3.1)&20.0 (1.0) & 18.8 (1.7) & 17.8 (1.7)\\ \hline M13a Scurve & 2 & 3 & 2.0 (0.2) & 2.0 (0.2) & 2.0 (0.1)&2.0 (0.1) & 1.9 (0.1) & 1.9 (0.1)&2.0 (0.1) & 2.0 (0.1) & 2.0 (0.1)&2.0 (0.1) & 2.0 (0.0) & 2.0 (0.1)\\ \hline M13b Spiral & 1 & 13 & 1.3 (0.1) & 1.3 (0.1) & 1.9 (0.1)&1.0 (0.1) & 1.0 (0.1) & 1.0 (0.1)&1.0 (0.0) & 1.0 (0.0) & 1.0 (0.0)&1.0 (0.0) & 1.0 (0.0) & 1.0 (0.0)\\ \hline Mbeta & 10 & 40 & 6.4 (0.8) & 6.4 (1.2) & 5.8 (0.7)&6.2 (0.9) & 6.2 (0.9) & 6.2 (0.9)&6.6 (0.7) & 6.6 (0.7) & 6.6 (0.7)&6.6 (0.4) & 6.1 (0.2) & 6.3 (0.4)\\ \hline Mn1 Nonlinear & 18 & 72 & 18.0 (3.0) & 18.2 (8.2) & 13.7 (2.7)&17.0 (1.7) & 13.7 (2.0) & 13.7 (2.0)&15.6 (0.7) & 15.5 (3.6) & 15.5 (3.6)&15.0 (0.5) & 13.9 (1.3) & 14.0 (1.3)\\ \hline Mn2 Nonlinear & 24 & 96 & 23.8 (8.0) & 20.5 (8.4) & 25.3 (11.8)&24.0 (3.1) & 19.8 (5.0) & 19.8 (5.0)&23.3 (1.3) & 18.3 (2.7) & 18.3 (2.7)&21.0 (0.7) & 18.6 (1.5) & 19.5 (2.5)\\ \hline Mp1 Paraboloid & 3 & 12 & 2.9 (0.3) & 2.9 (0.3) & 2.8 (0.2)&2.9 (0.2) & 2.9 (0.2) & 2.9 (0.2)&2.9 (0.2) & 2.9 (0.2) & 2.9 (0.2)&3.0 (0.1) & 2.9 (0.1) & 2.9 (0.1)\\ \hline Mp2 Paraboloid & 6 & 21 & 5.2 (0.9) & 5.2 (0.9) & 4.9 (0.6)&5.1 (0.7) & 5.1 (0.7) & 5.1 (0.7)&5.3 (0.4) & 5.3 (0.4) & 5.3 (0.4)&5.4 (0.3) & 5.1 (0.2) & 5.2 (0.2)\\ \hline Mp3 Paraboloid & 9 & 30 & 6.9 (1.4) & 6.9 (1.4) & 5.6 (0.7)&7.0 (0.7) & 7.0 (0.7) & 7.0 (0.7)&7.3 (0.7) & 7.3 (0.7) & 7.3 (0.7)&7.5 (0.6) & 7.1 (0.4) & 7.2 (0.4)\\ \hline
    \end{tabular}

\caption{\footnotesize \textbf{Number of samples}. KNN has good performance for low dimensional manifolds even with few points. \textbf{High dimensional datasets}. For high dimensional datasets such as M10 Cubics, the error in the estimated dimension is even high, and the estimated dimension can even be negative (see comment in \Cref{rmk:ph_knn_error}). 
\textbf{Variance}. For the same reasons outlined in \Cref{rmk:ph_knn_error}, the variance can be high for high dimensional datasets, such as the M10 Cubics. \textbf{Hyperparameter dependency}. The differences between different hyperparameter choices is small for most datasets, apart from high dimensional ones.  \textbf{Nonlinear datasets}. KNN has the tendency to underestimate on the nonlinear datasets, such as Mp2,3 Paraboloids, and Mn1, 2 Nonlinear. \label{tab:knn_bm} }
\end{sidewaystable}

%% file: tables/DanCoNEW.tex
\begin{sidewaystable}[h]
\centering
\tiny
\begin{tabular}{|l|ll||lll|lll|lll|lll|}
\hline            

&    &    & DanCo &              &              &              &              &              &              &              &              &              &              &              \\  \hline             &    &    &              &              &              &              &              &              &              &              &              &              &              &              \\  \hline              &    &    & 625          &              &              & 1250         &              &              & 2500         &              &              & 5000         &              &              \\  \hline              &    &    &              &              &              &              &              &              &              &              &              &              &              &              \\  \hline              & d  & n  & Best         & med abs      & med rel      & Best         & med abs      & med rel      & Best         & med abs      & med rel      & Best         & med abs      & med rel      \\  \hline              &    &    &              &              &              &              &              &              &              &              &              &              &              &              \\  \hline               M1Sphere & 10 & 11 & 10.7 (0.43) & 10.7 (0.43) & 10.8 (0.35)&10.7 (0.41) & 11.0 (0.21) & 10.9 (0.23)&10.9 (0.34) & 11.0 (0.0) & 11.0 (0.0)&10.5 (1.96) & 11.0 (0.0) & 11.0 (0.0)\\ \hline M2Affine3to5 & 3 & 5 & 3.0 (0.1) & 2.9 (0.08) & 2.8 (0.08)&3.0 (0.07) & 2.9 (0.06) & 2.9 (0.08)&3.0 (0.04) & 2.9 (0.05) & 2.9 (0.05)&3.0 (0.04) & 2.9 (0.04) & 2.9 (0.02)\\ \hline M3Nonlinear4to6 & 4 & 6 & 4.3 (0.07) & 4.8 (0.31) & 4.3 (0.07)&4.6 (0.18) & 5.0 (0.39) & 4.7 (0.26)&3.5 (1.65) & 5.0 (0.2) & 5.0 (0.2)&4.8 (0.19) & 5.2 (0.33) & 4.8 (0.19)\\ \hline M4Nonlinear & 4 & 8 & 4.5 (2.54) & 5.1 (0.25) & 4.8 (0.2)&3.6 (2.55) & 5.1 (0.27) & 4.9 (0.16)&4.0 (2.47) & 5.5 (0.29) & 5.5 (0.29)&3.8 (1.92) & 5.4 (0.23) & 5.1 (0.13)\\ \hline M5aHelix1d & 1 & 3 & 1.0 (0.0) & 1.0 (0.0) & 1.0 (0.0)&1.0 (0.0) & 1.0 (0.0) & 1.0 (0.0)&1.0 (0.0) & 1.0 (0.0) & 1.0 (0.0)&1.0 (0.0) & 1.0 (0.0) & 1.0 (0.0)\\ \hline M5bHelix2d & 2 & 3 & 1.9 (0.91) & 3.0 (0.0) & 3.0 (0.0)&2.1 (0.92) & 3.0 (0.0) & 3.0 (0.0)&1.9 (0.84) & 3.0 (0.0) & 3.0 (0.0)&2.1 (1.01) & 3.4 (0.04) & 3.4 (0.46)\\ \hline M6Nonlinear & 6 & 36 & 7.7 (0.33) & 7.7 (0.33) & 8.1 (0.22)&7.3 (0.3) & 7.4 (0.34) & 7.4 (0.3)&7.1 (0.05) & 7.4 (0.22) & 7.4 (0.22)&7.0 (0.02) & 7.7 (0.22) & 7.1 (0.11)\\ \hline M7Roll & 2 & 3 & 2.1 (0.78) & 2.3 (0.02) & 3.5 (0.14)&2.2 (0.02) & 2.3 (0.02) & 3.1 (0.03)&2.2 (0.01) & 2.3 (0.01) & 2.3 (0.01)&2.2 (0.01) & 2.2 (0.01) & 3.0 (0.24)\\ \hline M8Nonlinear & 12 & 72 & 18.0 (0.73) & 18.0 (0.73) & 18.3 (0.83)&17.4 (4.0) & 17.8 (0.56) & 17.9 (0.3)&12.4 (0.56) & 17.3 (0.53) & 17.3 (0.53)&12.3 (0.45) & 16.9 (0.21) & 17.0 (0.21)\\ \hline M9Affine & 20 & 20 & 19.7 (0.46) & 19.6 (0.58) & 19.3 (0.45)&20.0 (0.22) & 19.5 (0.49) & 18.5 (3.81)&19.9 (0.3) & 19.3 (0.46) & 19.3 (0.46)&19.9 (0.3) & 18.2 (3.72) & 19.1 (0.22)\\ \hline M10aCubic & 10 & 11 & 10.1 (0.3) & 10.4 (0.48) & 10.3 (0.45)&10.1 (0.29) & 10.1 (0.29) & 10.2 (0.35)&10.0 (0.02) & 10.1 (0.29) & 10.1 (0.29)&10.0 (0.01) & 10.1 (0.29) & 10.0 (0.01)\\ \hline M10bCubic & 17 & 18 & 17.2 (0.59) & 17.4 (0.49) & 17.2 (0.59)&17.0 (0.21) & 17.2 (0.47) & 17.0 (0.21)&17.0 (0.01) & 17.2 (0.36) & 17.2 (0.36)&17.0 (0.01) & 17.0 (0.01) & 17.0 (0.01)\\ \hline M10cCubic & 24 & 25 & 23.9 (0.62) & 24.6 (0.5) & 23.9 (0.62)&23.9 (0.5) & 24.2 (0.57) & 24.1 (0.22)&24.0 (0.0) & 24.0 (0.44) & 24.0 (0.44)&24.0 (0.0) & 24.0 (0.0) & 24.0 (0.0)\\ \hline M10dCubic & 70 & 72 & 70.6 (0.8) & 70.9 (0.36) & 68.2 (2.01)&70.7 (0.48) & 70.8 (0.6) & 70.9 (0.3)&70.8 (0.6) & 70.8 (0.6) & 70.8 (0.6)&69.6 (1.12) & 71.0 (0.0) & 71.0 (0.0)\\ \hline M11Moebius & 2 & 3 & 2.3 (0.04) & 2.7 (0.32) & 3.6 (0.43)&1.7 (0.95) & 2.3 (0.02) & 3.2 (0.04)&2.2 (0.01) & 2.3 (0.01) & 2.3 (0.01)&2.2 (0.01) & 2.3 (0.01) & 3.1 (0.01)\\ \hline M12Norm & 20 & 20 & 19.9 (0.23) & 19.9 (0.35) & 19.9 (0.35)&20.0 (0.0) & 20.0 (0.0) & 20.0 (0.0)&20.0 (0.0) & 20.0 (0.0) & 20.0 (0.0)&20.0 (0.0) & 20.0 (0.0) & 20.0 (0.0)\\ \hline M13aScurve & 2 & 3 & 2.2 (0.03) & 2.3 (0.02) & 3.5 (0.04)&2.2 (0.02) & 2.3 (0.02) & 3.0 (0.24)&2.2 (0.01) & 2.3 (0.01) & 2.3 (0.01)&2.2 (0.01) & 2.2 (0.01) & 3.0 (0.17)\\ \hline M13bSpiral & 1 & 13 & 1.2 (1.05) & 2.2 (0.04) & 2.3 (0.04)&1.2 (0.09) & 1.4 (0.17) & 2.1 (0.16)&1.0 (0.0) & 1.0 (0.04) & 1.0 (0.04)&1.0 (0.0) & 1.0 (0.0) & 1.0 (0.0)\\ \hline Mbeta & 10 & 40 & 7.1 (0.17) & 6.3 (0.16) & 6.3 (0.21)&7.6 (0.28) & 6.7 (0.16) & 6.4 (0.12)&8.1 (0.1) & 7.0 (0.03) & 7.0 (0.03)&8.8 (0.28) & 7.3 (0.1) & 7.0 (0.02)\\ \hline Mn1Nonlinear & 18 & 72 & 17.9 (0.66) & 17.9 (0.66) & 17.7 (0.53)&17.9 (0.38) & 17.7 (0.53) & 17.7 (0.43)&17.9 (0.38) & 17.9 (0.38) & 17.9 (0.38)&17.8 (0.35) & 17.7 (0.43) & 17.8 (0.35)\\ \hline Mn2Nonlinear & 24 & 96 & 24.0 (0.73) & 24.0 (0.73) & 23.7 (0.72)&24.1 (0.53) & 23.9 (0.76) & 24.1 (0.53)&23.8 (0.47) & 23.8 (0.62) & 23.8 (0.62)&24.0 (0.44) & 24.0 (0.44) & 23.8 (0.35)\\ \hline Mp1Paraboloid & 3 & 12 & 3.1 (0.21) & 3.3 (0.2) & 3.1 (0.21)&3.2 (0.12) & 3.3 (0.12) & 3.2 (0.14)&3.1 (0.08) & 3.2 (0.09) & 3.2 (0.09)&3.1 (0.04) & 3.2 (0.07) & 3.2 (0.07)\\ \hline Mp2Paraboloid & 6 & 21 & 6.3 (0.4) & 5.7 (0.2) & 4.8 (0.16)&5.9 (0.13) & 5.9 (0.13) & 5.6 (0.12)&5.9 (0.12) & 6.3 (0.17) & 6.3 (0.17)&6.0 (0.03) & 6.6 (0.17) & 6.2 (0.11)\\ \hline Mp3Paraboloid & 9 & 30 & 8.4 (9.19) & 6.2 (0.17) & 5.6 (0.19)&8.0 (0.11) & 7.0 (0.11) & 6.6 (0.19)&8.9 (0.21) & 7.7 (0.21) & 7.7 (0.21)&9.1 (0.16) & 8.0 (0.01) & 7.7 (0.16)\\ \hline   \end{tabular}
\label{tab:danco_bm}
\caption{\footnotesize \textbf{Number of samples}. Fairly accurate even with few points, except with some datasets with non-linearities, such as M3Nonlinear4to6, and M3Nonlinear, M13b spiral.  \textbf{High dimensional datasets}. Performs very well on the cubic datasets, even with few samples. Also on M12 Norm. 
\textbf{Variance}. Moderate variance, even with many point samples on some datasets such as M3Nonlinear 4to6, and M4Nonlinear. Note that the variance can also depend on the choice of hyperparameters; see for 5000 samples, M1Sphere, M6 Nonlinear, and Mp2,3  paraboloids. \textbf{Hyperparameter dependency}. The estimate  can vary depending on the hyperparameters on some datasets with nonlinearities and curvature, such as Mbeta, M5bhelix2d , M11 Moebius, M13ab, Mp3 paraboloid, and M4 M8 Nonlinear. Interestingly, the hyperparameter dependency can be worse with more samples; for example on M8 Nonlinear.  \textbf{Nonlinear datasets}. The performance can be dependent on the choice of hyperparameters on the nonlinear datasets. The performance is good on some  datasets with non-uniform sampling densities such as M12 Norm and Mn1,2 Nonlinear, but can be challenged by datasets with curvature.   }

\end{sidewaystable}

%% file: tables/MLE_benchmark.tex
\begin{sidewaystable}[h]
\centering
\tiny
\begin{tabular}{|l|ll||lll|lll|lll|lll|}
\hline               &    &    & MLE &              &              &              &              &              &              &              &              &              &              &              \\  \hline              &    &    &              &              &              &              &              &              &              &              &              &              &              &              \\  \hline             &    &    & 625          &              &              & 1250         &              &              & 2500         &              &              & 5000         &              &              \\  \hline               &    &    &              &              &              &              &              &              &              &              &              &              &              &              \\  \hline              & d  & n  & Best         & med abs      & med rel      & Best         & med abs      & med rel      & Best         & med abs      & med rel      & Best         & med abs      & med rel      \\  \hline              &    &    &              &              &              &              &              &              &              &              &              &              &              &              \\  \hline                M1Sphere & 10 & 11 & 9.9 (0.2) & 9.2 (0.21) & 9.9 (0.2)&10.1 (0.11) & 9.3 (0.1) & 10.1 (0.11)&10.3 (0.07) & 9.5 (0.08) & 10.3 (0.07)&9.7 (0.03) & 9.7 (0.04) & 9.7 (0.04)\\ \hline M2Affine3to5 & 3 & 5 & 2.9 (0.02) & 2.9 (0.06) & 3.2 (0.04)&3.0 (0.02) & 3.0 (0.04) & 3.2 (0.03)&3.0 (0.02) & 3.0 (0.02) & 3.3 (0.02)&3.0 (0.02) & 3.0 (0.02) & 3.0 (0.02)\\ \hline M3Nonlinear4to6 & 4 & 6 & 3.9 (0.05) & 3.9 (0.08) & 4.3 (0.09)&4.0 (0.05) & 3.9 (0.05) & 4.3 (0.06)&4.0 (0.04) & 4.0 (0.04) & 4.4 (0.04)&4.0 (0.02) & 4.0 (0.02) & 4.0 (0.02)\\ \hline M4Nonlinear & 4 & 8 & 4.1 (0.07) & 4.3 (0.08) & 4.7 (0.1)&4.0 (0.04) & 4.1 (0.06) & 4.5 (0.05)&4.0 (0.03) & 4.1 (0.03) & 4.4 (0.03)&4.0 (0.02) & 4.1 (0.02) & 4.1 (0.02)\\ \hline M5aHelix1d & 1 & 3 & 1.0 (0.02) & 1.0 (0.02) & 1.1 (0.01)&1.0 (0.01) & 1.0 (0.01) & 1.1 (0.01)&1.0 (0.0) & 1.0 (0.01) & 1.1 (0.01)&1.0 (0.0) & 1.0 (0.01) & 1.0 (0.01)\\ \hline M5bHelix2d & 2 & 3 & 2.6 (0.04) & 2.8 (0.06) & 3.1 (0.05)&2.6 (0.04) & 2.7 (0.04) & 3.1 (0.04)&2.4 (0.02) & 2.5 (0.03) & 2.9 (0.03)&2.2 (0.01) & 2.3 (0.02) & 2.3 (0.02)\\ \hline M6Nonlinear & 6 & 36 & 6.6 (0.15) & 7.0 (0.17) & 7.7 (0.16)&6.3 (0.1) & 6.7 (0.11) & 7.3 (0.11)&6.2 (0.05) & 6.5 (0.06) & 7.1 (0.05)&6.0 (0.04) & 6.3 (0.06) & 6.3 (0.06)\\ \hline M7Roll & 2 & 3 & 2.0 (0.04) & 2.0 (0.04) & 2.2 (0.03)&2.0 (0.01) & 2.0 (0.02) & 2.2 (0.02)&2.0 (0.01) & 2.0 (0.02) & 2.2 (0.02)&2.0 (0.01) & 2.0 (0.01) & 2.0 (0.01)\\ \hline M8Nonlinear & 12 & 72 & 12.0 (0.08) & 13.9 (0.24) & 15.1 (0.22)&12.5 (0.05) & 14.1 (0.19) & 15.4 (0.21)&13.0 (0.04) & 14.1 (0.13) & 15.4 (0.15)&13.3 (0.03) & 14.2 (0.07) & 14.2 (0.07)\\ \hline M9Affine & 20 & 20 & 15.8 (0.24) & 14.6 (0.24) & 15.8 (0.24)&16.2 (0.19) & 15.0 (0.23) & 16.2 (0.19)&16.6 (0.15) & 15.4 (0.17) & 16.6 (0.15)&17.0 (0.12) & 15.7 (0.11) & 15.7 (0.11)\\ \hline M10aCubic & 10 & 11 & 9.5 (0.13) & 8.8 (0.14) & 9.5 (0.13)&9.7 (0.14) & 9.0 (0.15) & 9.7 (0.14)&10.0 (0.06) & 9.2 (0.07) & 10.0 (0.06)&10.1 (0.05) & 9.3 (0.06) & 9.3 (0.06)\\ \hline M10bCubic & 17 & 18 & 14.4 (0.23) & 13.4 (0.22) & 14.4 (0.23)&14.8 (0.15) & 13.6 (0.17) & 14.8 (0.15)&15.2 (0.15) & 14.0 (0.16) & 15.2 (0.15)&15.4 (0.1) & 14.3 (0.1) & 14.3 (0.1)\\ \hline M10cCubic & 24 & 25 & 18.4 (0.28) & 17.1 (0.25) & 18.4 (0.28)&19.3 (0.24) & 17.8 (0.21) & 19.3 (0.24)&19.8 (0.17) & 18.3 (0.17) & 19.8 (0.17)&20.2 (0.1) & 18.7 (0.14) & 18.7 (0.14)\\ \hline M10dCubic & 70 & 72 & 37.8 (0.74) & 35.2 (0.56) & 37.8 (0.74)&39.8 (0.51) & 36.9 (0.47) & 39.8 (0.51)&41.6 (0.34) & 38.5 (0.4) & 41.6 (0.34)&43.2 (0.32) & 39.9 (0.24) & 39.9 (0.24)\\ \hline M11Moebius & 2 & 3 & 2.0 (0.03) & 2.1 (0.04) & 2.2 (0.03)&2.0 (0.02) & 2.0 (0.03) & 2.2 (0.02)&2.0 (0.01) & 2.1 (0.02) & 2.2 (0.02)&2.0 (0.01) & 2.1 (0.01) & 2.1 (0.01)\\ \hline M12Norm & 20 & 20 & 16.5 (0.3) & 15.3 (0.32) & 16.5 (0.3)&17.4 (0.15) & 16.0 (0.19) & 17.4 (0.15)&18.0 (0.16) & 16.6 (0.13) & 18.0 (0.16)&18.6 (0.1) & 17.2 (0.13) & 17.2 (0.13)\\ \hline M13aScurve & 2 & 3 & 2.0 (0.04) & 2.0 (0.04) & 2.2 (0.03)&2.0 (0.03) & 2.0 (0.03) & 2.2 (0.02)&2.0 (0.01) & 2.0 (0.02) & 2.2 (0.01)&2.0 (0.01) & 2.1 (0.01) & 2.1 (0.01)\\ \hline M13bSpiral & 1 & 13 & 1.4 (0.01) & 1.8 (0.05) & 2.0 (0.04)&1.5 (0.03) & 1.6 (0.03) & 2.0 (0.02)&1.1 (0.01) & 1.1 (0.01) & 1.3 (0.01)&1.0 (0.0) & 1.0 (0.01) & 1.0 (0.01)\\ \hline Mbeta & 10 & 40 & 6.7 (0.11) & 6.0 (0.13) & 6.7 (0.11)&7.0 (0.08) & 6.3 (0.1) & 7.0 (0.08)&7.1 (0.04) & 6.4 (0.05) & 7.1 (0.04)&7.3 (0.06) & 6.6 (0.05) & 6.6 (0.05)\\ \hline Mn1Nonlinear & 18 & 72 & 14.7 (0.23) & 13.6 (0.26) & 14.7 (0.23)&15.1 (0.14) & 14.0 (0.17) & 15.1 (0.14)&15.6 (0.13) & 14.4 (0.13) & 15.6 (0.13)&15.8 (0.11) & 14.6 (0.1) & 14.6 (0.1)\\ \hline Mn2Nonlinear & 24 & 96 & 18.2 (0.22) & 16.9 (0.17) & 18.2 (0.22)&18.9 (0.21) & 17.5 (0.16) & 18.9 (0.21)&19.5 (0.18) & 18.1 (0.19) & 19.5 (0.18)&20.0 (0.15) & 18.5 (0.15) & 18.5 (0.15)\\ \hline Mp1Paraboloid & 3 & 12 & 2.9 (0.05) & 2.9 (0.05) & 3.2 (0.04)&3.0 (0.06) & 3.0 (0.06) & 3.2 (0.04)&3.0 (0.03) & 3.0 (0.03) & 3.3 (0.02)&3.0 (0.01) & 3.0 (0.02) & 3.0 (0.02)\\ \hline Mp2Paraboloid & 6 & 21 & 5.2 (0.09) & 4.9 (0.08) & 5.2 (0.09)&5.5 (0.06) & 5.1 (0.06) & 5.5 (0.06)&5.7 (0.04) & 5.3 (0.04) & 5.7 (0.04)&5.9 (0.03) & 5.5 (0.03) & 5.5 (0.03)\\ \hline Mp3Paraboloid & 9 & 30 & 6.4 (0.13) & 6.0 (0.14) & 6.4 (0.13)&7.0 (0.09) & 6.5 (0.09) & 7.0 (0.09)&7.5 (0.08) & 6.9 (0.06) & 7.5 (0.08)&7.8 (0.06) & 7.2 (0.06) & 7.2 (0.06)\\ \hline 

\end{tabular}
\caption{\footnotesize \textbf{Number of samples}. Accurate on some low dimensional datasets with few samples, and needs more samples on those with challenging geometries such as M5b Helix2d, M6 Nonlinear, and M13b spiral.  \textbf{High dimensional datasets}. Performs poorly, and tends to underestimate on the M9 Affine and M10c-d Cubic datasets. 
\textbf{Variance}. Generally low variance and good consistency between samples. \textbf{Hyperparameter dependency}. The estimate is mostly insensitive to the choice of neighbourhood size, apart from some nonlinear datasets such as M8 Nonlinear and Mn1,2 Nonlinear, and moderately high dimensional ones such as M9 Affine, M12Norm, and the M9 Cubic datasets. \textbf{Nonlinear datasets}. There is a tendency to underestimate on nonlinear datasets such as Mn1,2 Nonlinear and Mp3 Paraboloid.  }
\label{tab:mle_bm}
\end{sidewaystable}

%% file: tables/MINDML_benchmark.tex
\begin{sidewaystable}[h]
\centering
\tiny
\begin{tabular}{|l|ll||lll|lll|lll|lll|}
\hline               &    &    & MiND\_ML &              &              &              &              &              &              &              &              &              &              &              \\  \hline               &    &    &              &              &              &              &              &              &              &              &              &              &              &              \\  \hline             &    &    & 625          &              &              & 1250         &              &              & 2500         &              &              & 5000         &              &              \\  \hline               &    &    &              &              &              &              &              &              &              &              &              &              &              &              \\  \hline               & d  & n  & Best         & med abs      & med rel      & Best         & med abs      & med rel      & Best         & med abs      & med rel      & Best         & med abs      & med rel      \\  \hline              &    &    &              &              &              &              &              &              &              &              &              &              &              &              \\  \hline               M1Sphere & 10 & 11 & 9.3 (0.52) & 9.3 (0.52) & 9.2 (0.62)&9.4 (0.5) & 9.4 (0.24) & 9.4 (0.24)&9.4 (0.5) & 9.4 (0.21) & 9.4 (0.21)&9.6 (0.48) & 9.6 (0.48) & 9.5 (0.11)\\ \hline 
M2 Affine 3to5 & 3 & 5 & 3.0 (0.0) & 3.0 (0.11) & 3.0 (0.0)&3.0 (0.0) & 3.0 (0.12) & 3.0 (0.12)&3.0 (0.0) & 3.0 (0.07) & 3.0 (0.07)&3.0 (0.0) & 3.0 (0.0) & 2.9 (0.05)\\ \hline 
M3 Nonlinear 4to6 & 4 & 6 & 4.0 (0.0) & 3.8 (0.18) & 4.0 (0.22)&4.0 (0.0) & 3.8 (0.14) & 3.8 (0.14)&4.0 (0.0) & 3.9 (0.1) & 3.9 (0.1)&4.0 (0.0) & 4.0 (0.0) & 3.9 (0.08)\\ \hline 
M4 Nonlinear & 4 & 8 & 4.0 (0.0) & 3.9 (0.22) & 4.0 (0.0)&4.0 (0.0) & 3.9 (0.13) & 3.9 (0.13)&4.0 (0.0) & 3.9 (0.1) & 3.9 (0.1)&4.0 (0.0) & 4.0 (0.0) & 3.9 (0.08)\\ \hline 
M5a Helix1d & 1 & 3 & 1.0 (0.0) & 1.0 (0.03) & 1.0 (0.0)&1.0 (0.0) & 1.0 (0.01) & 1.0 (0.01)&1.0 (0.0) & 1.0 (0.01) & 1.0 (0.01)&1.0 (0.0) & 1.0 (0.0) & 1.0 (0.01)\\ \hline 
M5b Helix2d & 2 & 3 & 2.0 (0.22) & 2.3 (0.12) & 2.0 (0.22)&2.0 (0.0) & 2.1 (0.07) & 2.1 (0.07)&2.0 (0.0) & 2.0 (0.05) & 2.0 (0.05)&2.0 (0.0) & 2.0 (0.0) & 2.0 (0.02)\\ \hline 
M6 Nonlinear & 6 & 36 & 6.1 (0.44) & 6.2 (0.32) & 6.1 (0.44)&6.0 (0.0) & 6.0 (0.22) & 6.0 (0.22)&6.0 (0.0) & 5.9 (0.15) & 5.9 (0.15)&6.0 (0.0) & 6.0 (0.0) & 5.9 (0.1)\\ \hline 
M7 Roll & 2 & 3 & 2.0 (0.0) & 2.0 (0.05) & 2.0 (0.0)&2.0 (0.0) & 2.0 (0.06) & 2.0 (0.06)&2.0 (0.0) & 2.0 (0.03) & 2.0 (0.03)&2.0 (0.0) & 2.0 (0.0) & 2.0 (0.02)\\ \hline 
M8 Nonlinear & 12 & 72 & 13.2 (0.43) & 13.4 (0.74) & 13.4 (0.8)&13.5 (0.67) & 10.0 (0.0) & 13.5 (0.59)&13.2 (0.4) & 13.3 (0.35) & 10.0 (0.0)&13.0 (0.22) & 10.0 (0.0) & 10.0 (0.0)\\ \hline 
M9 Affine & 20 & 20 & 15.2 (0.69) & 15.2 (0.69) & 15.2 (0.75)&15.4 (0.51) & 10.0 (0.0) & 15.4 (0.51)&15.6 (0.37) & 15.6 (0.37) & 10.0 (0.0)&16.0 (0.22) & 10.0 (0.0) & 10.0 (0.0)\\ \hline 
M10a Cubic & 10 & 11 & 9.0 (0.35) & 9.0 (0.35) & 9.0 (0.38)&9.1 (0.34) & 9.1 (0.31) & 9.1 (0.34)&9.2 (0.14) & 9.2 (0.14) & 9.2 (0.14)&9.3 (0.12) & 9.0 (0.22) & 9.3 (0.12)\\ \hline 
M10b Cubic & 17 & 18 & 13.8 (0.75) & 13.8 (0.62) & 13.8 (0.75)&13.7 (0.56) & 10.0 (0.0) & 13.7 (0.41)&14.2 (0.35) & 14.2 (0.35) & 10.0 (0.0)&14.3 (0.29) & 10.0 (0.0) & 10.0 (0.0)\\ \hline 
M10c Cubic & 24 & 25 & 18.0 (0.77) & 18.0 (0.66) & 18.0 (0.77)&18.2 (0.58) & 10.0 (0.0) & 18.2 (0.58)&18.6 (0.49) & 18.6 (0.49) & 10.0 (0.0)&19.0 (0.22) & 10.0 (0.0) & 10.0 (0.0)\\ \hline 
M10d Cubic & 70 & 72 & 38.5 (1.9) & 20.0 (0.0) & 20.0 (0.0)&39.4 (1.07) & 10.0 (0.0) & 20.0 (0.0)&40.2 (0.84) & 30.0 (0.0) & 10.0 (0.0)&41.9 (0.7) & 10.0 (0.0) & 10.0 (0.0)\\ \hline 
M11 Moebius & 2 & 3 & 2.0 (0.0) & 2.0 (0.05) & 2.0 (0.0)&2.0 (0.0) & 2.0 (0.04) & 2.0 (0.04)&2.0 (0.0) & 2.0 (0.02) & 2.0 (0.02)&2.0 (0.0) & 2.0 (0.0) & 2.0 (0.02)\\ \hline 
M12 Norm & 20 & 20 & 16.2 (0.61) & 16.2 (0.61) & 16.2 (0.75)&16.8 (0.52) & 10.0 (0.0) & 16.8 (0.52)&17.2 (0.43) & 17.2 (0.31) & 10.0 (0.0)&17.5 (0.27) & 10.0 (0.0) & 10.0 (0.0)\\ \hline 
M13a Scurve & 2 & 3 & 2.0 (0.0) & 2.0 (0.05) & 2.0 (0.0)&2.0 (0.0) & 2.0 (0.05) & 2.0 (0.05)&2.0 (0.0) & 2.0 (0.03) & 2.0 (0.03)&2.0 (0.0) & 2.0 (0.0) & 2.0 (0.02)\\ \hline 
M13b Spiral & 1 & 13 & 1.0 (0.0) & 1.1 (0.05) & 1.0 (0.0)&1.0 (0.0) & 1.0 (0.02) & 1.0 (0.02)&1.0 (0.0) & 1.0 (0.01) & 1.0 (0.01)&1.0 (0.0) & 1.0 (0.0) & 1.0 (0.01)\\ \hline 
Mbeta & 10 & 40 & 6.2 (0.27) & 6.2 (0.27) & 6.2 (0.36)&6.3 (0.21) & 6.3 (0.21) & 6.3 (0.21)&6.6 (0.5) & 6.5 (0.17) & 6.5 (0.17)&6.8 (0.36) & 6.8 (0.36) & 6.6 (0.11)\\ \hline 
Mn1 Nonlinear & 18 & 72 & 14.0 (0.61) & 14.0 (0.61) & 13.9 (0.7)&14.4 (0.48) & 10.0 (0.0) & 14.3 (0.42)&14.6 (0.35) & 14.6 (0.35) & 10.0 (0.0)&14.8 (0.36) & 10.0 (0.0) & 10.0 (0.0)\\ \hline 
Mn2 Nonlinear & 24 & 96 & 17.8 (0.83) & 17.7 (0.83) & 17.8 (0.83)&18.3 (0.71) & 10.0 (0.0) & 18.2 (0.69)&18.3 (0.54) & 18.3 (0.54) & 10.0 (0.0)&18.8 (0.51) & 10.0 (0.0) & 10.0 (0.0)\\ \hline 
Mp1 Paraboloid & 3 & 12 & 3.0 (0.0) & 2.9 (0.13) & 3.0 (0.0)&3.0 (0.0) & 3.0 (0.07) & 3.0 (0.07)&3.0 (0.0) & 3.0 (0.09) & 3.0 (0.09)&3.0 (0.0) & 3.0 (0.0) & 3.0 (0.04)\\ \hline 
Mp2 Paraboloid & 6 & 21 & 5.1 (0.18) & 5.1 (0.18) & 5.0 (0.22)&5.3 (0.19) & 5.3 (0.19) & 5.3 (0.19)&5.4 (0.17) & 5.4 (0.17) & 5.4 (0.17)&5.5 (0.08) & 5.4 (0.5) & 5.5 (0.08)\\ \hline 
Mp3 Paraboloid & 9 & 30 & 6.6 (0.33) & 6.6 (0.33) & 6.5 (0.5)&7.0 (0.0) & 7.0 (0.21) & 7.0 (0.21)&7.2 (0.17) & 7.2 (0.17) & 7.2 (0.17)&7.4 (0.13) & 7.4 (0.48) & 7.4 (0.13)\\  \hline    
\end{tabular} 
\caption{\footnotesize \textbf{Number of samples}. On most datasets, especially those of small dimension $< 7$, MiND\_ML is quite accurate even with few samples; for higher dimensional datasets where the estimate is inaccurate, adding more points does not appear to improve the estimation significantly. 
\textbf{High dimensional datasets}. MiND\_ML severely underestimates high dimensional datasets, such as M9 Affine and M10a,b,c,d Cubics. 
\textbf{Variance}. The variance is low in general, though higher dimensional datasets such as the M10b,c,d Cubics, and non-linear datasets such as Mn1,2 Nonlinear, can induce higher variances  \textbf{Hyperparameter dependency}. On low dimensional datasets, MiND\_ML has good agreement between Best, med abs, and med rel estimates, yet on higher dimensional datasets sucha as M10b,c,d Cubics, and nonlinear datasets such as M8 Nonlinear, a poor choice of the number of neighbours parameter can lead to a vast under-estimation of the dimension. \textbf{Nonlinear datasets}. 
MiND\_ML tends to underestimate on Mp2,3 paraboloids, as well as Mn1,2 Nonlinear.  }

\label{tab:mind_bm}
\end{sidewaystable}

%% file: tables/GRIDE_benchmark.tex
\begin{sidewaystable}[h]
\centering
\tiny
\begin{tabular}{|l|ll||lll|lll|lll|lll|}
\hline
               &    &    & GRIDE &              &              &              &              &              &              &              &              &              &              &              \\  \hline
               &    &    &              &              &              &              &              &              &              &              &              &              &              &              \\  \hline
n              &    &    & 625          &              &              & 1250         &              &              & 2500         &              &              & 5000         &              &              \\  \hline
               &    &    &              &              &              &              &              &              &              &              &              &              &              &              \\  \hline
               & d  & n  & Best         & med abs      & med rel      & Best         & med abs      & med rel      & Best         & med abs      & med rel      & Best         & med abs      & med rel      \\  \hline
               &    &    &              &              &              &              &              &              &              &              &              &              &              &              \\  \hline 
               M1\_Sphere & 10 & 11 & 9.2 (0.52) & 9.2 (0.52) & 9.2 (0.52)&9.4 (0.24) & 9.4 (0.24) & 9.4 (0.24)&9.4 (0.2) & 9.4 (0.2) & 9.4 (0.2)&9.5 (0.11) & 9.5 (0.11) & 9.5 (0.11)\\ \hline M2\_Affine\_3to5 & 3 & 5 & 2.9 (0.12) & 2.9 (0.12) & 2.9 (0.12)&2.9 (0.12) & 2.9 (0.12) & 2.9 (0.12)&2.9 (0.07) & 2.9 (0.07) & 2.9 (0.07)&2.9 (0.05) & 2.9 (0.05) & 2.9 (0.05)\\ \hline M3\_Nonlinear\_4to6 & 4 & 6 & 3.8 (0.18) & 3.8 (0.18) & 3.8 (0.18)&3.8 (0.07) & 3.8 (0.14) & 3.8 (0.14)&3.9 (0.1) & 3.9 (0.1) & 3.9 (0.1)&3.9 (0.07) & 3.9 (0.08) & 3.9 (0.08)\\ \hline M4\_Nonlinear & 4 & 8 & 4.0 (0.11) & 3.9 (0.22) & 3.9 (0.22)&4.0 (0.05) & 3.9 (0.13) & 3.9 (0.13)&4.0 (0.04) & 3.9 (0.1) & 3.9 (0.1)&4.0 (0.01) & 3.9 (0.08) & 3.9 (0.08)\\ \hline M5a\_Helix1d & 1 & 3 & 1.0 (0.01) & 1.0 (0.03) & 1.0 (0.03)&1.0 (0.01) & 1.0 (0.02) & 1.0 (0.02)&1.0 (0.0) & 1.0 (0.02) & 1.0 (0.02)&1.0 (0.0) & 1.0 (0.02) & 1.0 (0.02)\\ \hline M5b\_Helix2d & 2 & 3 & 2.3 (0.11) & 2.3 (0.11) & 2.3 (0.11)&2.1 (0.09) & 2.1 (0.09) & 2.1 (0.09)&2.0 (0.06) & 2.0 (0.06) & 2.0 (0.06)&2.0 (0.02) & 2.0 (0.04) & 2.0 (0.04)\\ \hline M6\_Nonlinear & 6 & 36 & 6.2 (0.32) & 6.2 (0.32) & 6.2 (0.32)&6.0 (0.22) & 6.0 (0.22) & 6.0 (0.22)&6.0 (0.09) & 5.9 (0.15) & 5.9 (0.15)&6.0 (0.06) & 5.9 (0.1) & 5.9 (0.1)\\ \hline M7\_Roll & 2 & 3 & 2.0 (0.02) & 2.0 (0.08) & 2.0 (0.08)&2.0 (0.09) & 2.0 (0.09) & 2.0 (0.09)&2.0 (0.05) & 2.0 (0.05) & 2.0 (0.05)&2.0 (0.04) & 2.0 (0.04) & 2.0 (0.04)\\ \hline M8\_Nonlinear & 12 & 72 & 12.1 (0.09) & 13.4 (0.74) & 13.4 (0.74)&12.1 (0.05) & 13.5 (0.59) & 13.5 (0.59)&12.5 (0.04) & 13.3 (0.35) & 13.3 (0.35)&13.0 (0.02) & 13.2 (0.22) & 13.2 (0.22)\\ \hline M9\_Affine & 20 & 20 & 15.2 (0.69) & 15.2 (0.69) & 15.2 (0.69)&15.4 (0.51) & 15.4 (0.51) & 15.4 (0.51)&15.6 (0.37) & 15.6 (0.37) & 15.6 (0.37)&15.9 (0.25) & 15.9 (0.25) & 15.9 (0.25)\\ \hline M10a\_Cubic & 10 & 11 & 9.0 (0.35) & 9.0 (0.35) & 9.0 (0.35)&9.1 (0.34) & 9.1 (0.34) & 9.1 (0.34)&9.2 (0.14) & 9.2 (0.14) & 9.2 (0.14)&9.3 (0.12) & 9.3 (0.12) & 9.3 (0.12)\\ \hline M10b\_Cubic & 17 & 18 & 13.8 (0.62) & 13.8 (0.62) & 13.8 (0.62)&13.7 (0.41) & 13.7 (0.41) & 13.7 (0.41)&14.2 (0.35) & 14.2 (0.35) & 14.2 (0.35)&14.3 (0.29) & 14.3 (0.29) & 14.3 (0.29)\\ \hline M10c\_Cubic & 24 & 25 & 17.9 (0.65) & 17.9 (0.65) & 17.9 (0.65)&18.2 (0.58) & 18.2 (0.58) & 18.2 (0.58)&18.6 (0.49) & 18.6 (0.49) & 18.6 (0.49)&18.9 (0.2) & 18.9 (0.2) & 18.9 (0.2)\\ \hline M10d\_Cubic & 70 & 72 & 38.4 (1.9) & 38.4 (1.9) & 38.4 (1.9)&39.3 (0.96) & 39.3 (0.96) & 39.3 (0.96)&40.1 (0.84) & 40.1 (0.84) & 40.1 (0.84)&41.8 (0.61) & 41.8 (0.61) & 41.8 (0.61)\\ \hline M11\_Moebius & 2 & 3 & 2.0 (0.07) & 2.0 (0.07) & 2.0 (0.07)&2.0 (0.02) & 2.0 (0.08) & 2.0 (0.08)&2.0 (0.01) & 2.0 (0.04) & 2.0 (0.04)&2.0 (0.01) & 2.0 (0.03) & 2.0 (0.03)\\ \hline M12\_Norm & 20 & 20 & 16.2 (0.6) & 16.2 (0.6) & 16.2 (0.6)&16.8 (0.52) & 16.8 (0.52) & 16.8 (0.52)&17.2 (0.31) & 17.2 (0.31) & 17.2 (0.31)&17.5 (0.27) & 17.5 (0.27) & 17.5 (0.27)\\ \hline M13a\_Scurve & 2 & 3 & 2.0 (0.06) & 2.0 (0.07) & 2.0 (0.07)&2.0 (0.08) & 2.0 (0.08) & 2.0 (0.08)&2.0 (0.05) & 2.0 (0.05) & 2.0 (0.05)&2.0 (0.04) & 2.0 (0.04) & 2.0 (0.04)\\ \hline M13b\_Spiral & 1 & 13 & 1.1 (0.05) & 1.1 (0.05) & 1.1 (0.05)&1.0 (0.02) & 1.0 (0.02) & 1.0 (0.02)&1.0 (0.01) & 1.0 (0.02) & 1.0 (0.02)&1.0 (0.0) & 1.0 (0.02) & 1.0 (0.02)\\ \hline Mbeta & 10 & 40 & 6.2 (0.27) & 6.2 (0.27) & 6.2 (0.27)&6.3 (0.21) & 6.3 (0.21) & 6.3 (0.21)&6.5 (0.17) & 6.5 (0.17) & 6.5 (0.17)&6.6 (0.11) & 6.6 (0.11) & 6.6 (0.11)\\ \hline Mn1\_Nonlinear & 18 & 72 & 14.0 (0.61) & 14.0 (0.61) & 14.0 (0.61)&14.3 (0.42) & 14.3 (0.42) & 14.3 (0.42)&14.6 (0.35) & 14.6 (0.35) & 14.6 (0.35)&14.7 (0.2) & 14.7 (0.2) & 14.7 (0.2)\\ \hline Mn2\_Nonlinear & 24 & 96 & 17.7 (0.82) & 17.7 (0.82) & 17.7 (0.82)&18.2 (0.69) & 18.2 (0.69) & 18.2 (0.69)&18.3 (0.54) & 18.3 (0.54) & 18.3 (0.54)&18.7 (0.35) & 18.7 (0.35) & 18.7 (0.35)\\ \hline Mp1\_Paraboloid & 3 & 12 & 2.9 (0.12) & 2.9 (0.12) & 2.9 (0.12)&3.0 (0.07) & 3.0 (0.07) & 3.0 (0.07)&2.9 (0.08) & 2.9 (0.08) & 2.9 (0.08)&3.0 (0.04) & 3.0 (0.04) & 3.0 (0.04)\\ \hline Mp2\_Paraboloid & 6 & 21 & 5.1 (0.18) & 5.1 (0.18) & 5.1 (0.18)&5.3 (0.19) & 5.3 (0.19) & 5.3 (0.19)&5.4 (0.17) & 5.4 (0.17) & 5.4 (0.17)&5.5 (0.08) & 5.5 (0.08) & 5.5 (0.08)\\ \hline Mp3\_Paraboloid & 9 & 30 & 6.6 (0.33) & 6.6 (0.33) & 6.6 (0.33)&7.0 (0.21) & 7.0 (0.21) & 7.0 (0.21)&7.2 (0.17) & 7.2 (0.17) & 7.2 (0.17)&7.4 (0.13) & 7.4 (0.13) & 7.4 (0.13)\\ \hline
    \end{tabular}
    \caption{\textbf{Number of samples}. The estimator is accurate on low dimensional datasets. More samples does not dramatically improve performance. \textbf{High dimensional datasets}. Struggles with high dimensional datasets, such as M9 Affine, and M10b,c,d Cubics. \textbf{Variance}. The variance is at a moderately high level compared to other estimators. \textbf{Hyperparameter dependency}. In most cases (72/96), the simplest set of arguments $(n_1=1, n_2=2)$ resulted in the best estimation. When $(n_1=1, n_2=2)$, the method is equivalent to \mle with input from  two nearest neighbour distances. \textbf{Non-linear datasets}.  This method tends to underestimate on Mbeta, Mn1,2 Nonlinear, and Mp2,3 paraboloids.\label{tab:gride_bm}}
\end{sidewaystable}

%% file: tables/CorrInt_benchmark.tex
\begin{sidewaystable}[h]
\centering
\tiny
\begin{tabular}{|l|ll||lll|lll|lll|lll|}
\hline              &    &    & CorrInt &              &              &              &              &              &              &              &              &              &              &              \\  \hline               &    &    &              &              &              &              &              &              &              &              &              &              &              &              \\  \hline             &    &    & 625          &              &              & 1250         &              &              & 2500         &              &              & 5000         &              &              \\  \hline              &    &    &              &              &              &              &              &              &              &              &              &              &              &              \\  \hline              & d  & n  & Best         & med abs      & med rel      & Best         & med abs      & med rel      & Best         & med abs      & med rel      & Best         & med abs      & med rel      \\  \hline               &    &    &              &              &              &              &              &              &              &              &              &              &              &              \\  \hline                M1Sphere & 10 & 11 & 8.8 (0.17) & 8.8 (0.17) & 8.4 (0.38)&9.0 (0.12) & 9.0 (0.12) & 8.6 (0.16)&9.1 (0.06) & 9.1 (0.06) & 9.1 (0.06)&9.3 (0.06) & 9.3 (0.06) & 9.3 (0.06)\\ \hline 
M2 Affine 3to5 & 3 & 5 & 2.8 (0.04) & 2.8 (0.04) & 2.8 (0.12)&2.9 (0.04) & 2.9 (0.04) & 2.8 (0.04)&2.9 (0.02) & 2.9 (0.02) & 2.9 (0.02)&2.9 (0.03) & 2.9 (0.03) & 2.9 (0.03)\\ \hline 
M3 Nonlinear 4to6 & 4 & 6 & 3.5 (0.07) & 3.5 (0.07) & 3.4 (0.13)&3.6 (0.08) & 3.6 (0.08) & 3.5 (0.08)&3.6 (0.03) & 3.6 (0.03) & 3.6 (0.03)&3.7 (0.03) & 3.7 (0.03) & 3.7 (0.03)\\ \hline 
M4 Nonlinear & 4 & 8 & 3.9 (0.11) & 3.8 (0.1) & 3.9 (0.13)&3.8 (0.06) & 3.8 (0.04) & 3.8 (0.06)&3.8 (0.04) & 3.8 (0.04) & 3.8 (0.04)&3.8 (0.02) & 3.8 (0.02) & 3.8 (0.02)\\ \hline 
M5a Helix1d & 1 & 3 & 1.0 (0.02) & 1.0 (0.02) & 1.0 (0.04)&1.0 (0.01) & 1.0 (0.01) & 1.0 (0.02)&1.0 (0.01) & 1.0 (0.01) & 1.0 (0.01)&1.0 (0.01) & 1.0 (0.01) & 1.0 (0.01)\\ \hline 
M5b Helix2d & 2 & 3 & 2.3 (0.1) & 2.7 (0.09) & 2.4 (0.15)&2.3 (0.04) & 2.7 (0.06) & 2.3 (0.05)&2.4 (0.02) & 2.4 (0.02) & 2.4 (0.02)&2.1 (0.02) & 2.1 (0.02) & 2.1 (0.02)\\ \hline
M6 Nonlinear & 6 & 36 & 6.0 (0.16) & 6.0 (0.16) & 6.1 (0.26)&6.0 (0.14) & 5.8 (0.11) & 6.0 (0.14)&6.0 (0.07) & 5.8 (0.06) & 5.8 (0.06)&5.8 (0.04) & 5.7 (0.03) & 5.7 (0.03)\\ \hline 
M7 Roll & 2 & 3 & 2.0 (0.04) & 2.0 (0.04) & 1.9 (0.09)&1.9 (0.03) & 1.9 (0.03) & 1.9 (0.03)&2.0 (0.05) & 2.0 (0.02) & 2.0 (0.02)&2.0 (0.03) & 2.0 (0.02) & 2.0 (0.02)\\ \hline 
M8 Nonlinear & 12 & 72 & 11.4 (0.28) & 11.4 (0.28) & 11.0 (0.49)&11.7 (0.2) & 11.7 (0.2) & 11.2 (0.2)&11.8 (0.14) & 11.8 (0.14) & 11.8 (0.14)&12.0 (0.06) & 12.0 (0.06) & 12.0 (0.06)\\ \hline 
M9 Affine & 20 & 20 & 13.5 (0.2) & 13.5 (0.2) & 12.7 (0.57)&13.9 (0.19) & 13.9 (0.19) & 13.1 (0.22)&14.4 (0.15) & 14.4 (0.15) & 14.4 (0.15)&14.8 (0.1) & 14.8 (0.1) & 14.8 (0.1)\\ \hline 
M10a Cubic & 10 & 11 & 8.5 (0.15) & 8.5 (0.15) & 8.1 (0.35)&8.7 (0.15) & 8.7 (0.15) & 8.3 (0.13)&8.9 (0.06) & 8.9 (0.06) & 8.9 (0.06)&9.1 (0.06) & 9.1 (0.06) & 9.1 (0.06)\\ \hline 
M10b Cubic & 17 & 18 & 12.6 (0.18) & 12.6 (0.18) & 11.9 (0.52)&12.9 (0.15) & 12.9 (0.15) & 12.2 (0.18)&13.4 (0.14) & 13.4 (0.14) & 13.4 (0.14)&13.7 (0.09) & 13.7 (0.09) & 13.7 (0.09)\\ \hline 
M10c Cubic & 24 & 25 & 15.9 (0.28) & 15.9 (0.28) & 15.0 (0.72)&16.6 (0.24) & 16.6 (0.24) & 15.4 (0.25)&17.1 (0.18) & 17.1 (0.18) & 17.1 (0.18)&17.7 (0.1) & 17.7 (0.1) & 17.7 (0.1)\\ \hline
M10d Cubic & 70 & 72 & 31.2 (0.66) & 31.2 (0.66) & 29.0 (0.94)&33.0 (0.42) & 33.0 (0.42) & 30.0 (0.42)&34.6 (0.32) & 34.6 (0.32) & 34.6 (0.32)&36.0 (0.24) & 36.0 (0.24) & 36.0 (0.24)\\ \hline 
M11 Moebius & 2 & 3 & 2.0 (0.02) & 2.0 (0.04) & 2.1 (0.07)&2.0 (0.03) & 2.0 (0.03) & 2.0 (0.03)&2.0 (0.04) & 2.0 (0.02) & 2.0 (0.02)&2.0 (0.02) & 2.0 (0.02) & 2.0 (0.02)\\ \hline 
M12 Norm & 20 & 20 & 12.4 (0.25) & 12.4 (0.25) & 11.7 (0.35)&13.1 (0.15) & 13.1 (0.15) & 12.0 (0.2)&13.6 (0.17) & 13.6 (0.17) & 13.6 (0.17)&14.2 (0.11) & 14.2 (0.11) & 14.2 (0.11)\\ \hline 
M13a Scurve & 2 & 3 & 2.0 (0.04) & 2.0 (0.04) & 1.9 (0.09)&2.0 (0.03) & 2.0 (0.03) & 1.9 (0.03)&2.0 (0.02) & 2.0 (0.02) & 2.0 (0.02)&2.0 (0.01) & 2.0 (0.01) & 2.0 (0.01)\\ \hline 
M13b Spiral & 1 & 13 & 1.3 (0.04) & 1.8 (0.07) & 1.3 (0.04)&1.5 (0.03) & 1.5 (0.03) & 1.6 (0.03)&1.1 (0.01) & 1.1 (0.01) & 1.1 (0.01)&1.0 (0.01) & 1.0 (0.01) & 1.0 (0.01)\\ \hline 
Mbeta & 10 & 40 & 3.4 (0.18) & 3.4 (0.18) & 3.1 (0.19)&3.5 (0.12) & 3.5 (0.12) & 3.2 (0.13)&3.7 (0.11) & 3.7 (0.11) & 3.7 (0.11)&3.9 (0.08) & 3.9 (0.08) & 3.9 (0.08)\\ \hline 
Mn1 Nonlinear & 18 & 72 & 12.3 (0.28) & 12.3 (0.28) & 11.6 (0.5)&12.8 (0.15) & 12.8 (0.15) & 11.9 (0.19)&13.2 (0.09) & 13.2 (0.09) & 13.2 (0.09)&13.6 (0.09) & 13.6 (0.09) & 13.6 (0.09)\\ \hline 
Mn2 Nonlinear & 24 & 96 & 15.0 (0.23) & 15.0 (0.23) & 14.4 (0.45)&15.7 (0.19) & 15.7 (0.19) & 14.6 (0.19)&16.3 (0.14) & 16.3 (0.14) & 16.3 (0.14)&16.9 (0.14) & 16.9 (0.14) & 16.9 (0.14)\\ \hline 
Mp1 Paraboloid & 3 & 12 & 2.1 (0.11) & 2.1 (0.07) & 2.1 (0.11)&2.2 (0.1) & 2.2 (0.08) & 2.1 (0.06)&2.2 (0.06) & 2.2 (0.06) & 2.2 (0.06)&2.2 (0.05) & 2.2 (0.07) & 2.2 (0.07)\\ \hline 
Mp2 Paraboloid & 6 & 21 & 2.8 (0.22) & 2.7 (0.21) & 2.8 (0.19)&2.7 (0.16) & 2.7 (0.16) & 2.7 (0.16)&2.8 (0.12) & 2.7 (0.09) & 2.7 (0.09)&2.7 (0.08) & 2.6 (0.08) & 2.6 (0.08)\\ \hline 
Mp3 Paraboloid & 9 & 30 & 3.1 (0.3) & 3.1 (0.29) & 3.1 (0.3)&3.1 (0.19) & 3.0 (0.22) & 3.0 (0.17)&3.1 (0.14) & 3.0 (0.13) & 3.0 (0.13)&3.0 (0.12) & 2.9 (0.13) & 2.9 (0.13)\\ \hline     
\end{tabular} 
\caption{\footnotesize \textbf{Number of samples}. In general, estimates do not see huge improvement as the sample size is increased. Some inaccuracies on low dimensional datasets even with high sample densities, such as Mp1 Paraboloid.  \textbf{High dimensional datasets}. CorrInt struggles with high M9 Affine, and M10b,c,d cubic datasets with a strong bias to underestimate. 
\textbf{Variance}. The variance is generally low and decreases as sample size increases across all datasets. \textbf{Hyperparameter dependency}. In general is insensitive to the choice of parameters \textbf{Nonlinear datasets}. Can be challenged by datasets with non-uniform sampling density, such as Mn1, Mn2 Nonlinear, and Mbeta; as well as those with some curvature, such as Mp1,2,3 Paraboloid.   }
\label{tab:corrint_bm}
\end{sidewaystable}

%% file: tables/WODCap_benchmark.tex
\begin{sidewaystable}[h]
\centering
\tiny
\begin{tabular}{|l|ll||lll|lll|lll|lll|}
\hline
               &    &    & WODCap &              &              &              &              &              &              &              &              &              &              &              \\  \hline
               &    &    &              &              &              &              &              &              &              &              &              &              &              &              \\  \hline
n              &    &    & 625          &              &              & 1250         &              &              & 2500         &              &              & 5000         &              &              \\  \hline
               &    &    &              &              &              &              &              &              &              &              &              &              &              &              \\  \hline
               & d  & n  & Best         & med abs      & med rel      & Best         & med abs      & med rel      & Best         & med abs      & med rel      & Best         & med abs      & med rel      \\  \hline
               &    &    &              &              &              &              &              &              &              &              &              &              &              &              \\  \hline 
               M1 Sphere & 10 & 11 & 7.74 (0.08) & 7.74 (0.08) & 6.93 (0.0) & 8.36 (0.07) & 6.93 (0.0) & 7.63 (0.04) & 9.15 (0.0) & 9.15 (0.0) & 8.06 (0.03) & 9.46 (0.03) & 9.15 (0.0) & 8.4 (0.02)  \\ \hline
M2 Affine 3to5 & 3 & 5 & 3.04 (0.06) & 3.72 (0.08) & 3.42 (0.0) & 3.15 (0.04) & 3.42 (0.0) & 3.45 (0.05) & 3.19 (0.04) & 3.42 (0.0) & 3.55 (0.03) & 3.24 (0.02) & 3.42 (0.0) & 3.61 (0.02)  \\ \hline
M3 Nonlinear 4to6 & 4 & 6 & 4.0 (0.23) & 4.84 (0.12) & 4.3 (0.0) & 4.04 (0.07) & 4.3 (0.0) & 4.31 (0.06) & 3.95 (0.04) & 4.3 (0.0) & 4.48 (0.04) & 4.03 (0.04) & 4.3 (0.0) & 4.61 (0.02)  \\ \hline
M4 Nonlinear & 4 & 8 & 4.3 (0.11) & 6.61 (0.19) & 5.43 (0.0) & 4.22 (0.07) & 5.43 (0.0) & 5.01 (0.07) & 4.17 (0.06) & 5.43 (0.0) & 4.93 (0.05) & 4.19 (0.03) & 5.43 (0.0) & 4.89 (0.02)  \\ \hline
M5a Helix1d & 1 & 3 & 1.22 (0.07) & 2.86 (0.34) & 1.3 (0.18) & 1.18 (0.05) & 1.19 (0.0) & 1.25 (0.04) & 1.18 (0.03) & 1.19 (0.0) & 1.25 (0.02) & 1.16 (0.04) & 1.19 (0.0) & 1.26 (0.02)  \\ \hline
M5b Helix2d & 2 & 3 & 3.22 (0.08) & 4.16 (0.1) & 3.46 (0.19) & 3.15 (0.07) & 3.55 (0.31) & 3.7 (0.06) & 3.14 (0.04) & 3.42 (0.0) & 3.63 (0.04) & 2.95 (0.03) & 3.46 (0.19) & 3.7 (0.04)  \\ \hline
M6 Nonlinear & 6 & 36 & 5.89 (0.15) & 8.36 (0.09) & 6.93 (0.0) & 5.88 (0.08) & 7.05 (0.48) & 7.23 (0.1) & 5.82 (0.05) & 6.93 (0.0) & 7.31 (0.06) & 5.77 (0.04) & 6.93 (0.0) & 7.22 (0.04)  \\ \hline
M7 Roll & 2 & 3 & 2.21 (0.04) & 3.21 (0.17) & 2.56 (0.26) & 2.28 (0.06) & 2.6 (0.22) & 2.46 (0.05) & 2.31 (0.04) & 2.7 (0.0) & 2.5 (0.03) & 2.32 (0.0) & 2.7 (0.0) & 2.54 (0.02)  \\ \hline
M8 Nonlinear & 12 & 72 & 9.48 (0.14) & 9.48 (0.14) & 9.15 (0.0) & 10.59 (0.09) & 9.15 (0.0) & 9.42 (0.11) & 11.53 (0.07) & 12.72 (1.19) & 10.27 (0.06) & 11.75 (0.02) & 13.12 (0.0) & 10.84 (0.03)  \\ \hline
M9 Affine & 20 & 20 & 10.09 (0.13) & 10.09 (0.13) & 9.15 (0.0) & 11.02 (0.07) & 9.15 (0.0) & 10.12 (0.07) & 13.12 (0.0) & 13.12 (0.0) & 10.84 (0.05) & 13.31 (0.03) & 13.12 (0.0) & 11.37 (0.03)  \\ \hline
M10a Cubic & 10 & 11 & 8.38 (0.12) & 8.38 (0.12) & 6.93 (0.0) & 9.15 (0.0) & 9.15 (0.0) & 8.07 (0.05) & 9.41 (0.05) & 9.15 (0.0) & 8.47 (0.04) & 10.0 (0.03) & 9.15 (0.0) & 8.71 (0.03)  \\ \hline
M10b Cubic & 17 & 18 & 9.81 (0.09) & 9.81 (0.09) & 9.15 (0.0) & 10.73 (0.07) & 9.15 (0.0) & 9.81 (0.07) & 11.72 (0.05) & 10.14 (1.72) & 10.45 (0.05) & 13.12 (0.0) & 13.12 (0.0) & 10.89 (0.05)  \\ \hline
M10c Cubic & 24 & 25 & 10.54 (0.07) & 10.54 (0.07) & 9.15 (0.0) & 12.92 (0.86) & 12.92 (0.86) & 10.65 (0.07) & 13.12 (0.0) & 13.12 (0.0) & 11.34 (0.04) & 14.3 (0.03) & 13.12 (0.0) & 11.8 (0.03)  \\ \hline
M10d Cubic & 70 & 72 & 13.12 (0.0) & 11.76 (0.08) & 13.12 (0.0) & 13.42 (0.13) & 13.12 (0.0) & 12.03 (0.05) & 16.68 (1.24) & 13.12 (0.0) & 12.52 (0.02) & 17.09 (0.0) & 13.12 (0.0) & 12.75 (0.02)  \\ \hline
M11 Moebius & 2 & 3 & 2.07 (0.07) & 3.45 (0.12) & 2.7 (0.0) & 2.32 (0.0) & 2.7 (0.0) & 2.64 (0.06) & 2.32 (0.0) & 2.7 (0.0) & 2.62 (0.05) & 2.32 (0.0) & 2.7 (0.0) & 2.6 (0.03)  \\ \hline
M12 Norm & 20 & 20 & 9.53 (0.15) & 9.53 (0.15) & 9.15 (0.0) & 10.79 (0.07) & 9.15 (0.0) & 9.82 (0.09) & 13.12 (0.0) & 13.12 (0.0) & 10.82 (0.07) & 13.12 (0.0) & 13.12 (0.0) & 11.56 (0.03)  \\ \hline
M13a Scurve & 2 & 3 & 2.09 (0.08) & 2.63 (0.06) & 2.29 (0.27) & 2.24 (0.06) & 2.65 (0.18) & 2.46 (0.04) & 2.31 (0.03) & 2.7 (0.0) & 2.5 (0.02) & 2.32 (0.0) & 2.7 (0.0) & 2.53 (0.02)  \\ \hline
M13b Spiral & 1 & 13 & 1.3 (0.06) & 2.35 (0.05) & 2.11 (0.0) & 1.8 (0.04) & 2.11 (0.0) & 2.27 (0.04) & 1.47 (0.28) & 2.7 (0.0) & 2.82 (0.02) & 1.19 (0.03) & 1.61 (0.0) & 2.42 (0.05)  \\ \hline
Mbeta & 10 & 40 & 7.63 (0.19) & 7.42 (0.21) & 6.63 (0.6) & 8.15 (0.12) & 6.93 (0.0) & 5.79 (0.12) & 8.47 (0.09) & 6.93 (0.0) & 6.08 (0.07) & 8.74 (0.06) & 6.93 (0.0) & 6.45 (0.05)  \\ \hline
Mn1 Nonlinear & 18 & 72 & 9.63 (0.15) & 9.63 (0.15) & 9.15 (0.0) & 10.68 (0.05) & 9.15 (0.0) & 9.71 (0.05) & 11.81 (1.81) & 11.81 (1.81) & 10.49 (0.05) & 13.12 (0.0) & 13.12 (0.0) & 11.07 (0.03)  \\ \hline
Mn2 Nonlinear & 24 & 96 & 10.15 (0.14) & 10.15 (0.14) & 9.15 (0.0) & 11.26 (0.05) & 9.75 (1.42) & 10.38 (0.07) & 13.12 (0.0) & 13.12 (0.0) & 11.24 (0.04) & 13.76 (0.05) & 13.12 (0.0) & 11.82 (0.02)  \\ \hline
Mp1 Paraboloid & 3 & 12 & 3.0 (0.08) & 3.62 (0.08) & 3.42 (0.0) & 2.93 (0.05) & 3.42 (0.0) & 3.39 (0.04) & 3.21 (0.04) & 3.42 (0.0) & 3.54 (0.03) & 3.29 (0.02) & 3.42 (0.0) & 3.63 (0.02)  \\ \hline
Mp2 Paraboloid & 6 & 21 & 4.75 (0.09) & 4.28 (0.13) & 3.77 (0.43) & 5.27 (0.07) & 4.3 (0.0) & 4.23 (0.05) & 5.72 (0.05) & 5.43 (0.0) & 4.87 (0.05) & 5.87 (0.04) & 5.43 (0.0) & 5.35 (0.03)  \\ \hline
Mp3 Paraboloid & 9 & 30 & 4.7 (0.1) & 3.92 (0.09) & 3.42 (0.0) & 5.47 (0.07) & 4.3 (0.0) & 4.02 (0.07) & 6.06 (0.07) & 5.43 (0.0) & 4.95 (0.06) & 6.92 (0.04) & 6.86 (0.33) & 5.8 (0.02)  \\
 \hline
    \end{tabular}
\label{tab:wodcap_bm}
\caption{\footnotesize \textbf{Number of samples}. 
Vulnerable even on M1Sphere to insufficently many samples, though as sample size increases, the performance improves. Biases can persist even as the number of samples increases.   \textbf{High dimensional datasets}. Significant bias to underestimate datasets with dimension beyond 10, even with many samples. 
\textbf{Variance}. Generally very consistent with low variance, even with few samples. \textbf{Hyperparameter dependency}. Can be quite sensitive to the neighbourhood size parameter, even with many points and on relatively simple and low dimensional datasets such as M7 Roll and M13b Spiral. \textbf{Nonlinear datasets}.  On some datasets, there is a bias to underestimate, as demondstrated on Mn1,2 Nonlinear, and Mp2,3 Paraboloids.  }
\end{sidewaystable}

%% file: tables/TwoNN_benchmark.tex
\begin{sidewaystable}[h]
\centering
\tiny
\begin{tabular}{|l|ll||lll|lll|lll|lll|}
\hline
    &    &    & TwoNN &              &              &              &              &              &              &              &              &              &              &              \\  \hline               &    &    &              &              &              &              &              &              &              &              &              &              &              &              \\  \hline              &    &    & 625          &              &              & 1250         &              &              & 2500         &              &              & 5000         &              &              \\  \hline              &    &    &              &              &              &              &              &              &              &              &              &              &              &              \\  \hline               & d  & n  & Best         & med abs      & med rel      & Best         & med abs      & med rel      & Best         & med abs      & med rel      & Best         & med abs      & med rel      \\  \hline              &    &    &              &              &              &              &              &              &              &              &              &              &              &              \\  \hline               M1Sphere & 10 & 11 & 9.2 (0.54) & 9.2 (0.54) & 9.2 (0.54)&9.4 (0.26) & 9.4 (0.26) & 9.4 (0.26)&9.4 (0.21) & 9.4 (0.21) & 9.4 (0.21)&9.5 (0.12) & 9.5 (0.12) & 9.5 (0.12)\\ \hline 

M2 Affine 3to5 & 3 & 5 & 2.9 (0.13) & 2.9 (0.13) & 2.9 (0.13)&2.9 (0.12) & 2.9 (0.12) & 2.9 (0.12)&2.9 (0.09) & 2.9 (0.09) & 2.9 (0.09)&2.9 (0.05) & 2.9 (0.05) & 2.9 (0.05)\\ \hline 
M3 Nonlinear4to6 & 4 & 6 & 3.8 (0.28) & 3.8 (0.18) & 3.8 (0.18)&3.8 (0.2) & 3.8 (0.16) & 3.8 (0.16)&3.9 (0.13) & 3.9 (0.11) & 3.9 (0.11)&3.9 (0.1) & 3.9 (0.08) & 3.9 (0.08)\\ \hline 
M4 Nonlinear & 4 & 8 & 4.0 (0.22) & 3.9 (0.24) & 3.9 (0.24)&4.0 (0.15) & 3.9 (0.14) & 3.9 (0.14)&3.9 (0.13) & 3.9 (0.1) & 3.9 (0.1)&3.9 (0.08) & 3.9 (0.08) & 3.9 (0.08)\\ \hline
M5a Helix1d & 1 & 3 & 1.0 (0.06) & 1.0 (0.05) & 1.0 (0.05)&1.0 (0.03) & 1.0 (0.04) & 1.0 (0.04)&1.0 (0.03) & 1.0 (0.03) & 1.0 (0.03)&1.0 (0.02) & 1.0 (0.02) & 1.0 (0.02)\\ \hline 
M5b Helix2d & 2 & 3 & 2.2 (0.14) & 2.2 (0.14) & 2.2 (0.14)&2.1 (0.08) & 2.1 (0.08) & 2.1 (0.08)&2.0 (0.06) & 2.0 (0.07) & 2.0 (0.07)&2.0 (0.05) & 2.0 (0.04) & 2.0 (0.04)\\ \hline 
M6 Nonlinear & 6 & 36 & 6.2 (0.31) & 6.2 (0.31) & 6.2 (0.31)&6.0 (0.24) & 6.0 (0.25) & 6.0 (0.25)&6.0 (0.16) & 5.9 (0.15) & 5.9 (0.15)&5.9 (0.15) & 5.9 (0.1) & 5.9 (0.1)\\ \hline 
M7 Roll & 2 & 3 & 2.0 (0.1) & 2.0 (0.1) & 2.0 (0.1)&2.0 (0.1) & 2.0 (0.1) & 2.0 (0.1)&2.0 (0.06) & 2.0 (0.06) & 2.0 (0.06)&2.0 (0.03) & 2.0 (0.04) & 2.0 (0.04)\\ \hline 
M8 Nonlinear & 12 & 72 & 13.4 (0.83) & 13.4 (0.83) & 13.4 (0.83)&13.5 (0.59) & 13.5 (0.59) & 13.5 (0.59)&13.2 (0.35) & 13.2 (0.35) & 13.2 (0.35)&13.2 (0.23) & 13.2 (0.23) & 13.2 (0.23)\\ \hline
M9 Affine & 20 & 20 & 15.2 (1.03) & 15.1 (0.7) & 15.1 (0.7)&15.3 (0.54) & 15.3 (0.54) & 15.3 (0.54)&15.6 (0.41) & 15.6 (0.41) & 15.6 (0.41)&15.9 (0.27) & 15.9 (0.27) & 15.9 (0.27)\\ \hline 
M10a Cubic & 10 & 11 & 9.0 (0.42) & 9.0 (0.38) & 9.0 (0.38)&9.1 (0.35) & 9.1 (0.34) & 9.1 (0.34)&9.2 (0.16) & 9.2 (0.15) & 9.2 (0.15)&9.2 (0.14) & 9.2 (0.14) & 9.2 (0.14)\\ \hline
M10b Cubic & 17 & 18 & 13.6 (0.6) & 13.6 (0.6) & 13.6 (0.6)&13.6 (0.44) & 13.6 (0.44) & 13.6 (0.44)&14.2 (0.36) & 14.2 (0.37) & 14.2 (0.37)&14.3 (0.29) & 14.3 (0.29) & 14.3 (0.29)\\ \hline 
M10c Cubic & 24 & 25 & 17.8 (0.72) & 17.8 (0.72) & 17.8 (0.72)&18.1 (0.62) & 18.1 (0.62) & 18.1 (0.62)&18.5 (0.5) & 18.5 (0.5) & 18.5 (0.5)&18.9 (0.21) & 18.9 (0.21) & 18.9 (0.21)\\ \hline
M10d Cubic & 70 & 72 & 38.1 (1.88) & 38.1 (1.88) & 38.1 (1.88)&39.2 (1.07) & 39.2 (1.07) & 39.2 (1.07)&39.9 (0.81) & 39.9 (0.82) & 39.9 (0.82)&41.7 (0.6) & 41.7 (0.6) & 41.7 (0.6)\\ \hline 
M11 Moebius & 2 & 3 & 2.0 (0.08) & 2.0 (0.08) & 2.0 (0.08)&2.0 (0.08) & 2.0 (0.09) & 2.0 (0.09)&2.0 (0.05) & 2.0 (0.05) & 2.0 (0.05)&2.0 (0.04) & 2.0 (0.04) & 2.0 (0.04)\\ \hline 
M12 Norm & 20 & 20 & 16.1 (0.65) & 16.1 (0.65) & 16.1 (0.65)&16.7 (0.56) & 16.7 (0.56) & 16.7 (0.56)&17.2 (0.29) & 17.2 (0.29) & 17.2 (0.29)&17.4 (0.29) & 17.4 (0.29) & 17.4 (0.29)\\ \hline 
M13a Scurve & 2 & 3 & 2.0 (0.09) & 2.0 (0.09) & 2.0 (0.09)&2.0 (0.09) & 2.0 (0.1) & 2.0 (0.1)&2.0 (0.06) & 2.0 (0.06) & 2.0 (0.06)&2.0 (0.03) & 2.0 (0.04) & 2.0 (0.04)\\ \hline 
M13b Spiral & 1 & 13 & 1.0 (0.06) & 1.0 (0.06) & 1.0 (0.06)&1.0 (0.03) & 1.0 (0.04) & 1.0 (0.04)&1.0 (0.03) & 1.0 (0.03) & 1.0 (0.03)&1.0 (0.02) & 1.0 (0.02) & 1.0 (0.02)\\ \hline 
Mbeta & 10 & 40 & 6.3 (0.44) & 6.2 (0.24) & 6.2 (0.24)&6.4 (0.27) & 6.2 (0.23) & 6.2 (0.23)&6.7 (0.24) & 6.5 (0.18) & 6.5 (0.18)&6.7 (0.16) & 6.6 (0.12) & 6.6 (0.12)\\ \hline 
Mn1 Nonlinear & 18 & 72 & 13.9 (0.67) & 13.9 (0.67) & 13.9 (0.67)&14.2 (0.44) & 14.2 (0.44) & 14.2 (0.44)&14.5 (0.34) & 14.5 (0.34) & 14.5 (0.34)&14.6 (0.22) & 14.6 (0.22) & 14.6 (0.22)\\ \hline 
Mn2 Nonlinear & 24 & 96 & 17.6 (0.79) & 17.6 (0.79) & 17.6 (0.79)&18.2 (0.68) & 18.2 (0.68) & 18.2 (0.68)&18.3 (0.54) & 18.3 (0.54) & 18.3 (0.54)&18.7 (0.39) & 18.7 (0.39) & 18.7 (0.39)\\ \hline 
Mp1 Paraboloid & 3 & 12 & 2.9 (0.21) & 2.8 (0.13) & 2.8 (0.13)&3.0 (0.09) & 3.0 (0.09) & 3.0 (0.09)&3.0 (0.11) & 2.9 (0.09) & 2.9 (0.09)&3.0 (0.08) & 3.0 (0.05) & 3.0 (0.05)\\ \hline 
Mp2 Paraboloid & 6 & 21 & 5.1 (0.4) & 5.1 (0.17) & 5.1 (0.17)&5.3 (0.22) & 5.3 (0.22) & 5.3 (0.22)&5.4 (0.18) & 5.4 (0.18) & 5.4 (0.18)&5.5 (0.08) & 5.5 (0.08) & 5.5 (0.08)\\ \hline 
Mp3 Paraboloid & 9 & 30 & 6.6 (0.36) & 6.6 (0.36) & 6.6 (0.36)&7.0 (0.21) & 7.0 (0.21) & 7.0 (0.21)&7.2 (0.17) & 7.2 (0.17) & 7.2 (0.17)&7.4 (0.13) & 7.4 (0.13) & 7.4 (0.13)\\ \hline   
\end{tabular}
\caption{\footnotesize \textbf{Number of samples}. Doesn't need many samples on low dimensional datasets for good accuracy. 
\textbf{High dimensional datasets}. Has a bias to underestimate, on moderately high dimensional datasets, such as M9 Affine; underestimation bias worsens with high dimensions, such as on M10b,c,d Cubic datasets. 
\textbf{Variance}. Moderate levels of variance that decreases as the number of samples increases. \textbf{Hyperparameter dependency}. A single hyperparameter can achieve consistent performance across all datasets that is close to the best possible performance. There is only a single hyperparameter; the discard fraction is varied between 0.1, 0.25, 0.5, 0.75. The discard fraction for both med abs and med rel columns are 0.1 for all $n$ except for 2500, which is 0.75. \textbf{Nonlinear datasets}. Struggles with non-uniform sampling, such as the Mbeta, and M20 Norm datasets, and also Mn1,2 Nonlinear. Challenged also by the higher dimensional paraboloids Mp2,3.  }
\label{tab:twonn_bm}
\end{sidewaystable}

%% file: tables/TLE_benchmark.tex
\begin{sidewaystable}[h]
\centering
\tiny
\begin{tabular}{|l|ll||lll|lll|lll|lll|}
\hline              &    &    & TLE &              &              &              &              &              &              &              &              &              &              &              \\  \hline               &    &    &              &              &              &              &              &              &              &              &              &              &              &              \\  \hline             &    &    & 625          &              &              & 1250         &              &              & 2500         &              &              & 5000         &              &              \\  \hline               &    &    &              &              &              &              &              &              &              &              &              &              &              &              \\  \hline               & d  & n  & Best         & med abs      & med rel      & Best         & med abs      & med rel      & Best         & med abs      & med rel      & Best         & med abs      & med rel      \\  \hline               &    &    &              &              &              &              &              &              &              &              &              &              &              &              \\  \hline              M1Sphere & 10 & 11 & 9.9 (0.14) & 9.2 (0.11) & 10.9 (0.2)&10.1 (0.09) & 10.1 (0.09) & 11.2 (0.12)&10.1 (0.05) & 10.3 (0.06) & 9.8 (0.05)&9.9 (0.04) & 10.5 (0.05) & 10.5 (0.05)\\ \hline 
M2 Affine 3to5 & 3 & 5 & 3.0 (0.03) & 3.0 (0.02) & 3.4 (0.04)&3.0 (0.02) & 3.2 (0.04) & 3.4 (0.03)&3.0 (0.01) & 3.2 (0.02) & 3.1 (0.01)&3.0 (0.01) & 3.3 (0.02) & 3.3 (0.02)\\ \hline 
M3 Nonlinear 4to6 & 4 & 6 & 4.1 (0.05) & 3.9 (0.04) & 4.5 (0.09)&4.0 (0.05) & 4.2 (0.06) & 4.6 (0.06)&4.0 (0.02) & 4.3 (0.04) & 4.1 (0.03)&4.0 (0.01) & 4.3 (0.03) & 4.3 (0.03)\\ \hline 
M4 Nonlinear & 4 & 8 & 4.5 (0.07) & 4.5 (0.07) & 5.1 (0.09)&4.4 (0.03) & 4.5 (0.05) & 4.9 (0.05)&4.3 (0.02) & 4.4 (0.03) & 4.3 (0.03)&4.2 (0.01) & 4.4 (0.02) & 4.4 (0.02)\\ \hline 
M5a Helix1d & 1 & 3 & 1.2 (0.01) & 1.2 (0.01) & 1.2 (0.01)&1.1 (0.0) & 1.2 (0.01) & 1.2 (0.01)&1.1 (0.0) & 1.1 (0.01) & 1.1 (0.0)&1.1 (0.0) & 1.1 (0.0) & 1.1 (0.0)\\ \hline 
M5b Helix2d & 2 & 3 & 2.7 (0.03) & 2.9 (0.03) & 3.2 (0.05)&2.8 (0.01) & 2.9 (0.05) & 3.2 (0.04)&2.7 (0.02) & 2.7 (0.02) & 2.9 (0.02)&2.5 (0.01) & 2.5 (0.01) & 2.5 (0.01)\\ \hline 
M6 Nonlinear & 6 & 36 & 6.5 (0.08) & 7.0 (0.11) & 8.1 (0.16)&6.7 (0.05) & 7.1 (0.1) & 7.8 (0.11)&6.7 (0.04) & 6.9 (0.05) & 6.8 (0.06)&6.5 (0.02) & 6.8 (0.04) & 6.8 (0.04)\\ \hline 
M7Roll & 2 & 3 & 2.1 (0.02) & 2.1 (0.02) & 2.3 (0.02)&2.1 (0.01) & 2.2 (0.02) & 2.3 (0.02)&2.0 (0.01) & 2.2 (0.02) & 2.1 (0.01)&2.0 (0.0) & 2.2 (0.01) & 2.2 (0.01)\\ \hline 
M8 Nonlinear & 12 & 72 & 12.0 (0.12) & 12.6 (0.15) & 15.3 (0.19)&12.3 (0.08) & 14.2 (0.16) & 15.7 (0.2)&12.1 (0.05) & 14.3 (0.13) & 13.6 (0.1)&12.6 (0.03) & 14.4 (0.06) & 14.4 (0.06)\\ \hline 
M9 Affine & 20 & 20 & 16.6 (0.25) & 13.6 (0.09) & 16.6 (0.25)&17.1 (0.19) & 15.5 (0.18) & 17.1 (0.19)&17.6 (0.15) & 16.0 (0.14) & 14.9 (0.09)&18.1 (0.12) & 16.4 (0.09) & 16.4 (0.09)\\ \hline 
M10a Cubic & 10 & 11 & 9.7 (0.12) & 8.7 (0.07) & 10.4 (0.13)&9.9 (0.16) & 9.7 (0.13) & 10.6 (0.15)&9.9 (0.06) & 9.9 (0.06) & 9.3 (0.07)&10.0 (0.05) & 10.0 (0.05) & 10.0 (0.05)\\ \hline 
M10b Cubic & 17 & 18 & 15.4 (0.23) & 12.7 (0.11) & 15.4 (0.23)&15.8 (0.15) & 14.3 (0.14) & 15.8 (0.15)&16.4 (0.14) & 14.8 (0.12) & 13.8 (0.1)&16.7 (0.11) & 15.1 (0.1) & 15.1 (0.1)\\ \hline 
M10c Cubic & 24 & 25 & 19.2 (0.28) & 15.8 (0.15) & 19.2 (0.28)&20.2 (0.24) & 18.4 (0.18) & 20.2 (0.24)&20.9 (0.17) & 18.9 (0.14) & 17.5 (0.09)&21.5 (0.1) & 19.4 (0.09) & 19.4 (0.09)\\ \hline 
M10d Cubic & 70 & 72 & 37.1 (0.74) & 29.6 (0.36) & 37.1 (0.74)&39.4 (0.5) & 35.6 (0.4) & 39.4 (0.5)&41.3 (0.33) & 37.4 (0.29) & 33.9 (0.26)&43.2 (0.33) & 38.9 (0.26) & 38.9 (0.26)\\ \hline 
M11 Moebius & 2 & 3 & 2.2 (0.02) & 2.2 (0.02) & 2.4 (0.02)&2.2 (0.01) & 2.2 (0.02) & 2.3 (0.01)&2.1 (0.01) & 2.2 (0.02) & 2.1 (0.01)&2.1 (0.01) & 2.2 (0.01) & 2.2 (0.01)\\ \hline 
M12 Norm & 20 & 20 & 16.4 (0.26) & 13.3 (0.1) & 16.4 (0.26)&17.4 (0.15) & 15.8 (0.14) & 17.4 (0.15)&18.1 (0.16) & 16.5 (0.09) & 15.1 (0.12)&18.9 (0.09) & 17.1 (0.08) & 17.1 (0.08)\\ \hline 
M13a Scurve & 2 & 3 & 2.0 (0.02) & 2.1 (0.02) & 2.3 (0.02)&2.0 (0.01) & 2.2 (0.02) & 2.3 (0.02)&2.0 (0.0) & 2.2 (0.01) & 2.1 (0.01)&2.0 (0.0) & 2.2 (0.01) & 2.2 (0.01)\\ \hline 
M13b Spiral & 1 & 13 & 1.6 (0.01) & 1.9 (0.02) & 2.0 (0.03)&1.8 (0.03) & 1.8 (0.03) & 2.0 (0.02)&1.2 (0.01) & 1.3 (0.01) & 1.9 (0.02)&1.1 (0.0) & 1.1 (0.0) & 1.1 (0.0)\\ \hline 
Mbeta & 10 & 40 & 6.5 (0.09) & 5.1 (0.08) & 6.5 (0.09)&6.8 (0.07) & 6.0 (0.07) & 6.8 (0.07)&7.1 (0.05) & 6.3 (0.05) & 5.9 (0.04)&7.4 (0.05) & 6.6 (0.05) & 6.6 (0.05)\\ \hline 
Mn1 Nonlinear & 18 & 72 & 15.3 (0.21) & 12.5 (0.13) & 15.3 (0.21)&15.8 (0.12) & 14.4 (0.11) & 15.8 (0.12)&16.4 (0.13) & 14.9 (0.11) & 13.8 (0.05)&16.8 (0.11) & 15.2 (0.09) & 15.2 (0.09)\\ \hline 
Mn2 Nonlinear & 24 & 96 & 18.6 (0.21) & 15.2 (0.15) & 18.6 (0.21)&19.5 (0.21) & 17.7 (0.17) & 19.5 (0.21)&20.2 (0.17) & 18.3 (0.14) & 16.9 (0.09)&20.9 (0.14) & 18.9 (0.12) & 18.9 (0.12)\\ \hline 
Mp1 Paraboloid & 3 & 12 & 3.0 (0.02) & 2.9 (0.03) & 3.3 (0.04)&3.0 (0.02) & 3.2 (0.04) & 3.4 (0.04)&3.0 (0.02) & 3.2 (0.02) & 3.1 (0.01)&3.0 (0.01) & 3.3 (0.01) & 3.3 (0.01)\\ \hline 
Mp2 Paraboloid & 6 & 21 & 5.5 (0.09) & 4.2 (0.08) & 5.5 (0.09)&5.9 (0.05) & 5.3 (0.06) & 5.9 (0.05)&6.1 (0.03) & 5.5 (0.04) & 5.2 (0.03)&6.0 (0.03) & 5.7 (0.03) & 5.7 (0.03)\\ \hline 
Mp3 Paraboloid & 9 & 30 & 6.7 (0.14) & 4.5 (0.13) & 6.7 (0.14)&7.4 (0.09) & 6.4 (0.08) & 7.4 (0.09)&7.9 (0.07) & 7.0 (0.08) & 6.6 (0.06)&8.3 (0.06) & 7.4 (0.05) & 7.4 (0.05)\\ \hline    
\end{tabular} 
\caption{\footnotesize \textbf{Number of samples}. On some tricky low-dimensional datasets, such as M11 Moebius and M13b spiral, more points are needed to converge to the correct dimension, but on most low dimensional datasets, not a lot of points are required to give an accurate estimate.  
\textbf{High dimensional datasets}. Like many other estimators, TLE underestimates on high dimensional datasets, such as M1b,c,d Cubic, and M9 Affine. 
\textbf{Variance}. The variance is generally low compared to other estimators. \textbf{Hyperparameter dependency}. For most datasets, there is a mild sensitivity to the choice of hyperparameter (neighbourhood size), but the dependency is more acute on nonlinear datasets such as M8, Mn1,2 Nonlinear, M12 Norm, Mbeta; and high dimensional datasets, such as M9 Affine, and M10a,b,c,d Cubic. \textbf{Nonlinear datasets}. Generally inaccurate and underestimates on the non-linear datasets referred to above. }
\label{tab:tle_bm}
\end{sidewaystable}

%% file: tables/ESS_benchmark.tex
\begin{sidewaystable}[h]
\centering
\tiny
\begin{tabular}{|l|ll||lll|lll|lll|lll|}
\hline

&    &    & ESS &              &              &              &              &              &              &              &              &              &              &              \\  \hline               &    &    &              &              &              &              &              &              &              &              &              &              &              &              \\  \hline              &    &    & 625          &              &              & 1250         &              &              & 2500         &              &              & 5000         &              &              \\  \hline               &    &    &              &              &              &              &              &              &              &              &              &              &              &              \\  \hline              & d  & n  & Best         & med abs      & med rel      & Best         & med abs      & med rel      & Best         & med abs      & med rel      & Best         & med abs      & med rel      \\  \hline              &    &    &              &              &              &              &              &              &              &              &              &              &              &              \\  \hline            
M1Sphere & 10 & 11 & 10.0 (0.08) & 10.2 (0.06) & 10.2 (0.06)&10.0 (0.04) & 10.1 (0.03) & 10.1 (0.03)&10.0 (0.03) & 10.1 (0.02) & 10.2 (0.01)&10.0 (0.02) & 10.1 (0.01) & 10.1 (0.01)\\ \hline 
M2 Affine 3to5 & 3 & 5 & 3.0 (0.02) & 2.8 (0.02) & 2.9 (0.01)&3.0 (0.01) & 2.9 (0.01) & 2.9 (0.01)&3.0 (0.01) & 2.9 (0.01) & 2.9 (0.01)&3.0 (0.01) & 3.0 (0.01) & 2.9 (0.0)\\ \hline 
M3 Nonlinear4to6 & 4 & 6 & 4.0 (0.05) & 3.8 (0.05) & 4.0 (0.05)&4.0 (0.04) & 3.9 (0.04) & 3.9 (0.03)&4.0 (0.02) & 3.9 (0.02) & 3.9 (0.02)&4.0 (0.02) & 4.1 (0.01) & 3.9 (0.01)\\ \hline 
M4 Nonlinear & 4 & 8 & 4.3 (0.06) & 5.2 (0.08) & 5.3 (0.08)&4.1 (0.04) & 4.8 (0.03) & 4.8 (0.03)&4.0 (0.03) & 4.6 (0.02) & 4.9 (0.02)&4.0 (0.02) & 4.5 (0.01) & 4.6 (0.01)\\ \hline 
M5a Helix1d & 1 & 3 & 1.1 (0.0) & 1.9 (0.04) & 2.0 (0.03)&1.0 (0.0) & 1.3 (0.01) & 1.2 (0.0)&1.0 (0.0) & 1.1 (0.0) & 1.2 (0.0)&1.0 (0.0) & 1.1 (0.0) & 1.1 (0.0)\\ \hline 

M5b Helix2d & 2 & 3 & 2.8 (0.02) & 2.8 (0.02) & 2.9 (0.01)&2.7 (0.03) & 2.9 (0.01) & 2.9 (0.01)&2.6 (0.02) & 2.9 (0.01) & 2.9 (0.01)&2.3 (0.01) & 2.8 (0.01) & 2.9 (0.01)\\ \hline 
M6 Nonlinear & 6 & 36 & 7.2 (0.16) & 8.8 (0.13) & 8.8 (0.11)&6.8 (0.08) & 8.3 (0.08) & 8.4 (0.08)&6.6 (0.05) & 7.9 (0.05) & 8.4 (0.05)&6.3 (0.03) & 7.6 (0.02) & 8.0 (0.02)\\ \hline 
M7 Roll & 2 & 3 & 2.0 (0.01) & 2.3 (0.02) & 2.3 (0.02)&2.0 (0.01) & 2.0 (0.01) & 2.0 (0.01)&2.0 (0.0) & 2.0 (0.0) & 2.0 (0.0)&2.0 (0.0) & 2.0 (0.0) & 2.0 (0.0)\\ \hline M8 Nonlinear & 12 & 72 & 14.9 (0.23) & 19.9 (0.19) & 20.0 (0.21)&14.6 (0.16) & 19.2 (0.12) & 19.3 (0.12)&14.3 (0.1) & 18.6 (0.09) & 19.5 (0.08)&14.0 (0.05) & 18.2 (0.06) & 18.9 (0.04)\\ \hline 
M9 Affine & 20 & 20 & 19.6 (0.06) & 19.3 (0.06) & 19.3 (0.08)&19.5 (0.05) & 19.2 (0.06) & 19.2 (0.06)&19.4 (0.04) & 19.1 (0.06) & 19.4 (0.04)&19.3 (0.02) & 19.0 (0.03) & 19.3 (0.02)\\ \hline 
M10a Cubic & 10 & 11 & 10.0 (0.06) & 10.2 (0.05) & 10.3 (0.04)&10.0 (0.07) & 10.1 (0.06) & 10.2 (0.06)&10.0 (0.04) & 10.0 (0.03) & 10.2 (0.02)&10.0 (0.02) & 10.1 (0.02) & 10.1 (0.02)\\ \hline 
M10b Cubic & 17 & 18 & 17.2 (0.09) & 17.2 (0.09) & 17.3 (0.08)&17.1 (0.04) & 17.1 (0.04) & 17.1 (0.05)&17.0 (0.04) & 17.0 (0.04) & 17.3 (0.04)&17.0 (0.03) & 17.0 (0.03) & 17.2 (0.02)\\ \hline 
M10c Cubic & 24 & 25 & 24.1 (0.13) & 24.1 (0.13) & 24.1 (0.13)&24.0 (0.08) & 24.0 (0.08) & 24.0 (0.09)&24.0 (0.06) & 23.9 (0.06) & 24.3 (0.04)&23.9 (0.03) & 23.9 (0.03) & 24.2 (0.02)\\ \hline 
M10d Cubic & 70 & 72 & 69.8 (0.37) & 67.4 (0.45) & 67.5 (0.41)&69.9 (0.18) & 67.5 (0.21) & 67.5 (0.21)&69.8 (0.1) & 67.3 (0.1) & 69.8 (0.1)&69.7 (0.06) & 67.6 (0.08) & 69.7 (0.06)\\ \hline 
M11 Moebius & 2 & 3 & 2.1 (0.02) & 2.6 (0.03) & 2.7 (0.02)&2.1 (0.01) & 2.3 (0.01) & 2.3 (0.01)&2.0 (0.01) & 2.1 (0.01) & 2.3 (0.01)&2.0 (0.0) & 2.1 (0.0) & 2.1 (0.0)\\ \hline 
M12 Norm & 20 & 20 & 19.8 (0.09) & 19.6 (0.1) & 19.6 (0.11)&19.8 (0.07) & 19.5 (0.08) & 19.6 (0.08)&19.8 (0.03) & 19.6 (0.04) & 19.8 (0.03)&19.8 (0.02) & 19.6 (0.03) & 19.8 (0.02)\\ \hline 
M13a Scurve & 2 & 3 & 2.0 (0.01) & 2.0 (0.01) & 2.1 (0.01)&2.0 (0.01) & 2.0 (0.01) & 2.0 (0.01)&2.0 (0.0) & 2.0 (0.0) & 2.0 (0.0)&2.0 (0.0) & 2.0 (0.0) & 2.0 (0.0)\\ \hline 
M13b Spiral & 1 & 13 & 1.6 (0.01) & 1.9 (0.01) & 1.9 (0.01)&1.9 (0.01) & 1.9 (0.01) & 2.0 (0.01)&1.0 (0.0) & 1.9 (0.0) & 2.0 (0.0)&1.0 (0.0) & 1.9 (0.0) & 2.0 (0.0)\\ \hline Mbeta & 10 & 40 & 6.1 (0.11) & 5.6 (0.11) & 5.9 (0.1)&6.1 (0.06) & 5.6 (0.07) & 5.8 (0.07)&6.2 (0.04) & 5.7 (0.06) & 5.8 (0.06)&6.3 (0.03) & 6.2 (0.02) & 5.9 (0.02)\\ \hline 
Mn1 Nonlinear & 18 & 72 & 18.2 (0.09) & 18.2 (0.09) & 18.4 (0.09)&18.1 (0.07) & 18.1 (0.07) & 18.1 (0.07)&18.0 (0.05) & 17.9 (0.05) & 18.3 (0.04)&18.1 (0.03) & 17.9 (0.03) & 18.1 (0.03)\\ \hline 
Mn2 Nonlinear & 24 & 96 & 23.6 (0.2) & 24.5 (0.18) & 24.6 (0.19)&24.4 (0.09) & 24.4 (0.09) & 24.4 (0.1)&24.2 (0.08) & 24.2 (0.08) & 24.7 (0.08)&24.0 (0.04) & 24.2 (0.04) & 24.5 (0.03)\\ \hline 
Mp1 Paraboloid & 3 & 12 & 3.0 (0.03) & 2.7 (0.04) & 2.9 (0.02)&3.0 (0.02) & 2.8 (0.02) & 2.9 (0.01)&3.0 (0.01) & 2.9 (0.01) & 2.9 (0.01)&3.0 (0.01) & 2.9 (0.0) & 2.9 (0.0)\\ \hline 
Mp2 Paraboloid & 6 & 21 & 5.2 (0.06) & 3.9 (0.12) & 4.8 (0.06)&5.4 (0.04) & 4.5 (0.06) & 5.1 (0.04)&5.5 (0.03) & 4.9 (0.04) & 5.1 (0.04)&5.6 (0.02) & 5.4 (0.01) & 5.3 (0.01)\\ \hline 
Mp3 Paraboloid & 9 & 30 & 7.0 (0.12) & 3.9 (0.13) & 5.8 (0.15)&7.4 (0.08) & 5.2 (0.1) & 6.8 (0.09)&7.7 (0.05) & 6.3 (0.07) & 6.8 (0.05)&7.9 (0.03) & 7.5 (0.02) & 7.4 (0.03)\\  \hline    
\end{tabular} 
\caption{\footnotesize \textbf{Number of samples}. For most low dimensoinal datasets, ESS does not need many points to obtain good accuracy. Notable exceptions are those with interesting geometry, such as M11 Moebius, M5b Helix2d, and M13b Spiral, though in the case of the latter two, the over-estimation bias persists even as the number of samples increases. 
\textbf{High dimensional datasets}. ESS is remarkably accurate on a lot of high dimensional datasets, such as the M10 Cubics, M9 Affine, and M12 Norm, that usually challenge other estimators. This accuracy is achieved even on low sample sizes. 
\textbf{Variance}. Low variance across all datasets. \textbf{Hyperparameter dependency}. ESS is mostly insensitive to the choice of neighbourhood parameter. with some exceptions, such as M13b Spiral even with many point samples. \textbf{Nonlinear datasets}. Unlike other estimators, ESS struggles with the lower dimensional non-linear datasets such as M4, M8,  Mbeta, and Mp1,2,3 Paraboloids, some of which other estimators perform well on; yet ESS performs superbly on the high dimensional Mn1,2, Nonlinear datasets which other estimators really struggle with. }
\label{tab:ess_bm}
\end{sidewaystable}

%% file: tables/FISHERS_benchmark.tex
\begin{sidewaystable}[h]
\centering
\tiny
\begin{tabular}{|l|ll||lll|lll|lll|lll|}
\hline               &    &    & FisherS &              &              &              &              &              &              &              &              &              &              &              \\  \hline              &    &    &              &              &              &              &              &              &              &              &              &              &              &              \\  \hline              &    &    & 625          &              &              & 1250         &              &              & 2500         &              &              & 5000         &              &              \\  \hline               &    &    &              &              &              &              &              &              &              &              &              &              &              &              \\  \hline               & d  & n  & Best         & med abs      & med rel      & Best         & med abs      & med rel      & Best         & med abs      & med rel      & Best         & med abs      & med rel      \\  \hline              &    &    &              &              &              &              &              &              &              &              &              &              &              &              \\  \hline                M1Sphere & 10 & 11 & 10.9 (0.16) & 10.9 (0.2) & 10.9 (0.2)&11.0 (0.06) & 11.0 (0.08) & 11.0 (0.08)&11.0 (0.02) & 11.0 (0.03) & 11.0 (0.03)&11.0 (0.02) & 11.0 (0.03) & 11.0 (0.03)\\ \hline 

M2 Affine 3to5 & 3 & 5 & 2.9 (0.01) & 2.5 (0.02) & 2.5 (0.02)&2.9 (0.01) & 2.5 (0.01) & 2.5 (0.01)&2.9 (0.01) & 2.5 (0.01) & 2.5 (0.01)&2.9 (0.0) & 2.5 (0.01) & 2.5 (0.01)\\ \hline 
M3 Nonlinear 4to6 & 4 & 6 & 3.4 (0.12) & 2.0 (0.09) & 2.0 (0.09)&3.4 (0.07) & 1.9 (0.01) & 1.9 (0.01)&3.4 (0.02) & 1.9 (0.0) & 1.9 (0.0)&3.4 (0.01) & 1.9 (0.0) & 1.9 (0.0)\\ \hline 
M4 Nonlinear & 4 & 8 & 4.1 (0.09) & 4.1 (0.09) & 4.1 (0.09)&4.1 (0.05) & 4.1 (0.05) & 4.1 (0.05)&4.1 (0.03) & 4.1 (0.03) & 4.1 (0.03)&4.1 (0.03) & 4.1 (0.03) & 4.1 (0.03)\\ \hline 
M5a Helix1d & 1 & 3 & 2.4 (0.18) & 2.4 (0.18) & 2.4 (0.18)&2.5 (0.01) & 2.5 (0.01) & 2.5 (0.01)&2.5 (0.01) & 2.5 (0.01) & 2.5 (0.01)&2.5 (0.01) & 2.5 (0.01) & 2.5 (0.01)\\ \hline 
M5b Helix2d & 2 & 3 & 2.0 (0.0) & 1.8 (0.01) & 1.8 (0.01)&2.0 (0.0) & 1.8 (0.0) & 1.8 (0.0)&2.0 (0.21) & 1.8 (0.0) & 1.8 (0.0)&2.0 (0.0) & 1.8 (0.0) & 1.8 (0.0)\\ \hline 
M6 Nonlinear & 6 & 36 & 5.8 (0.11) & 5.8 (0.11) & 5.8 (0.11)&5.8 (0.09) & 5.8 (0.09) & 5.8 (0.09)&5.8 (0.05) & 5.8 (0.05) & 5.8 (0.05)&5.8 (0.04) & 5.8 (0.04) & 5.8 (0.04)\\ \hline 
M7 Roll & 2 & 3 & 2.5 (0.02) & 2.5 (0.02) & 2.5 (0.02)&2.5 (0.02) & 2.5 (0.02) & 2.5 (0.02)&2.5 (0.01) & 2.5 (0.01) & 2.5 (0.01)&2.5 (0.01) & 2.5 (0.01) & 2.5 (0.01)\\ \hline 
M8 Nonlinear & 12 & 72 & 11.0 (0.31) & 11.0 (0.31) & 11.0 (0.31)&10.9 (0.21) & 10.9 (0.21) & 10.9 (0.21)&10.8 (0.15) & 10.8 (0.15) & 10.8 (0.15)&10.8 (0.08) & 10.8 (0.08) & 10.8 (0.08)\\ \hline 
M9 Affine & 20 & 20 & 19.5 (0.57) & 11.3 (0.42) & 11.3 (0.42)&19.2 (0.39) & 11.1 (0.2) & 11.1 (0.2)&18.9 (0.28) & 11.0 (0.12) & 11.0 (0.12)&18.8 (0.15) & 11.1 (0.05) & 11.1 (0.05)\\ \hline 
M10a Cubic & 10 & 11 & 10.5 (0.1) & 7.3 (0.09) & 7.3 (0.09)&10.4 (0.09) & 7.2 (0.06) & 7.2 (0.06)&10.4 (0.07) & 7.2 (0.03) & 7.2 (0.03)&10.3 (0.04) & 7.2 (0.02) & 7.2 (0.02)\\ \hline 
M10b Cubic & 17 & 18 & 17.5 (0.49) & 10.9 (0.23) & 10.9 (0.23)&17.1 (0.37) & 10.8 (0.16) & 10.8 (0.16)&17.0 (0.3) & 10.8 (0.09) & 10.8 (0.09)&16.9 (0.15) & 10.8 (0.05) & 10.8 (0.05)\\ \hline 
M10c Cubic & 24 & 25 & 24.5 (0.85) & 16.2 (1.25) & 16.2 (1.25)&24.2 (0.45) & 14.9 (1.11) & 14.9 (1.11)&23.9 (0.41) & 14.5 (0.32) & 14.5 (0.32)&23.6 (0.32) & 14.4 (0.14) & 14.4 (0.14)\\ \hline 
M10d Cubic & 70 & 72 & nan (nan) & nan (nan) & nan (nan)&nan (nan) & nan (nan) & nan (nan)&nan (nan) & nan (nan) & nan (nan)&nan (nan) & nan (nan) & nan (nan)\\ \hline 
M11 Moebius & 2 & 3 & 2.0 (0.0) & 2.1 (0.01) & 2.1 (0.01)&2.0 (0.0) & 2.1 (0.0) & 2.1 (0.0)&2.0 (0.0) & 2.1 (0.0) & 2.1 (0.0)&2.0 (0.0) & 2.1 (0.0) & 2.1 (0.0)\\ \hline 
M12 Norm & 20 & 20 & 19.9 (0.33) & 9.2 (0.18) & 9.2 (0.18)&19.9 (0.34) & 9.1 (0.13) & 9.1 (0.13)&19.9 (0.11) & 9.0 (0.08) & 9.0 (0.08)&20.0 (0.18) & 8.9 (0.05) & 8.9 (0.05)\\ \hline 
M13a Scurve & 2 & 3 & 1.9 (0.02) & 1.8 (0.01) & 1.8 (0.01)&1.9 (0.01) & 1.8 (0.01) & 1.8 (0.01)&1.9 (0.01) & 1.8 (0.01) & 1.8 (0.01)&1.9 (0.01) & 1.8 (0.0) & 1.8 (0.0)\\ \hline 
M13b Spiral & 1 & 13 & 2.0 (0.0) & 2.0 (0.0) & 2.0 (0.0)&2.0 (0.0) & 2.0 (0.0) & 2.0 (0.0)&2.0 (0.0) & 2.0 (0.0) & 2.0 (0.0)&2.0 (0.0) & 2.0 (0.0) & 2.0 (0.0)\\ \hline 
Mbeta & 10 & 40 & 5.3 (0.1) & 3.6 (0.03) & 3.6 (0.03)&5.3 (0.05) & 3.6 (0.02) & 3.6 (0.02)&5.3 (0.05) & 3.6 (0.02) & 3.6 (0.02)&5.3 (0.04) & 3.6 (0.01) & 3.6 (0.01)\\ \hline
Mn1 Nonlinear & 18 & 72 & 18.4 (0.78) & 9.9 (0.32) & 9.9 (0.32)&18.3 (0.53) & 9.9 (0.18) & 9.9 (0.18)&17.4 (0.43) & 9.9 (0.12) & 9.9 (0.12)&18.4 (0.22) & 9.9 (0.06) & 9.9 (0.06)\\ \hline 
Mn2 Nonlinear & 24 & 96 & 23.9 (0.92) & 13.6 (1.13) & 13.6 (1.13)&24.9 (1.0) & 12.9 (0.36) & 12.9 (0.36)&22.9 (0.75) & 12.9 (0.22) & 12.9 (0.22)&23.3 (0.31) & 12.9 (0.09) & 12.9 (0.09)\\ \hline 
Mp1 Paraboloid & 3 & 12 & 1.4 (0.01) & 1.4 (0.01) & 1.4 (0.01)&1.4 (0.01) & 1.4 (0.01) & 1.4 (0.01)&1.4 (0.01) & 1.4 (0.01) & 1.4 (0.01)&1.4 (0.0) & 1.4 (0.0) & 1.4 (0.0)\\ \hline 
Mp2 Paraboloid & 6 & 21 & 1.3 (0.01) & 1.3 (0.01) & 1.3 (0.01)&1.3 (0.01) & 1.3 (0.01) & 1.3 (0.01)&1.3 (0.01) & 1.3 (0.01) & 1.3 (0.01)&1.3 (0.0) & 1.3 (0.0) & 1.3 (0.0)\\ \hline 
Mp3 Paraboloid & 9 & 30 & 1.4 (0.01) & 1.4 (0.01) & 1.4 (0.01)&1.3 (0.01) & 1.3 (0.01) & 1.3 (0.01)&1.3 (0.01) & 1.3 (0.01) & 1.3 (0.01)&1.3 (0.0) & 1.3 (0.0) & 1.3 (0.0)\\ \hline    
\end{tabular} 
\caption{\footnotesize \textbf{Number of samples}. The performance of FisherS does not significantly improve when more samples are added. Even with many samples, FisherS can be biased for some low-dimensional datasets, such as M3 Nonlinear 4to6, M5a Helix1d, and M7 Roll. 
\textbf{High dimensional datasets}. With the correct hyperparameters, FisherS can have good performances on high dimensional datasets, but it is quite dependent on a good choice. As FisherS relies on inverting a non-linear function in its routine, for M10d Cubic, the estimator returned an error. 
\textbf{Variance}. Variance is mostly low, but for some nonlinear datasets such as Mn1,2 Nonlinear with few samples, the variance can be high.  \textbf{Hyperparameter dependency}. On nonlinear datasets such as M13 Norm, Mbeta, Mn1,2 Nonlinear, and M4 Nonlinear; and high dimensional datasets such as M9 Affine, and M10 Cubics, there is a significant dependency on the hyperparameter choice. \textbf{Nonlinear datasets}. The performance on nonlinear datasets depends on hyperparameter choices, though it grossly underestimates on Mp1,2,3 Paraboloids.  }
\label{tab:fishers_bm}
\end{sidewaystable}

%% file: tables/Cdim_benchmark1.tex
\begin{sidewaystable}[h]
\centering
\tiny
\begin{tabular}{|l|ll||lll|lll|lll|lll|}
\hline
               CDim&    &    & \multicolumn{3}{c|}{n = 625} & \multicolumn{3}{c|}{n = 1250} & \multicolumn{3}{c|}{n = 2500} & \multicolumn{3}{c|}{n = 5000} \\
\hline
               &    &    & Best         & Med Abs      & Med Rel      & Best         & Med Abs      & Med Rel      & Best         & Med Abs      & Med Rel      & Best         & Med Abs      & Med Rel      \\
\hline
M1\_Sphere      & 10 & 11 & 5.0          & 5.0          & 5.0          & 5.1          & 5.1          & 5.1          & 5.5          & 5.5          & 5.5          & 6.0          & 6.0          & 5.7          \\ \hline
M2\_Affine\_3to5 & 3  & 5  & 3.0          & 3.0          & 3.0          & 3.0          & 3.0          & 3.0          & 3.0          & 3.0          & 3.0          & 3.0          & 3.0          & 3.0          \\ \hline
M3\_Nonlinear\_4to6 & 4  & 6  & 4.0          & 4.0          & 4.0          & 4.0          & 3.9          & 3.9          & 4.0          & 4.0          & 4.0          & 4.0          & 4.0          & 4.0          \\ \hline
M4\_Nonlinear   & 4  & 8  & 4.0          & 4.0          & 4.0          & 4.0          & 4.2          & 4.2          & 4.0          & 4.2          & 4.2          & 4.0          & 4.0          & 4.1          \\ \hline
M5a\_Helix1d    & 1  & 3  & 1.0          & 2.0          & 2.0          & 1.0          & 1.7          & 1.4          & 1.0          & 1.1          & 1.1          & 1.0          & 1.0          & 1.0          \\ \hline
M5b\_Helix2d    & 2  & 3  & 2.0          & 3.0          & 3.0          & 2.0          & 3.0          & 3.0          & 2.0          & 3.0          & 3.0          & 2.0          & 3.0          & 3.0          \\ \hline
M6\_Nonlinear   & 6  & 36 & 5.0          & 4.0          & 5.0          & 5.0          & 4.6          & 4.4          & 5.0          & 4.7          & 4.7          & 5.0          & 5.0          & 4.8          \\ \hline
M7\_Roll        & 2  & 3  & 2.0          & 2.0          & 3.0          & 2.0          & 2.5          & 2.4          & 2.0          & 2.1          & 2.1          & 2.0          & 2.0          & 2.0          \\ \hline
M8\_Nonlinear   & 12 & 72 & 3.0          & 3.0          & 3.0          & 3.3          & 3.3          & 3.0          & 3.6          & 3.6          & 3.6          & 4.0          & 4.0          & 3.8          \\ \hline
M9\_Affine      & 20 & 20 & 3.3          & 3.0          & 3.0          & 4.0          & 3.7          & 3.3          & 4.0          & 4.0          & 4.0          & 4.2          & 4.0          & 4.2          \\ \hline
M10a\_Cubic     & 10 & 11 & 5.0          & 4.0          & 5.0          & 5.0          & 4.9          & 4.8          & 5.1          & 5.1          & 5.1          & 5.2          & 5.0          & 5.2          \\ \hline
M10b\_Cubic     & 17 & 18 & 4.0          & 3.0          & 4.0          & 4.0          & 4.0          & 3.7          & 4.3          & 4.3          & 4.3          & 5.0          & 5.0          & 4.6          \\ \hline
M10c\_Cubic     & 24 & 25 & 3.0          & 3.0          & 3.0          & 3.0          & 3.0          & 2.7          & 3.4          & 3.4          & 3.4          & 4.0          & 4.0          & 3.6          \\ \hline
M10d\_Cubic     & 70 & 72 & 1.0          & 1.0          & 1.0          & 1.0          & 1.0          & 1.0          & 1.0          & 1.0          & 1.0          & 1.1          & 1.0          & 1.1          \\ \hline
M11\_Moebius    & 2  & 3  & 2.0          & 3.0          & 3.0          & 2.0          & 2.8          & 2.7          & 2.0          & 2.3          & 2.3          & 2.0          & 2.0          & 2.1          \\ \hline
M12\_Norm       & 20 & 20 & 3.1          & 3.0          & 3.0          & 3.3          & 3.3          & 2.8          & 3.5          & 3.5          & 3.5          & 4.0          & 4.0          & 3.6          \\ \hline
M13a\_Scurve    & 2  & 3  & 2.0          & 2.0          & 2.0          & 2.0          & 2.0          & 2.0          & 2.0          & 2.0          & 2.0          & 2.0          & 2.0          & 2.0          \\ \hline
M13b\_Spiral    & 1  & 13 & 1.8          & 2.0          & 2.0          & 1.6          & 2.0          & 2.0          & 1.0          & 2.0          & 2.0          & 1.0          & 2.0          & 2.0          \\ \hline
Mbeta           & 10 & 40 & 4.0          & 3.0          & 4.0          & 4.0          & 3.8          & 3.7          & 4.0          & 4.0          & 4.0          & 4.1          & 4.0          & 4.1          \\ \hline
Mn1\_Nonlinear  & 18 & 72 & 3.4          & 3.0          & 3.0          & 4.0          & 3.7          & 3.3          & 4.0          & 3.9          & 3.9          & 4.2          & 4.0          & 4.2          \\ \hline
Mn2\_Nonlinear  & 24 & 96 & 3.0          & 2.0          & 3.0          & 3.0          & 2.9          & 2.5          & 3.1          & 3.1          & 3.1          & 3.4          & 3.0          & 3.4          \\ \hline
Mp1\_Paraboloid & 3  & 12 & 3.0          & 3.0          & 3.0          & 3.0          & 3.0          & 3.0          & 3.0          & 3.0          & 3.0          & 3.0          & 3.0          & 3.0          \\ \hline
Mp2\_Paraboloid & 6  & 21 & 4.0          & 4.0          & 4.0          & 4.1          & 4.1          & 3.9          & 4.4          & 4.4          & 4.4          & 5.0          & 5.0          & 4.5          \\ \hline
Mp3\_Paraboloid & 9  & 30 & 4.0          & 3.0          & 4.0          & 4.0          & 3.8          & 3.6          & 4.1          & 4.1          & 4.1          & 4.4          & 4.0          & 4.4          \\
\hline
\end{tabular}
\caption{Only one run of the data sets was performed due to the computation time, so no variance is reported. Overall, the estimator struggles with higher dimensions but remains accurate on lower dimensions regardless of codimension. The very slight improvements on higher-dimensional datasets when there are more points suggests that the number of points needed to accurately estimate dimension in these cases is extremely large.}
\end{sidewaystable}

%% file: tables/noise/6_11_gaussian.tex
\begin{table}[!ht]
\footnotesize
    \centering
    \begin{tabular}{|l||l|l|l|l|}
    \hline
        Estimator & $\sigma^2=0.0$ & $\sigma^2=0.01$ & $\sigma^2=0.1$ & $\sigma^2=1.0$ \\ \hline
        \lpca & 7.00 (0.000) & 10.00 (0.000) & 10.00 (0.000) & 10.00 (0.000) \\ \hline
        \mle & 5.89 (0.06) & 5.92 (0.06) & 8.04 (0.10) & 10.36 (0.17) \\ \hline 
        \phzero & 5.94 (0.13) & {5.97 (0.10)} & 8.21 (0.14) & 10.69 (0.33)  \\ \hline
        \knnestimator & 6.0 (0.5) & 5.97 (0.45) & 9.17 (1.15) & 10.75 (1.09)  \\ \hline
        \wodcap & 6.93 (0.0) & 6.93 (0.0) & 6.93 (0.0) & 9.15 (0.0)   \\ \hline
        \gride & 5.86 (0.123) & 8.60 (0.150) & 10.29 (0.191) & 10.45 (0.207) \\ \hline
        \twonn & 5.85 (0.153) & 8.58 (0.209) & 10.23 (0.200) & 10.42 (0.242) \\ \hline
        \danco & 6.92 (0.067) & 8.95 (0.288) & 11.00 (0.000) & 11.00 (0.000) \\ \hline
        \mindml & {6.00 (0.000)} & 9.00 (0.490) & 10.00 (0.000) & 10.00 (0.000) \\ \hline
        \corrint & 5.82 (0.070) & 7.48 (0.083) & 9.00 (0.100) & 8.95 (0.088) \\ \hline
        \ess & 6.07 (0.027) & 7.87 (0.043) & 10.36 (0.052) & 10.68 (0.063) \\ \hline
        \fishers & {6.98 (0.018)} & {5.63 (0.021)} & {5.82 (0.033)} & {5.51 (0.047)} \\ \hline        
        \tle & 6.28 (0.108) & 7.83 (0.141) & 9.74 (0.176) & 10.00 (0.177) \\ \hline
    \end{tabular}
    \vspace{1em}
    \caption{Mean estimated dimension along with standard deviation for the unit sphere $S^6 \subset \mathbb{R}^{11}$ with ambient Gaussian noise. The variance of the noise is given at the top of the columns. }
    \label{tab:611_gauss}
\end{table}

%% file: tables/noise/6_11_outliers.tex
\begin{table}[!ht]
\footnotesize
    \centering
    \begin{tabular}{|l||l|l|l|l|}
    \hline
        Estimator & n=0 & n=25 & n=125 & n=250 \\ \hline
        \lpca & 7.00 (0.000) & 7.00 (0.000) & 7.00 (0.000) & 7.00 (0.000) \\ \hline
        \mle & 5.89 (0.06) & 5.93 (0.07) & 5.98 (0.07) & 6.01 (0.07) \\ \hline
        \phzero & 5.95 (0.12) & 6.01 (0.10) & 6.18 (0.21) & 6.18 (0.24) \\ \hline
        \knnestimator & 6.0 (0.5) & 6.3 (0.67) & 6.26 (0.85) & 6.24 (0.79)   \\ \hline
        \wodcap & 6.93 (0.0) & 6.93 (0.0) & 6.93 (0.0) & 6.93 (0.0)  \\ \hline
        \gride & 5.96 (0.167) & 5.85 (0.102) & 5.97 (0.131) & 5.85 (0.133) \\ \hline
        \twonn & 5.88 (0.144) & 5.92 (0.146) & 5.87 (0.141) & 5.91 (0.120) \\ \hline
        \danco & 6.95 (0.117) & 6.98 (0.055) & 7.01 (0.019) & 6.97 (0.015) \\ \hline
        \mindml & {6.00 (0.000)} & {6.00 (0.000)} & {6.00 (0.000)} & {6.00 (0.000)} \\ \hline
        \corrint & 5.83 (0.053) & 5.79 (0.063) & 5.80 (0.050) & 5.83 (0.052) \\ \hline
        \ess & 6.07 (0.025) & 6.08 (0.029) & {6.00 (0.066)} & 5.97 (0.080) \\ \hline
        \fishers & 6.99 (0.016) & 5.81 (0.038) & 4.40 (0.052) & 3.71 (0.069) \\ \hline
        \tle & 6.29 (0.109) & 6.27 (0.107) & 6.26 (0.107) & 6.30 (0.116) \\ \hline 
    \end{tabular}
    \vspace{1em}
    \caption{Mean estimated dimension along with standard deviations for $S^6 \subset \mathbb{R}^{11}$ with uniform background noise (outliers). The number of outlier points is given at the top of the columns.}
\end{table}


%% file: tables/noise/10_11_gaussian.tex
\begin{table}[!ht]
\footnotesize
    \centering
    \begin{tabular}{|l||l|l|l|l|}
    \hline
        Estimator & $\sigma^2=0.0$ & $\sigma^2=0.01$ & $\sigma^2=0.1$ & $\sigma^2=1.0$ \\ \hline
        \lpca & 10.0 (0.000) & {10.00 (0.000)} & {10.00 (0.000)} & {10.00 (0.000)} \\ \hline
        \mle & 9.26 (0.14) &  9.25 (0.13) & 9.61 (0.14) & 10.31 (0.15) \\ \hline
        \phzero & 9.38 (0.16) & 9.34 (0.18) & 9.78 (0.16) & 10.54 (0.27) \\ \hline
        \knnestimator & 9.91 (1.03) & 9.99 (0.99) & 10.29 (0.9) & 10.11 (0.98)  \\ \hline
        \wodcap & 9.15 (0.0) & 9.15 (0.0) & 9.15 (0.0) & 9.15 (0.0)  \\ \hline 
        \gride & 9.4 (0.200) & 9.77 (0.178) & 10.48 (0.213) & 10.58 (0.227) \\ \hline
        \twonn & 9.4 (0.210) & 9.82 (0.178) & 10.51 (0.265) & 10.55 (0.268) \\ \hline
        \mindml & 9.4 (0.05) & {10.00 (0.218)} & {10.00 (0.000)} & {10.00 (0.000)} \\ \hline
        \corrint &9.1 (0.060)  & 9.32 (0.097) & 9.22 (0.066) & 9.02 (0.072) \\ \hline
        \ess & 10.0 (0.030) & 10.27 (0.055) & 10.65 (0.058) & 10.69 (0.054) \\ \hline
        \fishers & 11.0 (0.020) & 7.87 (0.038) & 5.82 (0.039) & 5.47 (0.046) \\ \hline
        \danco & 10.9 (0.300) & 11.00 (0.000) & 11.00 (0.000) & 11.00 (0.000) \\ \hline
        \tle & 10.1 (0.050) & 9.89 (0.192) & 10.02 (0.178) & 9.97 (0.181) \\ \hline
    \end{tabular}
    \vspace{1em}
    \caption{Mean estimated dimension along with standard deviations for $S^{10} \subset \mathbb{R}^{11}$ with ambient Gaussian noise. The variance of the noise is given at the top of the columns. }
\end{table}

%% file: tables/noise/10_11_outliers.tex
\begin{table}[!ht]
\footnotesize
    \centering
    \begin{tabular}{|l||l|l|l|l|}
    \hline
        Estimator & n=0 & n=25 & n=125 & n=250 \\ \hline
        \lpca & {10.00 (0.000)} & {10.00 (0.000)} & {10.00 (0.000)} & {10.00 (0.000)} \\ \hline
        \mle & 9.26 (0.14) & 9.33 (0.14) & 9.49 (0.13) & 9.58 (0.13) \\ \hline
        \phzero & 9.36 (0.22) & 9.47 (0.30)& 9.74 (0.38) & 10.05 (0.45) \\ \hline
        \knnestimator & 9.91 (1.03) & 10.17 (1.31) & 9.72 (1.37) & 9.18 (1.42)     \\ \hline
        \wodcap & 9.15 (0.0) & 9.15 (0.0) & 9.15 (0.0) & 9.15 (0.0)  \\ \hline 
        \gride & 9.4 (0.200) & 9.48 (0.222) & 9.53 (0.204) & 9.52 (0.245) \\ \hline
        \twonn & 9.4 (0.210) & 9.49 (0.275) & 9.49 (0.263) & 9.48 (0.230) \\ \hline
        \danco & 10.9 (0.300) & 11.00 (0.000) & 11.00 (0.000) & 11.00 (0.000) \\ \hline
        \mindml & 9.4 (0.050) & {10.00 (0.458)} & {10.00 (0.490)} & {10.00 (0.497)} \\ \hline
        \corrint & 9.1 (0.060) & 9.14 (0.090) & 9.11 (0.081) & 9.16 (0.136) \\ \hline
        \ess & 10.0 (0.030) & 9.97 (0.051) & 9.88 (0.106) & 9.66 (0.248) \\ \hline
        \fishers & 11.0 (0.020) & 6.65 (0.052) & 4.78 (0.051) & 4.07 (0.068) \\ \hline
         \tle & 10.1 (0.050) & 9.72 (0.189) & 9.64 (0.201) & 9.55 (0.193) \\ \hline

    \end{tabular}
    \vspace{1em}
    \caption{Mean estimated dimension along with standard deviations for $S^{10} \subset \mathbb{R}^{11}$ with uniform background noise (outliers). The number of outlier points is given at the top of the columns. }
\label{tab:1011_out}
\end{table}

%% file: tables/noise/6_16_so_gaussian.tex
\begin{table}[!ht]
\footnotesize
    \centering
    \begin{tabular}{|l||l|l|l|l|}
    \hline
        Estimator & $\sigma^2=0.0$ & $\sigma^2=0.01$ & $\sigma^2=0.1$ & $\sigma^2=1.0$ \\ \hline
        \lpca & 10.00 (0.000) & 10.00 (0.000) & 10.00 (0.000) & 10.00 (0.000) \\ \hline
        \mle & 6.14 (0.10) & 6.15 (0.162) & {7.19 (0.10)} & 13.95 (0.12) \\ \hline
        \phzero & 6.16 (0.14) & {6.14 (0.13)} & {7.20 (0.15)} & 14.48 (0.31) \\ \hline
        \knnestimator & 5.91 (0.53) & 5.87 (0.65) & 7.17 (0.63) & 15.2 (2.69)  \\ \hline
        \wodcap & 6.15 (0.03) & 6.16 (0.03) & 6.36 (0.05) & 8.81 (0.03) \\ \hline 
        \gride & 6.17 (0.127) & 7.46 (0.144) & 12.10 (0.302) & 14.38 (0.327) \\ \hline
        \twonn & 6.22 (0.180) & 7.45 (0.162) & 12.12 (0.348) & 14.26 (0.265) \\ \hline
        \danco & 7.02 (0.016) & 8.00 (0.014) & 13.01 (0.293) & 16.00 (0.000) \\ \hline
        \mindml & {6.00 (0.218)} & 8.00 (0.458) & 10.00 (0.000) & 10.00 (0.000) \\ \hline
        \corrint & 6.20 (0.089) & 6.94 (0.077) & 10.62 (0.091) & 11.83 (0.174) \\ \hline
        \ess & 7.63 (0.043) & 8.40 (0.047) & 12.90 (0.083) & 15.34 (0.074) \\ \hline
        \fishers & 9.03 (0.040) & 8.76 (0.024) & 8.22 (0.046) & {7.50 (0.061)} \\ \hline
        \tle & 6.88 (0.131) & 7.51 (0.136) & 11.01 (0.229) & 13.01 (0.251) \\ \hline
    \end{tabular}
        \vspace{1em}
    \caption{Mean estimated dimension along with standard deviation for the $SO(4)$ dataset with ambient Gaussian noise. The variance of the noise is given at the top of the columns. }
    \label{tab:616_gauss}
\end{table}

%% file: tables/noise/6_16_so_outliers.tex
\begin{table}[!ht]
\footnotesize
    \centering
    \begin{tabular}{|l||l|l|l|l|}
    \hline
        Estimator & n=0 & n=25 & n=125 & n=250 \\ \hline
        \lpca & 10.00 (0.000) & 10.00 (0.000) & 10.00 (0.000) & 10.00 (0.000) \\ \hline
        \mle & 6.14 (0.10) & 6.20 (0.10) & 6.34 (0.11) & 6.46 (0.11) \\ \hline
        \phzero & 6.15 (0.13) & 6.31 (0.18) & 6.55 (0.21) & 6.70 (0.27) \\ \hline
        \knnestimator & 5.91 (0.53) & 6.26 (0.58) & 6.34 (0.75) & 6.07 (0.62)  \\ \hline
        \wodcap & 6.15 (0.03) & 6.18 (0.03) & 6.29  (0.04) & 6.43 (0.04) \\ \hline
        \gride & 6.10 (0.127) & 6.16 (0.124) & 6.26 (0.102) & 6.15 (0.151) \\ \hline
        \twonn & 6.06 (0.158) & 6.07 (0.135) & 6.22 (0.103) & 6.10 (0.170) \\ \hline
        \danco & 7.01 (0.108) & 7.37 (0.101) & 7.95 (0.065) & 7.98 (0.014) \\ \hline
        \mindml & 6.00 (0.000) & 6.00 (0.000) & 6.00 (0.000) & 6.00 (0.218) \\ \hline
        \corrint & 6.19 (0.052) & 6.18 (0.067) & 6.18 (0.060) & 6.19 (0.067) \\ \hline
        \ess & 7.61 (0.044) & 7.62 (0.039) & 7.61 (0.072) & 7.49 (0.151) \\ \hline
        \fishers & 9.03 (0.039) & 6.98 (0.042) & 5.25 (0.076) & 4.53 (0.053) \\ \hline
        \tle & 6.88 (0.128) & 6.85 (0.129) & 6.82 (0.132) & 6.77 (0.132) \\ \hline
    \end{tabular}
    \vspace{1em}
    \caption{Mean estimated dimension along with standard deviations for $SO(4)$ with uniform background noise (outliers). The number of outlier points is given at the top of the columns. }
\end{table}